\documentclass[journal,twoside]{IEEEtran}

\usepackage{color}
\usepackage{amssymb}
\usepackage{rotating}
\usepackage[T1]{fontenc}
\usepackage{lscape}
\usepackage{dblfloatfix}
\usepackage{xcolor}
\usepackage{subcaption}
\usepackage[ruled,vlined]{algorithm2e}
\usepackage[autostyle]{csquotes}
\usepackage{tikz}
\usepackage{acro}
\usepackage{siunitx}
\usepackage{enumitem}
\usepackage{multirow}
\usepackage{cite}
\usepackage{hyperref}
\usepackage{amsmath,amssymb,amsfonts}
\usepackage{tikz}
\usepackage{float}
\usepackage{algorithmic}
\usepackage{graphicx}
\usepackage{textcomp}
\usepackage{xcolor}
\usepackage{url}
\usepackage{todonotes}
\usepackage{tabularray}
\usepackage{longtable}
\usepackage{tabularx}
\usepackage{mathrsfs}
\DeclareUnicodeCharacter{221E}{\ensuremath{\infty}}
\DeclareUnicodeCharacter{043F}{\cyrr}

\usepackage{comment}
\def\BibTeX{{\rm B\kern-.05em{\sc i\kern-.025em b}\kern-.08em
    T\kern-.1667em\lower.7ex\hbox{E}\kern-.125emX}}

\ifCLASSINFOpdf
\else
 
\fi

\usepackage[a4paper, total={170mm,257mm}]{geometry}

\begin{document}

%\title{Perspectives in hybrid edge-cloud computing: a case study on energy efficiency in buildings}
%\title{A hybrid tow-stage edge-based anomaly detection of building energy consumption}

%\title{Anomaly-Net: Deep Anomaly Detection of Building Energy Consumption Using Time-Series Imaging}

\title{A Comprehensive Review of Recent Research Trends on UAVs}


\author{\IEEEauthorblockN{Kaled Telli\IEEEauthorrefmark{1},
Okba Kraa\IEEEauthorrefmark{1}, 
Yassine Himeur\IEEEauthorrefmark{2}, 
Abdelmalik Ouamane\IEEEauthorrefmark{3},
Mohamed Boumehraz\IEEEauthorrefmark{1},
Shadi Atalla\IEEEauthorrefmark{2} and
Wathiq Mansoor\IEEEauthorrefmark{2}
}\\
\IEEEauthorblockA{\IEEEauthorrefmark{1}
Energy Systems Modelling (MSE) Laboratory, Mohamed Khider University, Biskra, Algeria}\\
\IEEEauthorblockA{\IEEEauthorrefmark{2}College of Engineering and Information Technology, University of Dubai, Dubai, UAE}\\
\IEEEauthorblockA{\IEEEauthorrefmark{3}Laboratory of LI3C, Mohamed Khider University, Biskra, Algeria}\\
}


\maketitle

\begin{abstract}
The growing interest in unmanned aerial vehicles (UAVs) from both scientific and industrial sectors has attracted a wave of new researchers and substantial investments in this expansive field. However, due to the wide range of topics and subdomains within UAV research, newcomers may find themselves overwhelmed by the numerous options available. It is therefore crucial for those involved in UAV research to recognize its interdisciplinary nature and its connections with other disciplines.
This paper presents a comprehensive overview of the UAV field, highlighting recent trends and advancements. Drawing on recent literature reviews and surveys, the review begins by classifying UAVs based on their flight characteristics. It then provides an overview of current research trends in UAVs, utilizing data from the Scopus database to quantify the number of scientific documents associated with each research direction and their interconnections.
The paper also explores potential areas for further development in UAVs, including communication, artificial intelligence, remote sensing, miniaturization, swarming and cooperative control, and transformability. Additionally, it discusses the development of aircraft control, commonly used control techniques, and appropriate control algorithms in UAV research.
Furthermore, the paper addresses the general hardware and software architecture of UAVs, their applications, and the key issues associated with them. It also provides an overview of current open-source software and hardware projects in the UAV field.
By presenting a comprehensive view of the UAV field, this paper aims to enhance understanding of this rapidly evolving and highly interdisciplinary area of research.
\end{abstract}


\IEEEpeerreviewmaketitle
\section{Introduction}
Artificial intelligence (AI) has become increasingly significant across various sectors due to its transformative capabilities, such as robotics, manufacturing and automation, healthcare \cite{himeur2023face}, cybersecurity \cite{kheddar2023deep1}, education \cite{atalla2023intelligent}, energy and utilities \cite{copiaco2023innovative,himeur2022next}, smart cities \cite{elnour2022performance}, natural language processing and human-computer interaction \cite{kheddar2023deep2}, agriculture \cite{atalla2023iot}, transportation and logistics \cite{al2022smart}. Similarly, AI plays a critical role in unmanned aerial vehicles (UAVs), enhancing their capabilities in navigation \cite{khalife2022achievability}, object detection \cite{elharrouss2021panoptic}, and mission planning \cite{liu2022hybrid}.

The versatility and efficiency of UAVs have made them increasingly popular for a wide range of tasks. As the demand for UAVs continues to grow, it is crucial to stay updated with the latest developments and research in this field. UAVs have attracted significant attention in the scientific community, as demonstrated by numerous review papers \cite{1,2,3,4,5,6,pina2022uavs} that explore various aspects of UAV development and research across different applications. Several key areas have garnered particular interest in UAV \cite{6,7,8}, including the use of open-source hardware and software in recent UAVs \cite{9,10,12}, frame designing and optimization \cite{13,14}, control systems \cite{15,17}, both conventional \cite{18} and modern \cite{19,20} communication modalities (such as 5G networks), integration of AI \cite{21}, recognition and detection algorithms, and path planning strategies \cite{22}. These areas play a pivotal role in advancing UAV technology and are critical subjects of investigation for researchers and practitioners alike.

Due to the wide range of subdomains and extensive scope of UAV research, coupled with significant investments in this multifaceted field, several important research questions have arisen. These include the interdisciplinary nature of UAV research, the challenges and opportunities presented by UAV technology, and the future directions of UAV research \cite{chen2023yolo}.
The field of UAVs has experienced rapid growth and has captured the attention of researchers worldwide. With its diverse subdomains and expansive nature, this field has become a vibrant and active area of study, attracting substantial investments. As researchers in this field, we are constantly seeking answers to pressing research questions. We recognize the need for a comprehensive guide to navigating the array of options available in UAV research \cite{othman2023development}.

The rapid growth and wide-ranging applications of unmanned aerial vehicles (UAVs) have given rise to several important research questions. One of these questions pertains to the interdisciplinary nature of UAV research. UAVs involve a convergence of various disciplines, including aerospace engineering, computer science, robotics, and remote sensing. Understanding the interplay between these disciplines and identifying effective collaboration strategies are crucial for advancing UAV technology \cite{chen2023yolo}.
Another important area of inquiry is the exploration of the challenges and opportunities presented by UAV technology. While UAVs offer numerous advantages such as improved efficiency, cost-effectiveness, and enhanced data collection capabilities, they also face challenges such as regulatory frameworks, privacy concerns, and safety issues. Investigating these challenges and finding solutions will contribute to the responsible and effective integration of UAVs into society \cite{othman2023development}.
%Furthermore, researchers are actively investigating the future directions of UAV research. As technology advances and new capabilities emerge, exploring innovative applications of UAVs becomes paramount. Future research directions may include advancements in autonomous flight, artificial intelligence integration, swarm robotics, energy efficiency, sensor miniaturization, and beyond-visual line of sight (BVLOS) operations. Understanding these trends and anticipating future developments is essential for shaping the future of UAV technology.
As researchers in the field of UAVs, we recognize the importance of addressing these research questions. Our aim is to provide a comprehensive guide that navigates the vast landscape of UAV research. By synthesizing the latest findings and insights from various subdomains, we hope to provide a valuable resource for researchers and practitioners in this dynamic field. Through collaboration and knowledge sharing, we can collectively advance UAV technology and unlock its full potential in a wide range of applications.

In the realm of Unmanned Aerial Vehicles (UAVs) or drones, several pressing research questions exist. Key among these includes exploring how we can enhance UAV autonomy by integrating machine learning and AI into their systems for improved functionality and decision-making \cite{chen2023yolo}. The improvement of navigation and control systems for precision manoeuvres in unpredictable environments is a significant area of focus, as is developing advanced "sense and avoid" systems for reliable obstacle detection. The application of swarm intelligence in UAVs to facilitate collaborative tasks and the implementation of efficient algorithms is being studied \cite{othman2023development}. Researchers are also looking into extending UAV battery life and investigating efficient power management strategies and alternative energy sources. The ability to increase UAV payload capacity without compromising efficiency or manoeuvrability and how drones can be adapted for specific payload types is under scrutiny. Equally important are the security concerns surrounding UAV systems against potential cyber-attacks or hijacking, and measures to protect individual privacy from misuse of surveillance-capable UAVs. Questions abound about how UAVs can be safely incorporated into crowded airspace, particularly in urban environments or near airports, and what changes to air traffic control systems are required to accommodate them. Regulatory implications of widespread UAV use are also on the table, focusing on how laws and regulations should adapt to handle UAV use. Finally, the aspect of human-UAV interaction in terms of safe and effective design for human interaction and improving the user experience is being delved into. Each of these research questions presents an exciting challenge in shaping the future of UAV technology.

In response to these needs, this paper serves as a comprehensive guide for new researchers venturing into the multifaceted and expansive subdomains of UAV research. Recognizing the vast scope of this field, the guide aims to establish a strong foundation for novice researchers by providing a thorough review and survey of each subfield. It encompasses popular UAV classifications \cite{24}, which categorize UAVs based on their size, range, and endurance.
In addition to covering UAV classifications, this paper provides an overview of crucial aspects such as hardware architecture, recent research trends, open-source initiatives, and software tools employed in UAV development and research. The research direction for UAVs has witnessed remarkable growth in recent years, and this paper meticulously analyzes these trends and investigates the interconnections among various research directions.
Critical areas addressed in the paper include communication and antennas, the internet of things (IoTs), aircraft detection, control and autonomous flight, perception and sensing, energy-efficient flight, human-UAV interaction, swarm behaviour, and more \cite{6,7,8}. Notably, the paper highlights the significant impact of utilizing UAVs in animal studies, enabling non-invasive monitoring, precise data collection, and reduced disturbance to wildlife habitats \cite{pina2022uavs,fudala2022use}.
By providing a comprehensive overview and synthesizing reliable references, this paper equips researchers with the necessary knowledge and resources to make significant contributions to the field of UAV research. It aims to foster exploration, innovation, and collaboration, ultimately driving the advancement and potential of UAV technology.

\textcolor{black}{The paper explores potential open-development axes for UAVs, including AI integration \cite{21}, environmental monitoring \cite{fudala2022use,26,27,28,29,zmarz2018application}, conservation \cite{29,ezequiel2014uav}, miniaturization \cite{32}, swarming, and cooperative control \cite{1,2,5,8}, and transformability systems \cite{34,35,36,37}. Aircraft control development is a crucial aspect of UAV research, necessitating consideration of appropriate control algorithms and commonly used techniques \cite{15,17}, providing valuable insights into UAV research controls.} The main contributions of this study are as follows:
\begin{itemize}
\item A comprehensive collection of relevant references related to the drone field, serving as a reliable and accessible source for researchers in this domain.
\item Insights and predictions established through a rigorous scientific approach regarding the most active and rapidly expanding research directions in the UAV field over the past three years. The analysis is based on growth rate per year and acceleration, supported by robust evidence.
\item Identification of potential UAV Open Development Axes, offering valuable insights and ideas for future research directions.
A systematic address of the Consideration for Appropriate Control Algorithm of UAVs, providing an in-depth analysis of this critical aspect of UAV research.
\item An overview of high-level UAV development software achieved through a systematic classification process, serving as an accessible guide to available options in this area of UAV research.
\item A rigorous extraction of the most prominent research directions in the UAV domain over the past three years, employing a scientifically sound methodology for a comprehensive understanding of the current state-of-the-art in UAV research.
\item Presentation of a numerical analysis of the interrelationships among UAV research directions, offering clear insights into the current landscape of UAV research, facilitating effective charting of future UAV research efforts.

\end{itemize}

%The paper also discusses potential UAV open-development axes, including AI integration \cite{21}, environmental monitoring and conservation \cite{26,28,29}, miniaturization \cite{32}, swarming and cooperative control \cite{1,2,5,8}, and transformability systems \cite{34,35,36,37}. The aircraft control development axis is a crucial area in UAVs research, and it is essential to consider appropriate control algorithms and the most commonly used control techniques in UAVs research and provides valuable insights into the control techniques used in UAVs research \cite{15,17}. 

\section{Popular UAV Classification in Research} \label{sec2}

UAVs, also known as drones, can be classified based on several factors, such as their flying principle, mission, weight, propulsion, control, altitude range, configuration, purpose, launch method, payload, autonomy level, size, endurance, and range \cite{39,24}. Common classifications of UAVs are as follows:
\begin{itemize}
\item Flying Principle: This category includes fixed-wing, rotary-wing, hybrid, flapping-wing, and other types of UAVs that differ in their flying mechanism.
\item Mission: UAVs can be classified based on their mission, such as reconnaissance, surveillance, attack, transport, search and rescue, and more.
\item Weight: UAVs can be classified based on their weight, such as micro UAVs, small UAVs, tactical UAVs, MALE (medium altitude long endurance) UAVs, HALE (high altitude long endurance) UAVs, and more.
\item Propulsion: UAVs can be powered by electric, fuel, solar, or other sources.
\item Control: UAVs can be remotely piloted, autonomous, semi-autonomous, or have other types of control.
\item Altitude Range: UAVs can be classified based on their altitude range, such as low-altitude UAVs, high-altitude UAVs, and stratospheric UAVs.
\item Configuration: UAVs can have different configurations, such as mono-rotor, multi-rotor, tilt-rotor, tilt-wing, and others.
\item Purpose: UAVs can have different purposes, such as military, civilian, commercial, industrial, scientific, and more.
\item Launch Method: UAVs can be launched from the ground, air, sea, or have other types of launch methods.
\item Payload: UAVs can carry various payloads, such as sensors, cameras, communication systems, weapons, cargo, and others.
\item Autonomy Level: UAVs can have different levels of autonomy, such as fully autonomous, semi-autonomous, human-operated, and others.
\item Size: UAVs can have different sizes, such as mini UAVs, handheld UAVs, man-portable UAVs, vehicle-mounted UAVs, and more.
\item Endurance: UAVs can have different endurance levels, such as short endurance UAVs, long endurance UAVs, ultra-long endurance UAVs, and more.
\item Range: UAVs can have different range levels, such as short-range UAVs, intermediate-range UAVs, long-range UAVs, and more.
\end{itemize}
These classifications enable us to categorize UAVs and understand their capabilities, limitations, and potential applications. The continuous development and evolution of UAV technology have led to the creation of new classifications and the blurring of traditional boundaries between them.

%%%%%%%%%%%%%%%%%%%%%%%%%%%%%%%%%%%%%%%%%%
\section{Navigating the Latest UAV Research Chalenges} \label{sec3}

The primary objective of this review paper is to assess recent trends in UAV research over the past three years, using the Scopus database as a reliable source. The database was queried using relevant keywords such as "drone," "UAV," "unmanned aerial vehicle," and "unmanned aerial systems." The obtained results were meticulously analyzed to identify the prominent research directions within this field. The number of scientific publications associated with each research direction was employed as an indicator of its significance and influence. This comprehensive analysis provides a comprehensive overview of the current state of UAV research and highlights the most promising avenues for future investigations. By gaining valuable insights into the prevailing areas of focus within the UAV research community, we can better comprehend the potential of this technology and its profound impact across various domains.

A systematic search was performed on the Scopus database using predetermined keywords in the Title, Abstract, and Keywords fields. The search yielded a total of 47,635 references published in the UAV field between 2020 and 2023. The search was conducted on March 14, 2023. The chart below illustrates the resulting research directions, derived using the formula: (TITLE-ABS-KEY (uav) OR TITLE-ABS-KEY (drone) OR TITLE-ABS-KEY (unmanned AND aerial AND vehicle) OR TITLE-ABS-KEY (unmanned AND aerial AND systems)) AND PUBYEAR > 2019 AND PUBYEAR < 2024.

\begin{figure*}[t!]
\includegraphics[width=1.0\textwidth]{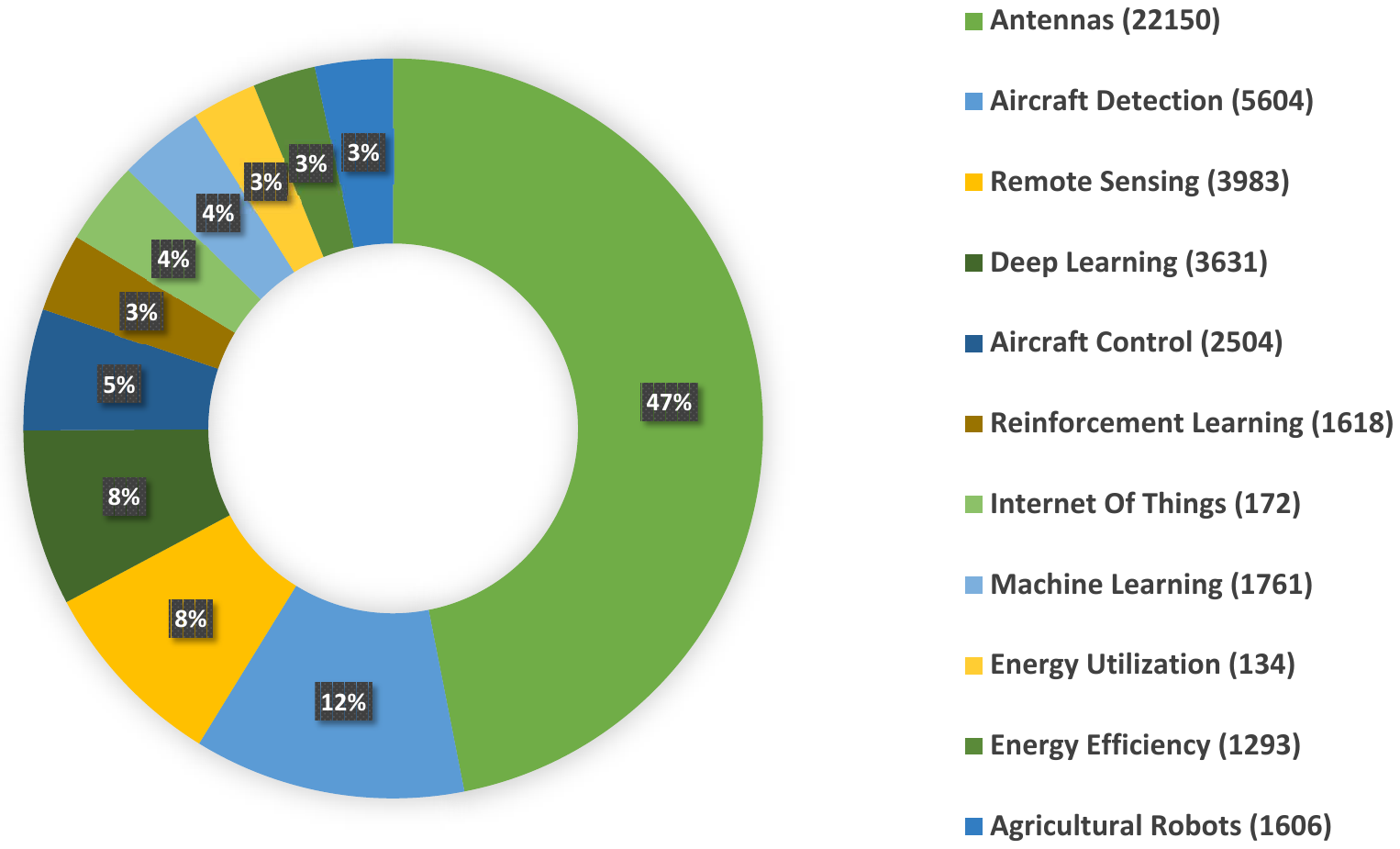}
\caption{Chart of last three years research directions in UAV filed.}
\label{fig1}
\end{figure*}   
%\unskip

Over the past three years, there has been a surge in research efforts in the field of UAVs, with various areas of study being explored. Antennas, aircraft detection, remote sensing, deep learning (DL), reinforcement learning (RL), machine learning (ML), aircraft control, the IoTs, trajectories, energy utilization, and energy efficiency have emerged as the most prominent research directions \cite{mohsan2023unmanned}. The development of the aforementioned AI tools has revolutionized the UAV field, leading to improved performance in areas such as object detection, trajectory optimization, and mission planning. Moreover, research on human-UAV interaction, swarm behaviour, environmental sensing, safety and reliability, integration with other platforms, application-specific development, and legal and ethical issues has also garnered significant attention in recent years \cite{arafat2023vision}.

\begin{figure*}[t!]
\includegraphics[width=1.0\textwidth]{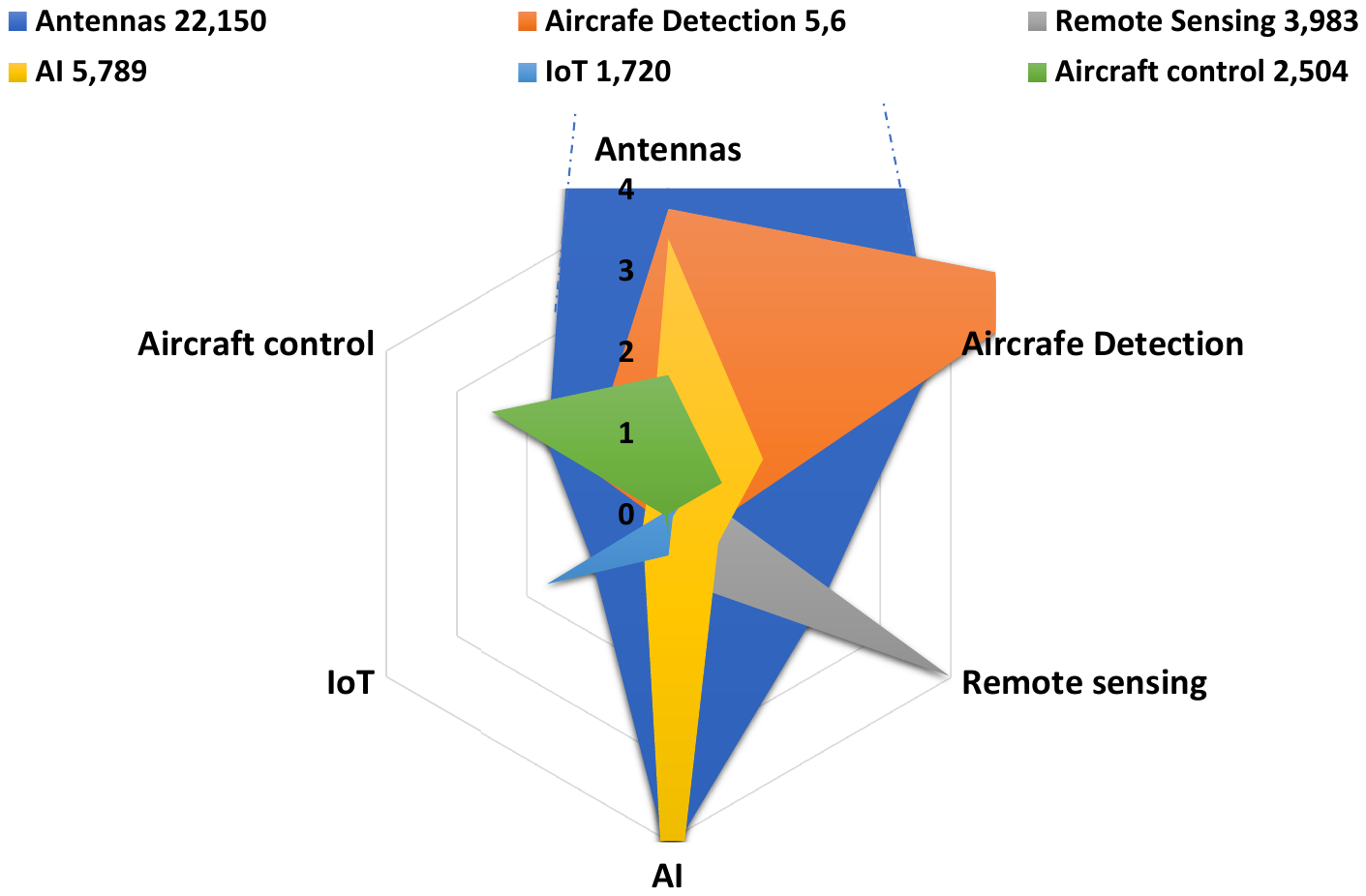}
\caption{View in UAV Research Direction and their Inner-Interaction, recent three years\label{fig2}}
\end{figure*}   
%\unskip

The research direction of antennas in the UAV field has received the most attention, with 22150 documents including journal papers, books, and conference papers, among others. This field has strong links to other research areas, such as aircraft detection, remote sensing, AI, IoT, and aircraft control. Aircraft detection is the research area that interacts the most with antennas in UAVs, with 3749 documents, followed by AI with 4231 documents, remote sensing with 2176 documents, aircraft control with 1707 documents, and IoT with 1092 documents. In the UAV field, AI is the second most referenced research direction with 5789 documents, while aircraft detection is also strong with 5604 documents. Other research areas, such as remote sensing with 3983 documents, IoT with 1702 documents, and aircraft control with 2504 documents, have also received notable attention. The interconnections between these research areas are depicted in Figure \ref{fig2} and are further elaborated upon in Table \ref{tab1}. These results suggest that there is substantial overlap between research areas in UAV technology, which could lead to more integrated and efficient solutions in the future. The data was collected on March 14, 2023.

\begin{table*}[t!] 
\centering
\caption{Charting the Course of UAV Research: Exploring Emerging Directions and Inter-Interaction. \label{tab1}}
\begin{tabular}{|c|c|c|c|c|c|c|}
\hline
\textbf{Research Direction} & \textbf{Antennas} & \textbf{Aircraft Detection} & \textbf{Remote Sensing} & \textbf{AI} & \textbf{IoT} & \textbf{Aircraft Control}\\
\hline
\textbf{Antennas}           & 2,215  & 3,749  & 2,176  & 3,38   & 1,092 & 1,707\\
\textbf{Aircraft Detection} & 3,749  & 5,604  & 0,462  & 1,343  & 0,133 & 0,758\\
\textbf{Remote Sensing}     & 2,176  & 0,462  & 3,983  & 0,707  & 0,063 & 0\\
\textbf{AI}                 & 4,231  & 2,203  & 0,777  & 5,789  & 0,512 & 0,294\\
\textbf{IoT}                & 1,092  & 0,133  & 0,063  & 0,365  & 1,72  & 0,043\\
\textbf{Aircraft Control}   & 1,707  & 1,152  & 0      & 0,311  & 0,043 & 2,504\\
\hline
\end{tabular}
\end{table*}

\subsection{Communication and Antennas }
The transmission and reception of signals are essential for the operation of UAVs, making antennas a critical component in their design. For UAV applications like communication \cite{42,43,44,45}, antennas need to be lightweight, compact, and durable enough to withstand harsh environmental conditions. Recent research in UAV antennas \cite{42,43,44,45,53,55,56,58} has focused on developing advanced technologies to enhance UAV performance. Researchers are exploring ML and DL approaches for antenna design and optimization \cite{55,59} as well as designing high-gain, wideband, and multibeam antennas and integrating them with other subsystems like power and control systems \cite{45}. Table \ref{tab2} portrays a comparison between the various communication technologies for FANETs.

\begin{table*}[t!] 
\caption{Comparison between the various communication technologies for FANETs \cite{SHI2021104340,hy}.} 
\label{tab2}

\begin{tabular}{ m{4cm} | m{3.6cm}| m{4cm}| m{4cm} }
\hline
\textbf{Technology}  & \textbf{Standard} & \textbf{Data Rate} & \textbf{Range}        \\ \hline
\textbf{WiFi} \cite{550,560,580,590,600} & 802.11    \cite{1}          & Up to 2 Mbps   & Up to 100m    \\ 
LCY<5ms| DM:Y |      & 802.11a \cite{560}          & Up to 54 Mbps  & Up to 120m   \\ 
UL NT:WLAN      & 802.11b  \cite{550}         & Up to 11 Mbps  & Up to 140m    \\ 
                       & 802.11n \cite{580}         & Up to 600 Mbps & Up to 250m    \\ 
                    & 802.11g \cite{590}         & Up to 54 Mbps  & Up to 140m    \\ 
                     & 802.11ac \cite{600}        & Up to 866.7Mbps & Up to 120m   \\ \hline
\textbf{ZigBee} \cite{zigbee,santoszigbee}                  & 802.15.4         & Up to 25kbps  & Up to 100m    \\ 
LCY<15ms|DM:Y|UL  & & & \\

\hline
\textbf{blacktooth V5}  \cite{2022comparative}           & 802.15.1         & Up to 2Mbps   & Up to 200m    \\ 
LCY<3ms|DM:Y|UL & & & \\
\hline
\textbf{LoRaWAN} \cite{paredes2023lora}     & IEEE    & Up to 50 kbps  & Up to 15km \\ 
DCD|DM:Y|UL & 802.15.4g & & \\
NT:WPAN   & & & \\
\hline
\textbf{Sigfox}  \cite{noor2020review}                 & -                & Up to 100 bps  & Up to 30km   \\ 
LCY about 2s|DM:Y|UL & & & \\

\hline
\textbf{NB-IoT}   \cite{martinez2019iot}                & -                & Up to 250 kbps & Up to 35km    \\ 
LCY:1.6 to 10s|DM:Y|L & & & \\
 
\hline

\textbf{Cellular}                       &               &     &      \\ 
 \textbf{3G}   \cite{500,510,520}                     & HSPA+           & Up to 21.1 Mbps   & Wide Area    \\
LTE / LTEM                     &              & Up to 100 Mbps   &      \\

LCY:500ms NT:LPWAN & & & \\
%\hline
\textbf{4G} \cite{noor2020review}    LCY:4ms                & HSPA+            & Up to 100 Mbps   & Wide Area   \\
%\hline
\textbf{5G}    \cite{noor2020review,20,371,colajanni2022service}                   & mMTC             & Up to 1 Gbps   & Wide Area    \\ 
LCY:1ms |DM:Y|L                       & URLLC            & Up to 1 Gbps   & Wide Area    \\
 %\hline
\textbf{B5G}  \cite{amponis2022drones}      & eMBB/Hybrid             & Up to 100 Gbps & Wide Area    \\ 
LCY:1ms |DM:Y|L          &  URLLC      &             Up to 100 Gbps      &     Wide Area    \\ 
\textbf{6G}  \cite{grasso2022tailoring,amponis2022drones}                       & MBRLLC           & Up to 1 Tbps   & Wide Area    \\ 
LCY< 1ms|DM:Y|L                     & mURLLC           & Up to 1 Tbps   & Wide Area    \\ 
                       & HCS / MPS             & Up to 1 Tbps   & Wide Area    \\ 
\hline
\end{tabular}
\end{table*}

Where:
\textbf{DM}:Device Mobility
\textbf{LCY}:Latency
\textbf{Y}:YES
\textbf{DCD}:Device Class Dependent 
\textbf{ST}:Spectrum Type
\textbf{UL}:Unlicensed
\textbf{L}:licensed
\textbf{NT}:Network type\\

 Miniaturization of antennas and AI is also being studied to develop smaller and more agile UAVs with enhanced capabilities \cite{55}. Further, research on novel materials and manufacturing techniques such as 3D printing \cite{58,60,61,62} has great potential to produce efficient and low-cost antennas for UAVs. Overall, the advancement of UAV technology and the development of more efficient and effective UAV systems for various applications depend on continued research into UAV antennas.
 
Flying ad-hoc networks (FANETs) are wireless communication networks composed of drones. FANETs enable communication and collaboration among drones in a decentralized manner, without relying on a fixed infrastructure. FANETs are designed to operate in the sky, and they offer advantages such as improved coverage, increased mobility, and access to remote or inaccessible areas. FANETs utilize wireless communication technologies and specialized protocols to establish and maintain connections between drones, facilitating data and message exchange \cite{noor2020review}. Through reviewing extensive literature, we have extracted the information presented in this Table \ref{tab2}. It provides a comprehensive comparison of different communication technologies utilized in FANETs.

\subsection{IoTs }
The integration of UAVs with IoTs has opened up new possibilities for data collection, analysis, and communication in various fields. By combining UAVs and IoT, a network of connected devices, sensors, and UAVs can collect, process, and share data in real-time. Recent research \cite{64,65,67} on IoT-enabled UAVs has focused on developing efficient and scalable communication protocols, network architectures \cite{68}, and data processing algorithms to enable seamless integration and interoperability between UAVs and other IoT devices like sensors and data centres.
IoT-enabled UAVs offer numerous benefits, such as improving the efficiency and effectiveness of various applications \cite{72} and disaster management. For example, UAVs equipped with sensors can collect data on crop health, soil moisture, and temperature, which can be analyzed in real-time to inform irrigation and fertilization decisions. Similarly, UAVs can be used to monitor natural disasters and assess damage, enabling a more rapid and accurate response.
Despite the immense potential of IoT-enabled UAVs, there are significant challenges that need to be addressed. Ensuring the security and privacy of IoT-enabled UAVs is crucial, and developing effective mechanisms for data processing and analysis is essential. However, the integration of UAVs with IoT is a promising area of research that has the potential to revolutionize various fields and enable new applications \cite{74}.

\subsection{Aircraft Detection }

Detecting and avoiding collisions with manned aircraft is a crucial task in the operation of UAVs, especially in shared airspace. The ability to detect aircraft is essential for ensuring the safe operation of both manned and unmanned aircraft. Recent research in the area of aircraft detection for UAVs has been focused on developing advanced systems that can detect and track aircraft in real-time using a range of sensors, including radar, LIDAR, and optical cameras. However, detecting small aircraft such as general aviation aircraft using traditional radar systems can be challenging. To address this issue, researchers are exploring the use of ML algorithms, such as DL, to improve the accuracy and reliability of aircraft detection systems.
Integrating aircraft detection systems with UAVs' navigation and control systems is also a significant research direction. This integration can enable automatic adjustment of UAVs' flight paths in response to the detected aircraft, ensuring the safe operation of both manned and unmanned aircraft. The research on aircraft detection for UAVs is crucial for enabling the widespread adoption of UAV technology in various domains, such as delivery, inspection, and surveillance, while ensuring safe operation in shared airspace.

\subsection{Control and Autonomous Flight}
Autonomous flight refers to the ability of UAVs to operate without human intervention, achieved through the use of advanced control algorithms \cite{5,17} and navigation systems that allow UAVs to fly, navigate, and perform tasks autonomously \cite{5}. Autonomous flight is a complex and challenging area of research in the field of UAVs, requiring the integration of multiple technologies, such as sensors \cite{88}, computer vision \cite{94}, and AI \cite{89}. The objective is to develop UAVs that can perform complex tasks in a safe and efficient manner, such as precision landing \cite{90, 94}, without human intervention. One of the critical challenges in autonomous flight is developing UAVs that can navigate and avoid obstacles in real-time while maintaining stability and control. This requires the development of advanced control algorithms \cite{91} and sensors \cite{88} that can accurately detect and respond to changes in the environment. Another significant challenge in autonomous flight is ensuring the safety and reliability of UAVs, particularly in scenarios with limited human intervention or hazardous areas \cite{93}. To address these challenges, researchers are developing new approaches for monitoring, controlling, and diagnosing UAVs, including the integration of backup systems, failsafe mechanisms, and real-time monitoring systems. The goal of autonomous flight is to develop UAVs capable of performing a broad range of tasks safely, efficiently, and reliably without the need for human intervention. This has the potential to revolutionize various industries, including agriculture, logistics, military operations, search and rescue missions, and civil engineering applications.

\subsection{Perception and Sensing}
Perception and sensing are crucial capabilities of UAVs, allowing them to gather, process, and interpret information from their surroundings. These capabilities are essential for enabling UAVs to perform various tasks, such as navigation, mapping, inspection, and surveillance. However, these areas of research are complex and challenging, requiring the integration of multiple technologies, including sensors, computer vision, and AI.
The ultimate aim is to develop UAVs that can perceive and understand their environment and make informed decisions based on that information. One of the key components of perception and sensing is the integration of various sensors, such as cameras, LiDAR, and radar \cite{99,100}. These sensors provide UAVs with information about the environment, including the position and location of obstacles, terrain, and other objects.
Another crucial aspect of perception and sensing is the development of computer vision algorithms that can process and interpret the information collected by the sensors. This includes identifying objects, recognizing patterns, and tracking movement. Additionally, researchers are exploring the integration of AI techniques, such as ML and DL \cite{21}, to enable UAVs to learn from their experiences and improve their perception and sensing capabilities over time.
The ultimate goal of perception and sensing in UAVs is to develop systems that can accurately perceive and understand their environment and make informed decisions based on that information. This has the potential to revolutionize various industries, including agriculture \cite{104}, construction, civil applications \cite{105}, marine applications \cite{27}, mining \cite{107}, military operations, and search and rescue missions, such as wildfire remote sensing \cite{108}. Furthermore, researchers are exploring the potential of cooperative perception using multiple UAVs to enhance their capabilities \cite{109}.

\subsection{Energy-Efficient Flight   }
Energy-efficient flight is a critical area of research in the field of UAVs, aiming to develop drones that can fly for extended periods while consuming minimal energy. Achieving energy efficiency is essential to enhance the performance and capabilities of UAVs, including flight time, payload capacity, and range. Researchers are exploring several approaches \cite{110}, such as aerodynamic design optimization \cite{110}, lightweight materials, and integration of alternative energy sources such as solar power \cite{112}.
Reducing the weight of UAVs is one of the key challenges in achieving energy-efficient flight. To address this, researchers are exploring the use of lightweight materials such as composites and new manufacturing techniques that can reduce the weight of UAVs. Moreover, the optimization of the propulsion system \cite{124,115} is critical in achieving energy efficiency, including the use of more efficient engines and the development of new propulsion technologies \cite{124}. Integrating electric and hybrid propulsion systems that offer improved energy efficiency compared to traditional internal combustion engines is also under research.
Another area of research is the integration of alternative energy sources, such as solar power \cite{116,117}. This involves developing new lightweight solar panels and energy storage systems capable of providing power for extended periods. Energy-efficient flight has the potential to significantly improve UAVs' capabilities and performance, enabling new applications and uses. However, massive data transfer during communication and surveillance can result in delays and considerable energy usage. Thus, deep reinforcement learning (DRL) and other AI approaches have been used in this context \cite{119,120,121}.

\subsection{Human-UAV Interaction}
Human-UAV interaction is an emerging field of research that investigates the interaction between humans and UAVs across various contexts, such as entertainment, education, and research. It has the potential to revolutionize several industries. Human-UAV interaction involves developing new technologies and interfaces that enable intuitive and innovative ways for humans to interact with UAVs \cite{125}. These technologies include virtual and augmented reality, gestures \cite{128}, and other forms of human-machine interaction.
One of the major challenges in human-UAV interaction is developing UAVs that can respond to human input in real-time while maintaining stability and control. This requires integrating advanced control algorithms and sensors and developing new user-friendly human-machine interfaces. Another challenge in human-UAV interaction is ensuring the safety and reliability of UAVs, particularly in scenarios where there is limited human intervention.
Numerous approaches have been proposed for controlling UAVs using natural language, hand gestures, and physical movements. Intelligent human-UAV interaction systems have been developed, utilizing ML and DL techniques to recognize gestures and enable efficient control of the UAV \cite{126,127,128,129}. Furthermore, some research papers have proposed novel architectures that allow users to control the UAV using natural body movements \cite{130,131}. Alongside technical approaches, several studies have examined human factors and challenges related to the use of UAVs, such as user interfaces, training, and workload \cite{134}.
Another active area of research in human-UAV interaction is understanding human decision-making when controlling UAVs, particularly in search and rescue applications \cite{135}. This highlights the importance of developing intuitive and efficient human-UAV interaction systems that can be used across various domains and applications. Survey papers have reviewed the current state-of-the-art in human-UAV interaction, such as a scoping review identifying areas like entertainment, transportation, and public safety, and another survey providing an overview of various control interfaces, gesture recognition techniques, and autonomous operation methods \cite{133}.

Furthermore, the literature has explored the human factors and challenges associated with using UAVs, including issues related to user interfaces, training, and workload \cite{134}. Understanding human decision-making when controlling UAVs is an active area of research, particularly in search and rescue applications \cite{135}. This highlights the importance of developing intuitive and efficient human-UAV interaction systems that can be used in a wide range of domains and applications.

\subsection{Swarm behaviour}

Swarm behaviour is a research area within the field of UAVs that aims to study the collective behaviour of groups, or "swarms," of UAVs \cite{136,150}. The potential impact of swarm behaviour on various industries, such as military operations, search and rescue missions, and environmental monitoring, has driven the development of algorithms and control systems that enable UAVs to coordinate their actions and work together to achieve a common goal.
One of the challenges in swarm behaviour is to ensure effective collaboration between UAVs while adapting to changing conditions and environments. To address this challenge, researchers are developing new algorithms for cooperation and coordination, as well as new approaches for task allocation and resource management \cite{17}. Additionally, the development of algorithms that enable UAVs to operate autonomously and make decisions based on their environment, including the integration of AI techniques such as ML and DL, is an important aspect of swarm behaviour \cite{22}.
Communication and control architectures are essential for the successful operation of UAV swarms \cite{140}. A review of UAV swarm communication and control architectures \cite{17} highlights the need for scalable and flexible architectures to support different swarm configurations and tasks. Similarly, a review of UAV swarm communication architectures \cite{140} discusses the challenges and future directions for communication in UAV swarms. Examples of communication and control architectures that can be used in UAV swarms include high-level control of UAV swarms with RSSI-based position estimation \cite{141} and a self-coordination algorithm (SCA) for multi-UAV systems using fair scheduling queue \cite{151}.

Path planning is another critical aspect of swarm behaviour. A recent review of AI applied to path planning in UAV swarms \cite{22} discusses the latest developments in this field. The importance of motion planning in swarm behaviour is highlighted in another paper that examines the motion planning of UAV swarms, recent challenges, and approaches \cite{143}. Additionally, a study on collaborative UAV swarms towards coordination and control mechanisms \cite{144} proposes a method for collaborative motion planning of UAV swarms. Another important issue addressed in swarm behaviour is localization in UAV swarms, which is tackled by high-level control of UAV swarms with RSSI-based position estimation \cite{141}.
Finally, there are various applications of swarm behaviour, including continuous patrolling in uncertain environments using the UAV swarm \cite{148} and autonomous drone swarm navigation and multi-target tracking in 3D environments with dynamic obstacles \cite{149}. In summary, research on swarm behaviour for UAVs has been a rapidly growing area of study, with numerous advances made in recent years. The development of algorithms and control systems for cooperation, coordination, task allocation, and resource management has been a primary focus of research, as has the integration of AI techniques. Communication and control architectures, path planning, localization, and resiliency are also vital areas of research. The numerous applications of swarm behaviour demonstrate the significant potential impact of this technology on various industries.

\subsection{AI}
The use of AI has had a significant impact on the field of UAVs in recent years. By enabling UAVs to perform tasks autonomously, AI has made them more efficient and effective in a variety of applications. Some of the most significant contributions and works related to using AI in UAVs include object detection \cite{77, 79, 82, 83, 84} and tracking, path planning \cite{22}, autonomous navigation \cite{89, 211}, swarm intelligence \cite{136, 22}, image and video analysis \cite{82,84} and cybersecurity \cite{himeur2022latest}.

AI algorithms, such as Convolutional Neural Networks (CNNs) and Recurrent Neural Networks (RNNs), have been used to detect and track aircraft \cite{82}, objects \cite{AI2, AI3, AI4}, and real-time objects from UAVs \cite{AI1,elharrouss2021panoptic}, with applications in surveillance \cite{121}, search and rescue \cite{AI1}, and agriculture \cite{atalla2023iot}. AI also enables UAVs to navigate autonomously in complex environments, avoid obstacles, and make real-time decisions based on sensor data \cite{99}, with applications in delivery, inspection, and surveillance \cite{himeur2023video}. Additionally, AI algorithms have been used to coordinate the behaviour of multiple UAVs, forming a swarm that can perform tasks collaboratively \cite{22, 136}, with applications in search and rescue, surveillance, and military operations.

Moreover, AI algorithms have been utilized to analyze images and videos captured by UAVs, extracting valuable information such as object recognition, semantic segmentation, and anomaly detection. These capabilities have a broad range of applications in fields such as agriculture, environmental monitoring, and disaster management. Additionally, AI has been used to detect and prevent cyberattacks on UAVs, ensuring their safety and security, with applications in military operations and commercial UAVs.
Some notable works related to using AI in UAVs include recent examples that have received attention from the research community. These include DRL for UAV control, vision-based autonomous landing of a fixed-wing UAV \cite{94}, swarm intelligence for collaborative UAV mission planning \cite{136, 22, 212}, autonomous navigation of UAVs in indoor environments \cite{89}, and AI-based object detection and tracking for UAV surveillance \cite{AI1, AI3, 121}. These examples demonstrate the wide range of applications for AI in UAVs, including navigation, mission planning, object detection and tracking, and surveillance.

\section{UAV Active and Expanding Research Directions} \label{sec4}
To conduct a comprehensive analysis, the Scopus database has been used as the primary data source for extracting pertinent information regarding the research direction of UAVs over the past three years. The goal of this analysis is to (i) identify the main open challenges, and (ii) provide insights into the most active and rapidly expanding research directions in the UAV field over the last three years. To achieve this objective, a systematic search of the Scopus database was conducted using predefined keywords in the title, abstract, and keywords fields from January 2020 to December 2022. Research directions were delineated based on the identified publications, and the resultant growth trajectories were plotted, taking into account the linear relationship between the number of publications and time (y=ax+b), where y represents the number of publications, x represents the year, and b represents the number of publications in the previous year. The parameter a reflects the magnitude and rate of growth observed from year to year.

\begin{figure*}[t!]
\centering
%\raggedleft 
\includegraphics[width=1.0\textwidth]{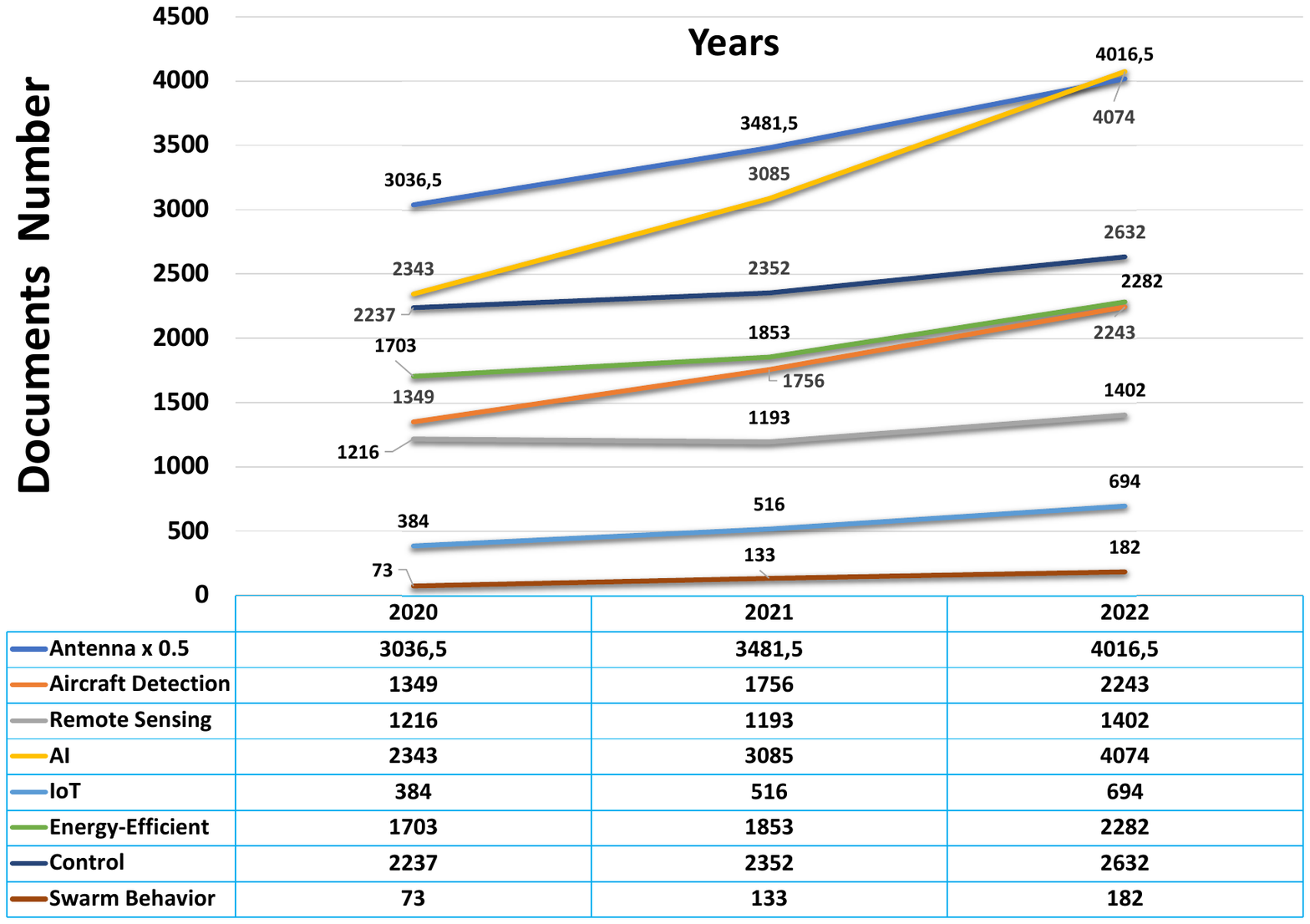}
\caption{UAV Research Direction growth trajectories last three years view.\label{fig3}}
\end{figure*}

The results of this analysis are presented in Figure \ref{fig3}, depicting the growth trajectories of each research direction. Furthermore, Table \ref{table2} tabulates the average growth rate for each research direction to facilitate a better understanding of the growth dynamics in the UAV field.

\begin{table*}[t!] 
\centering
\caption{UAV Research Direction, Acceleration of Growth.}
\label{table2}
\begin{tabular}{|c|c|c|}
\hline
\textbf{Research Direction} & \textbf{Rate of Growth/Year (Number of Document/Year)} & \textbf{Acceleration of Growth} \\
\hline
Antenna & 980 & 1.20 \\
\hline
Aircraft Detection & 447 & 1.19 \\
\hline
Remote Sensing & 93 & 9.08 \\
\hline
AI & 865 & 1.33 \\
\hline
IoT & 155 & 1.34 \\
\hline
Energy-Efficient & 289 & 2.86 \\
\hline
Control & 197 & 2.43 \\
\hline
Swarm & 54 & 0.81 \\
\hline
\end{tabular}
\end{table*}

Table \ref{table2}'s Acceleration of Growth demonstrates a high degree of efficacy in predicting the areas of interest that attract significant attention from researchers in the UAV domain. These predictions offer a nuanced perspective of the developments and trends within the field. For example, the Antenna field has a considerable number of publications and an impressive growth rate of approximately 980 new documents per year. However, its growth ratio remains relatively constant at 1.20. In contrast, the field of Remote Sensing experiences a remarkable surge in interest each year, reflected in an acceleration of growth ratio of 9.08. Consequently, these findings provide valuable insights into the most promising research directions within the UAV domain.

\section{Potential UAV's Open Research Directions} \label{sec5}

The open development axis for UAVs involves creating a collaborative ecosystem where developers, researchers, and users can work together to build open-source platforms, tools, and standards for UAV design, development, and operation. This approach allows for greater innovation and flexibility in the UAV industry \cite{7}, including the integration of AI and ML algorithms to enhance autonomous flight and decision-making capabilities \cite{208,209}.
Further challenges and open research directions can be elaborated on in the following points:
\subsection{Integration of AI}  
AI algorithms are revolutionizing the way UAVs operate. By using these algorithms, UAVs can achieve improved autonomous flight and decision-making capabilities, which can result in more efficient and safer operations \cite{208,209}.
One way AI and ML algorithms can be used in UAVs is through the development of advanced sensor systems. For example, computer vision and LIDAR can be integrated into UAVs to provide real-time data to the AI system \cite{211}. This allows the UAV to make decisions based on its environment and react to changes dynamically. Computer vision can enable the UAV to recognize objects and people in its environment, which can be used to improve safety and prevent collisions. Meanwhile, LIDAR can provide detailed information about the UAV's surroundings, including distance, size, and speed of objects, which can be used to navigate complex environments more effectively.

Another way AI can be used in UAVs is through the optimization of flight paths. By using ML algorithms, UAVs can learn from past flights and optimize their routes to reduce energy consumption and increase efficiency. This can be achieved by analyzing data such as wind speed, temperature, and other environmental factors \cite{212}.

\subsubsection{Generative AI and ChatGPT for UAVs} 
Natural Language Processing (NLP) models like Chat Generative Pre-trained Transformer (ChatGPT) \cite{OpenAI,sohail2023using} are designed to understand natural language input from users and generate human-like responses. ChatGPT can be adapted to many robotics tasks, such as high-level agent planning \cite{chatgpt3} or code generation \cite{chatgpt4,sohail2023future}. As such, it can serve as an intuitive language-based interface between non-technical users and UAVs.

"Efforts to incorporate language into robotics systems have largely focused on using language token embedding models,multi-modal model features, and  LLM features for specific form factors or scenarios. Applications range from visual-language navigation \cite{chatgpt6}, language-based human-robot interaction \cite{chatgpt7}, and visual-language manipulation control." \cite{vemprala2023chatgpt} 
ChatGPT, for example, can be used via API libraries to enable many tasks \cite{chatgpt2}, such as zero-shot task planning in drones, where it accesses functions that control a real drone and serves as an interface between the user and the drone \cite{chatgpt1}. This can allow non-technical users to easily and safely operate UAVs without needing specialized training.

A real drone was operated using ChatGPT through a separate API implementation, which offered a user-friendly natural language interface between the user and the robot, allowing the model to create intricate code structures for drone movement such as circular and lawnmower inspections \cite{GPT-Drone2}. Using the Microsoft AirSim \cite{chatgpt1, chatgpt2} simulator, ChatGPT has also been applied to a simulated domain, where the possibility of a model being used by a non-technical user to operate a drone and carry out an industrial inspection scenario has been investigated \cite{GPT-Drone2}. It can be seen from the snippet that ChatGPT can accurately control the drone by reading user input for geometrical clues and purpose.

\subsection{ Environmental Monitoring and Conservation} 
The utilization of UAVs for environmental monitoring and conservation presents a promising application of this advanced technology \cite{26}. These unmanned aerial vehicles, equipped with high-resolution cameras and other cutting-edge sensors, offer immense potential for a wide range of purposes, including monitoring wildlife populations \cite{28}, tracking changes in ecosystems \cite{tiwary2021monitoring}, and detecting environmental hazards \cite{}. By leveraging the capabilities of UAVs in these areas, we can greatly enhance our capacity to collect precise and reliable data, thereby gaining deeper insights into the overall health and well-being of our planet \cite{31}. Such invaluable information empowers us to develop and implement effective conservation strategies, ensuring the preservation and safeguarding of our environment for future generations \cite{himeur2022ai}.  \textcolor{black}{The use of drones in ecological and glaciological research in regions like Antarctica is on the rise, as demonstrated by studies \cite{fudala2022use,zmarz2018application,wojcik2019investigation}. Drones facilitate detailed geomorphological mapping, precise vegetation monitoring over expansive areas, and health indicator assessments. They enhance the identification and characterization of cryospheric features, including subsurface applications, and revolutionize faunal studies by enabling non-invasive counting and morphometrics of diverse animal species \cite{fudala2022use}. UAV atmospheric surveys allow swift and versatile data collection, including aerosol sample collection. The design and development of platforms tailored to the harsh Antarctic environment have been crucial for the success of these applications. UAVs capable of collecting physical samples from remote or inaccessible areas are available, and further advances in autonomy and robustness will enhance their utility for Antarctic fieldwork \cite{pina2022uavs}. UAV usage for environmental monitoring and conservation serves both planetary and human interests.}

\subsection{Urban Air Mobility (UAM)}
UAM, an emerging field, holds immense promise for the future of transportation \cite{220,221,222}. UAVs have the potential to revolutionize urban transportation by offering a rapid, efficient, and eco-friendly alternative to traditional ground-based systems \cite{220}. However, realizing this potential necessitates significant technological advancements in navigation, autonomous flight, and safety systems. Moreover, the development of tailored air traffic management systems designed to address the unique challenges of urban environments is crucial for ensuring the safe and efficient operation of UAVs in densely populated areas \cite{223}. With ongoing growth and investment in this domain, UAVs are poised to become a pivotal component of future urban transportation systems.

\subsection{Miniaturization} 
In the UAV industry, miniaturization is a prominent trend that focuses on developing smaller and more compact drones capable of diverse applications \cite{33}. These applications encompass search and rescue, delivery services, surveillance, and more. Nonetheless, accomplishing miniaturization necessitates substantial technological advancements, including the development of compact and efficient propulsion systems, as well as lighter and more durable materials. Consequently, miniaturization has emerged as a key area of research and development within the UAV industry, unlocking new possibilities for drone utilization across a wide range of fields.

\subsection{Swarming and Cooperative Control} 
The open development axis in the UAV industry focuses on establishing a collaborative ecosystem that brings together developers, researchers, and users to advance algorithms and techniques for swarming and cooperative control of multiple UAVs, enabling them to perform complex tasks \cite{136,17}. These tasks encompass a wide range of applications, including surveillance, search and rescue operations, and environmental monitoring. To achieve this, significant technological advancements are necessary, such as the development of robust communication protocols \cite{17,140}, distributed sensing and control systems \cite{144}, and adaptive decision-making capabilities. For instance, the design of effective communication protocols facilitates seamless information exchange and coordination among multiple UAVs, enabling them to work together efficiently towards common objectives. Similarly, distributed sensing and control mechanisms empower each UAV to carry out specific tasks within a coordinated framework, greatly enhancing the efficiency and effectiveness of complex missions performed by UAV swarms. Additionally, the implementation of adaptive decision-making algorithms equips UAVs with the ability to make rapid and accurate decisions based on real-time data, further augmenting the capabilities of UAV swarms in various scenarios.

\subsection{Beyond Visual Line of Sight (BVLOS) Operations}
Beyond Visual Line of Sight (BVLOS) Operations refers to the technologies that enable UAVs to operate beyond the visual line of sight of their pilot \cite{232}. Achieving BVLOS capabilities requires the development of advanced sense-and-avoid systems capable of detecting obstacles and avoiding collisions \cite{233}. Additionally, reliable communication and control protocols are necessary to ensure safe and efficient operations. To enable the widespread adoption of BVLOS operations, regulatory frameworks must be established to ensure compliance with safety standards and mitigate potential risks \cite{236}.

\subsection{Long-Range and High-Altitude Flights}
Another area of development in the UAV industry is the advancement of long-range and high-altitude flights. This entails equipping UAVs with the ability to fly for extended periods and at greater altitudes. To achieve this, there is a need for the development of more energy-efficient propulsion systems capable of sustaining long flights \cite{240}. Additionally, the integration of renewable energy sources, such as solar panels, into UAV designs is being explored to extend their range and increase their endurance \cite{241,242}.

\subsection{Flight Safety}
As the use of UAVs continues to grow across different applications, ensuring flight safety has become a crucial concern. In response, developers are actively working on integrating new technologies that can enhance the safety of UAV operations. For instance, there has been an increasing focus on developing collision avoidance systems \cite{243,245} that can prevent mid-air collisions with other UAVs, manned aircraft, or obstacles in the environment. Additionally, automatic landing systems \cite{246} can help to reduce the risk of accidents during landing, while onboard obstacle detection and avoidance systems \cite{243} can enable UAVs to detect and avoid obstacles during flight, reducing the risk of collisions. Such technologies are critical for ensuring the safe and responsible use of UAVs, as they can mitigate potential risks and prevent accidents.

\subsection{UAV Suspension Payload Capabilities}
UAV suspension payload refers to the development and optimization of suspension systems for UAVs that are capable of carrying various types of payloads, including heavier items such as medical supplies, food, and other essential goods. The suspension system plays a critical role in ensuring stable flight during missions that involve payload dropping, as it helps to mitigate vibrations and provide shock absorption to protect the payload and sensitive equipment on board \cite{253}.

Recent advancements in drone suspension payload technology have focused on improving the performance and efficiency of suspension systems, as well as integrating them with other components of the drone. Some examples of these advancements include the use of advanced materials and manufacturing techniques, the development of active suspension systems that can adjust to changing flight conditions in real time, and the integration of suspension systems with propulsion, control, and payload systems to ensure seamless operation and maximum efficiency.

Moreover, recent advances in controlling quadrotors with suspended loads have focused on developing new algorithms and control strategies that can handle the additional complexity and challenges introduced by the suspended payload \cite{312,313,314}. Some recent studies have proposed methods to improve the accuracy and stability of quadrotors with suspended loads, including predictive control strategies and the use of adaptive learning algorithms \cite{312,313,314}. These innovations in UAV suspension payload technology will lead to more efficient and reliable delivery and transport capabilities, further expanding the applications of UAVs in various fields.

\subsection{Transformability or Convertibility} 
Transformability or Convertibility is an emerging technology in the field of unmanned aerial vehicles that enables them to change their shape or configuration in flight \cite{35,36,256,34}. This advancement has the potential to enhance the versatility and efficiency of UAVs by allowing them to adapt to different operational environments and missions. There are several approaches to achieving transformability in UAVs, including:

One important area of transformability that we will explore is the utilization of morphing wings. This innovative approach involves designing wings capable of changing their shape during flight to enhance efficiency and manoeuvrability \cite{259}. By incorporating morphing wings technology, drones can adapt their wing configurations to varying flight conditions, such as alterations in altitude, speed, and wind direction. Through these adaptable wings, drones can optimize their aerodynamic performance and overall efficiency, thereby improving their range, endurance, and stability \cite{261}.
There are several mechanisms employed to achieve morphing wings, including shape memory alloys, smart materials, and mechanical systems. These mechanisms enable drones to adjust the wing angle, alter the curvature of the airfoil, or even completely change the wing shape.
One notable example of morphing wings technology is the "RoboSwift", developed by the Delft University of Technology in the Netherlands. Resembling a swift bird in nature, this small drone has the ability to morph its wings during flight, allowing for enhanced efficiency and reduced noise. The RoboSwift has gained considerable recognition in the scientific community due to its innovative morphing wings technology and its potential applications in various fields, such as surveillance, environmental monitoring, and wildlife research. Its remarkable features have been highlighted in numerous research papers \cite{263,264}.
Another notable example is the "FlexFoil" developed by FlexSys Inc., an American engineering firm. The FlexFoil incorporates a unique "morphing trailing edge" technology, enabling the rear edge of the wing to bend and twist in response to changes in the airflow \cite{267}. This design feature enhances the drone's aerodynamic performance and adaptability to different flight conditions, resulting in improved efficiency.
By harnessing the power of morphing wings technology, drones can revolutionize the field of aviation by achieving greater agility, range, and stability. The development of such transformative capabilities opens up new possibilities for various industries, from surveillance and monitoring to research and exploration.

The concept of foldable Unmanned Aerial Vehicles (UAVs), equipped with collapsible arms or wings, presents an intriguing area for exploration \cite{frigioescu2023preliminary}. This design feature leads to a decrease in the overall dimensions of the drones, thereby enhancing portability and facilitating more streamlined transportation and storage. Such foldable drones, including models like the Mavic Pro, DJI Mavic Air 2, Parrot Anafi, PowerVision PowerEgg X, and Robotics EVO, have gained substantial popularity due to their adaptability and convenience \cite{gokbel2023improvement}. The ability of these drones to easily fold their arms or wings offers flexibility, allowing users to transport them in compact cases or bags. This feature not only improves portability, but also boosts the drones' durability and protection during transportation. Consequently, the potential damage is minimized, ensuring the drones are well-protected and ready for operation in a variety of environments and scenarios \cite{defrangesco2022big}.

Moving on, \textbf{Reconfigurable airframes} represents another avenue of transformation in UAVs. With reconfigurable airframes, drones have the ability to change their shape or configuration during flight to adapt to different missions or operational environments. This versatility can involve modifying their wing configuration, adding or removing payloads, or adjusting their centre of gravity. By incorporating reconfigurable airframes, drones can cater to a wide range of mission requirements, making them more cost-effective and capable compared to traditional fixed-design drones.
While reconfigurable airframes in UAVs are still an evolving technology, there are a few noteworthy examples of companies and organizations that are actively developing such drones \cite{268}. For instance, roboticists from the University of Zurich and EPFL have developed quadrotors that feature foldable designs, allowing them to morph their shape in mid-air between “X” and “O” configurations \cite{272,268}. These innovative designs demonstrate the potential of reconfigurable airframe drones, showcasing their adaptability and agility in various flight scenarios.

Besides, a significant development in UAV technology is the incorporation of \textbf{variable pitch propellers} \cite{273}. These propellers are equipped with blades that can adjust their angle or pitch during flight. Variable pitch propellers, also known as adjustable or controllable pitch propellers, provide a higher level of control over the drone's flight and performance, particularly in challenging or dynamic conditions.
By altering the pitch of the propeller blades, the drone can finely tune its thrust and lift, enabling it to maintain stable flight even in varying wind conditions, altitudes, or flight modes. This capability greatly enhances the drone's manoeuvrability, efficiency, and overall performance across a wide range of applications, including aerial surveying, mapping, and inspection.
Variable pitch propellers are commonly found in more advanced or specialized UAVs \cite{274,275}, such as industrial or military drones, where precise control and optimal performance are crucial. However, they are increasingly becoming accessible in consumer drones as well, allowing hobbyists and enthusiasts to leverage their benefits and enjoy greater control and versatility in their aerial endeavours.

Lastly, a significant advancement in UAV technology is the integration of \textbf{transformable rotors} \cite{276,277}. UAVs equipped with transformable rotors have the ability to modify the configuration of their rotors during flight, enabling them to adapt to various flight conditions or mission requirements. This includes the capability to change the number or orientation of the rotors.
The development of transformable UAVs holds tremendous potential in revolutionizing the field of unmanned aerial vehicles. It empowers UAVs to perform a broader range of missions with increased effectiveness and efficiency. One remarkable example is the VA-X4, which features four rotors that can tilt forward, transitioning from vertical takeoff and landing (VTOL) mode to forward flight mode. This design allows the UAV to achieve higher speeds of up to 200 mph, enabling it to cover longer distances more efficiently \cite{276,277}.
Another notable transformable rotor UAV is the Voliro Hexcopter developed by the ETH Zurich team \cite{280,281}. This hexacopter utilizes multiple rotors capable of providing thrust in various directions, granting the drone the ability to translate freely and manoeuvre in complex environments.
NASA's Greased Lightning GL-10 \cite{282} is yet another remarkable transformable rotor UAV. It can seamlessly transition between a vertical takeoff and landing (VTOL) mode and a fixed-wing mode, optimizing efficiency during forward flight. This UAV is equipped with ten electric motors powering ten rotors, enabling it to achieve high speeds and exceptional manoeuvrability.
These examples demonstrate the immense potential of transformable rotor UAVs in expanding the capabilities and versatility of unmanned aerial systems, paving the way for more efficient and adaptable aerial missions across various industries.

Overall, transformable UAVs have the potential to greatly improve the versatility and efficiency of UAVs, enabling them to adapt to different operational environments and missions. By transforming their shape or configuration in flight, these UAVs can optimize their performance for different flight conditions and mission requirements, making them a valuable tool for a wide range of applications.

\section{Advancements in Aircraft Control: An Overview of the Development Axes} \label{sec6}
Flight dynamics control is a critical area of aerospace engineering that focuses on the stability and control of an aircraft during flight \cite{17}. The main objective of flight dynamics control is to ensure that the aircraft remains stable, controllable, and safe across its entire flight envelope \cite{284}. Various approaches and algorithms are employed in flight dynamics control to design control systems that stabilize and govern the behaviour of aircraft throughout different flight tasks, including path following \cite{285}, morphing capabilities \cite{256,34,36,37}, navigation and surveillance \cite{371}, swarm flights \cite{17}, autonomous manoeuvring \cite{5}, mapping \cite{290}, and sprayer operations \cite{291}. These approaches and algorithms \cite{17,15}, along with their architectural considerations \cite{17,294}, utilize mathematical models of aircraft dynamics and control theory to generate control inputs that modify the aircraft's behaviour to achieve specific performance objectives \cite{15}. The ultimate goal is to ensure safe, stable, and efficient flight operations.

There are several ways to classify UAV control methodologies \cite{294,91,311}, based on factors such as the type of UAV being controlled, the control objectives, and the mathematical techniques employed. Here are some common classifications in control theory: feedback control vs. feedforward control, linear control vs. nonlinear control \cite{297}, continuous-time control vs. discrete-time control, deterministic control vs. stochastic control, model-based control vs. model-free control, robust control \cite{298,299}, and adaptive control \cite{300,301}, centralized control vs. decentralized control, optimal control vs. suboptimal control, and time-invariant control vs. time-varying control. These classifications represent different approaches and algorithms used in flight dynamics control. However, four main categories are widely recognized: classical control, modern control, intelligent control, and adaptive control, each offering distinct methodologies and techniques to address the complexities of flight dynamics control.

\subsection{Classical Control}
Classical control \cite{303,304} refers to traditional control theory based on mathematical models of linear systems. This approach is widely employed in controlling aircraft dynamics and involves the use of proportional, integral, and derivative (PID) control algorithms. The advantage of classical control lies in its simplicity and well-established theoretical foundations. However, it has limitations in handling nonlinear systems and managing disturbances that impact the aircraft's performance.

\subsection{Modern Control}
Modern control \cite{304} encompasses the use of advanced control theory, including state-space and optimal control, to design control systems capable of handling nonlinearities and disturbances. This approach is extensively applied in flight dynamics control, as it provides more precise control and can manage complex systems. The advantage of modern control is its ability to handle nonlinear systems and uncertainties \cite{306}. However, it requires more computational power and sophisticated algorithms, posing challenges for real-time implementation. The Linear Quadratic Regulator (LQR) is a popular control algorithm widely used in flight dynamics control for optimal control of linear systems.

\subsection{Intelligent Control}
Intelligent control \cite{307,308,309} is a branch of control theory that employs AI techniques, such as neural networks, fuzzy logic, and genetic algorithms, to design control systems. This approach finds extensive use in flight dynamics control, as it can adapt to changing conditions and provide a high level of robustness. The advantage of intelligent control lies in its ability to handle complex and nonlinear systems, along with its adaptive nature. However, it requires significant computational resources and can be challenging to analyse and debug. Neural networks are widely employed in flight dynamics control, particularly for control and fault diagnosis. They have found applications in autopilots, flight control systems, and engine control systems.

%-----------------------------------------------------------------

\subsection{Adaptive Control}
Adaptive control is a class of control algorithms that can dynamically adjust control parameters in real-time based on the aircraft's behaviour and environmental conditions \cite{300}. This approach is widely employed in flight dynamics control as it effectively handles uncertainties and disturbances, making it suitable for variable operating conditions. The advantage of adaptive control lies in its ability to accommodate uncertain systems and adapt to changing conditions. However, it requires a substantial amount of data to learn the system's behaviour, which can be challenging to obtain in certain cases. The Model Reference Adaptive Control (MRAC) is a widely used adaptive control algorithm in flight dynamics control, finding applications in various aircraft control systems, including autopilots, flight directors, and flight control systems \cite{301}.

In summary, flight dynamics control approaches and algorithms play a critical role in ensuring the safe and efficient operation of aircraft. The different classes of control techniques possess their respective strengths and weaknesses, and the choice of a particular approach depends on the specific requirements of the control problem. Researchers and practitioners continue to develop and enhance control systems for aircraft, and new approaches and algorithms are expected to emerge. In conclusion, flight dynamics control is an essential field of aerospace engineering that ensures the stability and controllability of an aircraft during its flight. Static stability control, dynamic stability control, and manoeuvrability control are the three primary classifications of flight dynamics control. Each class has its own advantages and disadvantages, and their selection depends on the specific design requirements of the aircraft.

\subsection{Pushing the Boundaries of UAV Control: Exploring Advanced Techniques}
In addition to the widely used PID control, the field of UAV control employs several advanced control techniques that aim to enhance performance and stability \cite{414,417,421}. These techniques are particularly suited for complex and dynamic UAV systems. One prominent advanced control technique in UAV control is Model Predictive Control (MPC). MPC utilizes mathematical models of the UAV's dynamics to predict its future behaviour \cite{412,413}. Based on these predictions, control inputs are computed to optimize a predefined performance metric, such as energy consumption, stability, or trajectory tracking accuracy. By considering the UAV's entire future trajectory, MPC offers improved performance compared to traditional control techniques. Another significantly advanced control technique is Adaptive Control. Adaptive control adjusts its parameters in real-time to adapt to changes in the UAV's environment and dynamics. This enables the control system to continuously enhance its performance over time, even in the presence of uncertainties and disturbances \cite{415,420,301}. These advanced control techniques contribute to the development of more robust and efficient UAV control systems. Sliding Mode Control (SMC) is another advanced control technique that is widely used in UAV control. SMC is a nonlinear control technique that provides robust performance in the presence of uncertainties and disturbances. SMC works by maintaining the UAV's states within a desired operating region, known as the sliding mode, which ensures stability and robustness \cite{480,481}. Finally, there are several advanced control techniques that are based on ML as AI, such as RL and DRL \cite{309,417}, and neural network-based control \cite{307,417}. These techniques enable UAVs to learn from their experiences and improve their control performance over time. In conclusion, there are several advanced control techniques that are being used in the field of UAV control. These techniques provide improved performance and stability compared to traditional control techniques, and they are well-suited to the complex and dynamic requirements of UAV systems. Cooperative control is a technique that enables multiple UAVs to work together to achieve a common mission objective. Cooperative control is particularly useful for UAVs that need to perform complex tasks \cite{316}, such as surveillance or search and rescue, that require coordination and collaboration between multiple UAVs.

%------------------ I am here

\color{black}
\textbf{SMC} is a widely utilized advanced control technique in UAV control. SMC is a nonlinear control method that ensures robust performance even in the presence of uncertainties and disturbances. By maintaining the UAV's states within a desired operating region, known as the sliding mode, SMC guarantees stability and robustness \cite{480,481}. 
The core concept of SMC lies in the creation of a sliding surface, which is a manifold in the state space. The sliding surface is carefully designed to have attractive properties, such as finite-time convergence or robustness against parameter variations. The control law then acts upon the system to drive its state trajectory towards this sliding surface and maintain it there \cite{edwards1998sliding}.
The distinguishing feature of SMC is its discontinuous nature (Chattering) \cite{lee2007chattering}. The control law consists of multiple control actions, or switching functions, that are activated based on the relative position of the system state with respect to the sliding surface. When the system's state is not on the sliding surface, the control law switches between different modes or control actions to robustly drive the state towards the sliding surface. Once the state reaches the sliding surface, the control law switches to a different mode to maintain the system's trajectory on the surface \cite{young1999control,lee2007chattering,edwards1998sliding}.
The discontinuous nature of SMC offers robustness against uncertainties and disturbances since the control actions adapt rapidly based on the state's proximity to the sliding surface. Moreover, the sliding mode itself provides inherent robustness properties as the system's behaviour on the sliding surface is less sensitive to parameter variations or external disturbances.
Mathematically, SMC relies on the theory of differential inclusions and Lyapunov stability analysis to guarantee the system's convergence to the sliding surface and the subsequent maintenance of the system's behaviour on the surface \cite{young1999control}.

The general approach to designing SMC components is to start by defining a sliding surface that represents the desired behaviour of the system. The sliding surface should have attractive properties, such as ensuring stability, convergence, or robustness. The choice of the sliding surface depends on the specific control objective and system dynamics. One commonly used sliding surface in quadrotor control is based on the error between the desired state and the actual state of the UAV. The sliding surface is designed to drive the error dynamics to zero. There are various possible sliding surfaces that can be used based on different UAV objectives and system dynamics. Several sliding surface design strategies have been proposed to minimize or eliminate the reaching mode \cite{tokat2015classification}. These methods can be classified based on dimensions, linearity, time dependence, and the nature of their moving algorithm \cite{tokat2015classification,zheng2014second}.
The aforementioned conventional sliding surface naturally yields a proportional-derivative (PD) sliding surface. To obtain PID structures, an integral action can also be included. Incorporating the integral term is commonly done in conjunction with a boundary layer SMC approach. By including the integral term, the steady-state error resulting from the boundary layer can be eliminated \cite{tokat2015classification}.
The author in \cite{tokat2015classification} has conducted extensive work in classifying the sliding surface into different categories, including Linear Constant Sliding Surface, Linear Discretely-Moving Sliding Surface, Linear Continuously-Moving Sliding Surface, Constant Nonlinear Sliding Surface, and Nonlinear Time-Varying Sliding Surface. All these sliding surfaces are designed to improve controller performance by minimizing or eliminating the time required to reach the sliding phase.
The second step in SMC design is to determine the equivalent control law and switching Law. After defining the sliding surface, the equivalent control law is derived to drive the system dynamics onto the sliding surface and maintain them there. The equivalent control law is typically obtained by analyzing the system's dynamic equation is designed to ensure that the system exhibits desirable behaviour and achieves the control objectives. It involves determining a control signal that will force the system state to follow the sliding surface and stabilize the system. The control law should be designed to counteract the effects of uncertainties, disturbances, and non-linearities in the system. The switching law is a crucial component that ensures the system's states remain on the sliding surface. It plays a key role in rejecting disturbances, uncertainties, and other external factors that may affect the system's performance.  by employing a discontinuous control signal, specifically a set-valued control signal. This signal compels the system to "slide" along a section of its typical behaviour \cite{bartoszewicz2015new}. There exist various types of reaching laws for SMC, including switching and non-switching reaching laws \cite{bartoszewicz2015new}.
Additionally, a new non-switching reaching law has been introduced, demonstrating improved system robustness without amplifying the magnitude of critical signals in the system \cite{bartoszewicz2015new}. Non-switching reaching laws eliminate the need for switching across the sliding hyperplane in each subsequent step \cite{latosinski2021non}. 
Additionally, a new non-switching reaching law has been introduced, demonstrating improved system robustness without amplifying the magnitude of critical signals in the system \cite{bartoszewicz2015new}. Non-switching reaching laws eliminate the need for switching across the sliding hyperplane in each subsequent step \cite{latosinski2021non}.\\

\textbf{ML}. These techniques, including RL, DRL \cite{309,417}, and neural network-based control \cite{307,417}, empower UAVs to learn from their experiences and continually enhance their control performance over time. 
UAVs have greatly benefited from the application of ML, enabling them to efficiently perform assigned tasks \cite{choi2019unmanned}. Researchers have explored the potential of UAVs in various areas such as inspection, delivery, and surveillance \cite{choi2019unmanned}. ML techniques have been employed to provide control strategies, including adaptive control in uncertain environments, real-time path planning, and object recognition \cite{choi2019unmanned}.
An interesting approach to UAV control using ML is DRL. This method allows UAVs to autonomously discover optimal control laws by interacting with the system and handling complex nonlinear dynamics \cite{khan2019unmanned}. Remarkably, DRL has demonstrated success in attitude control of fixed-wing UAVs using the original nonlinear dynamics with as little as three minutes of flight data \cite{khan2019unmanned}.
The integration of ML has not only enhanced UAV capabilities but also reduced challenges, opening doors to various sectors \cite{ben2022uav}. This combination has yielded fast and reliable results \cite{ben2022uav}. In the realm of UAV flight controller designs, Model-Based Control (MBC) techniques have traditionally dominated. However, they heavily rely on accurate mathematical models of the real plant and face complexity issues. Artificial Neural Networks (ANNs) offer a promising solution to address these challenges due to their unique features and advantages in system identification and controller design \cite{gu2020uav}.
A comprehensive survey examines the combination of MBC and ANNs for UAV flight control, particularly in low-level control \cite{gu2020uav}. The objective is to establish a foundation and facilitate efficient controller designs with performance guarantees \cite{gu2020uav}. Fuzzy logic has been utilized to design autonomous flight control systems for UAVs \cite{cordoba2007autonomous}. For instance, a study focused on UAV flight dynamics and developed longitudinal and lateral controllers based on fuzzy logic \cite{hajiyev2015fuzzy}. Despite not employing optimization techniques or dynamic model knowledge, the fuzzy logic controller exhibited satisfactory performance \cite{hajiyev2015fuzzy}.
Another example involves an ANFIS-based autonomous flight controller for UAVs, which utilizes three fuzzy logic modules to control the UAV's position in three-dimensional space, including altitude and longitude-latitude location \cite{kurnaz2010adaptive}.
In summary, the field of UAV control encompasses several advanced techniques that offer improved performance and stability compared to traditional control approaches. These techniques are well-suited to meet the complex and dynamic requirements of UAV systems.\\

\textbf{Cooperative control} is a technique that facilitates the collaboration of multiple UAVs to accomplish a shared mission objective. Particularly for tasks like surveillance or search and rescue, where coordination and collaboration among multiple UAVs are essential, cooperative control proves to be highly beneficial \cite{316}.
The cooperative control of UAVs entails the effective coordination and collaboration among multiple drones to accomplish shared goals \cite{sargolzaei2020control}. UAV swarms bring forth advantages in terms of improved efficiency, flexibility, accuracy, robustness, and reliability \cite{sargolzaei2020control}. Nevertheless, the integration of external communications introduces the possibility of encountering additional faults, failures, uncertainties, and cyberattacks, which can potentially result in the propagation of errors \cite{sargolzaei2020control}.
For the purpose of ensuring operational safety, the field of cooperative control has seen the development of Fault Detection and Diagnosis (FDD) and Fault-Tolerant Control (FTC) methods \cite{ziquan2022review}. These methods are designed to identify and tolerate faults that may occur in the individual components of UAVs \cite{ziquan2022review}. The FDD unit is responsible for diagnosing faults, while the FTC unit offers appropriate compensation measures \cite{ziquan2022review}.
Cooperative UAVs find wide-ranging applications in diverse fields such as search and rescue operations, border patrol,  mapping tasks, surveillance missions, and military operations. \cite{ryan2004overview}. These tasks are well-suited for autonomous vehicles due to their repetitive or dangerous nature. The utilization of multiple UAVs substantially enhances the efficiency of execution \cite{ryan2004overview}.
In the realm of cooperative control, recent advancements encompass the introduction of a distributed consensus algorithm for multi-agent systems (MAS). This algorithm ensures the delivery of seamless input signals to control channels, effectively mitigating the undesired chattering effect associated with conventional control protocols \cite{kada2020distributed}. Another innovative concept is Cooperative Fault Detection and Diagnosis (CFDD) and Fault-Tolerant Cooperative Control (FTCC), which mitigate the negative impact of component and communication faults that may arise during formation or swarm flights \cite{ziquan2022review}.
Within the cooperative control architecture, drones collect sensory data, communicate, and share information \cite{sargolzaei2020control}. Tasks are assigned based on mission requirements, and optimal paths are generated \cite{sargolzaei2020control}. Control algorithms ensure precise trajectory tracking, while collision avoidance mechanisms ensure safe operations \cite{sargolzaei2020control}. Continuous feedback and adaptation maintain accurate trajectory tracking \cite{sargolzaei2020control}.
By incorporating these steps and advancements, cooperative control enables drones to effectively work together, accomplish complex tasks, and achieve synchronized trajectory following \cite{sargolzaei2020control, ziquan2022review, kada2020distributed}. This enhances efficiency and effectiveness in various applications, such as aerial formations, surveillance, and coordinated mapping missions.
 The cooperative control architecture for UAV begin by collect sensory data and communicate with each other to exchange information. Tasks and roles are assigned based on mission requirements, considering capabilities and resource availability. Optimal paths and trajectories are generated to accomplish assigned tasks, considering mission objectives and environmental constraints.
The control layer translates high-level commands and planned paths into low-level actions, ensuring stability, motion control, and trajectory tracking. The mission management layer oversees the overall mission, adapting plans and allocating resources as needed based on real-time feedback from UAVs.
This integrated approach enables efficient task allocation, precise control, and adaptive mission management, facilitating effective cooperation among the UAVs to accomplish complex objectives.

\color{black}
 \textbf{Fault-tolerant control} (FTC) is a methodology employed to uphold acceptable performance and ensure the safety of a system even when faults or failures occur in its hardware or software components \cite{334}. Such faults or failures may arise from diverse causes, including sensor malfunctions, actuator failures, communication losses, or software errors. The objective of fault-tolerant control is to detect faults or failures and mitigate their effects by reconfiguring the control system or adapting the control law. This can be accomplished through the utilization of redundancy, fault detection and isolation (FDI) techniques \cite{330,331,332}, and fault accommodation strategies.
Redundancy involves the presence of multiple copies of critical hardware or software components that can assume control in the event of a failure or fault. For instance, a UAV may possess redundant sensors or actuators that can be employed if the primary ones fail. FDI techniques employ sensor data to detect and isolate faults or failures in the UAV's hardware or software. FDI can be achieved using a variety of techniques, including observer-based approaches, statistical methods, or analytical redundancy.
Fault accommodation strategies adapt the control law or reconfigure the control system to maintain the UAV's stability and performance in the presence of faults or failures. These strategies may involve switching to backup control law, adjusting control gains, or employing adaptive control techniques. Overall, fault-tolerant control plays a crucial role in ensuring the safe and reliable operation of UAVs even in the presence of faults or failures \cite{333,334}. It enables UAVs to detect and mitigate the effects of faults or failures, allowing them to continue operating effectively and achieve their mission objectives. 
Recent advancements in the field of FTC for UAVs have been the subject of several studies. A survey article provides a comprehensive overview of recent research on fault diagnosis, FTC, and anomaly detection specifically tailored for UAVs \cite{fourlas2021survey}. Additionally, a separate review focuses on the topic of fault-tolerant cooperative control, specifically addressing the control of multiple UAVs in a fault-tolerant manner, this study delves into the recent developments in Fault-Tolerant Cooperative Control and offers a systematic analysis of FTCC methods applicable to multi-UAV scenarios. The study initially summarizes and analyzes formation control strategies for fault-free flight conditions of multi-UAVs \cite{ziquan2022review}. Furthermore, an adaptive fault-tolerant control method integrated with fast terminal sliding mode control (FTSMC) technology and neural network is proposed for the attitude system of a quadrotor UAV in another study \cite{gao2022adaptive}. The utilization of the NN allows for the approximation of uncertain terms within the system, thereby enhancing fault-tolerant capabilities.

\color{black}
 \textbf{Prescribed Performance Control (PPC)} is a control methodology specifically designed to fulfil prescribed performance criteria. In PPC, prescribed performance refers to ensuring that the tracking error converges to a predefined small residual set while satisfying a predetermined convergence rate and limiting the maximum overshoot to a sufficiently small constant, Consequently, the desired transient performance metrics, such as overshoot and convergence time , are successfully achieved \cite{bu2023prescribed}. This passage emphasizes the significance of achieving good transient performance in aircraft control systems, encompassing traditional  airplanes, hypersonic flight vehicles, and unmanned aerial vehicles (UAVs). It discusses several studies and methodologies that employ PPC to achieve this goal \cite{bu2023prescribed}.  The study \cite{song2015integrated} focuses on the traditional PPC approach to develop an integrated guidance and control method for interceptors, ensuring excellent transient performance during target interception. 
In the case of hypersonic flight vehicles, studies \cite{huang2022prescribed, chen2020adaptive} indicate that the normal PPC guarantees satisfactory transient performance. However, satisfying the strict initial condition for tracking error poses a challenge. To address this, modified versions of PPC have been proposed \cite{bu2018prescribed} as non-affine models to reduce reliance on initial error. Nevertheless, these methods may result in large overshoot due to the initial value selection of the performance function. To mitigate this issue, a newly designed performance function \cite{bu2017prescribed}  is applied to develop a concise neural tracking controller for hyper-sonic flight vehicles. Simulation results demonstrate small or zero overshoot in velocity and altitude tracking.
UAVs, which hold significant potential in military and civil applications, greatly benefit from good transient performance to carry out tasks effectively. Numerous tracking control methodologies utilizing PPC have been proposed. \cite{li2020adaptive} present a fuzzy-back-stepping-based tracking controller with prescribed performance for a single UAV. Additionally, PPC is applied to platoon control , leader-follower control , Decentralized, finite-time, adaptive fault-tolerant , synchronization control Multi-UAVs \cite{yu2020decentralized}, and quadrotor UAVs  Backstepping-based \cite{koksal2020backstepping}, simulation results validate their superior performance in achieving both transient and steady-state performance. In summary, this passage underscores the application of PPC in achieving desirable transient performance across a range of aircraft, including airplanes, hypersonic flight vehicles, and UAVs. The referenced studies provide evidence of the effectiveness of PPC methodologies.

\color{black}

By extensively examining various literary sources, we have derived Table \ref{table_control}, which outlines the pros and cons of each controller as per the opinions of different authors.

\begin{table*}[t!]
\caption{Summarizing and Comparing Pros and Cons: Control Techniques in UAVs Field.}
\label{table_control}
\begin{tabular}{
m{25mm}
m{68mm}
m{68mm}
}
\hline

Control Technique & Advantage & Disadvantage   \\

\hline

\textbf{ PID \newline
\cite{16,91,294,304,amin2016review,roy2021review,hasseni2021parameters,hasseni2018decentralized,417,421}
  } & 

(1)Implementation is simple. 
(2) The reduction of steady state error can be achieved by increasing parameter gains. 
(3) It consumes minimal memory. 
(4) The design is user-friendly, and it responds well.
&  
 
(1) Conducting experiments can be a time-consuming process.
(2) In certain cases, aggressive gain and overshooting may occur.
(3) There is a possibility of overshoot occurrences when adjusting the parameters
 
	 \\
\hline  \textbf { SMC \newline
 \cite{castillo2018comparison,480,10110876,16,91,294,304,amin2016review,roy2021review} 
 } 
& 
(1) It exhibits high insensitivity to variations in parameters and disturbances.
(2) It is capable of delivering significant implementation efforts.
(3) Linearization of dynamics is not necessary for its operation.
(4) It is efficient in terms of time.
(5) Filtering techniques can be employed to reduce chattering effects.

&
(1) Severe chattering effects occur during switching.
(2) The process of designing such a controller is intricate.
(3) The sliding control scheme heavily depends on the sliding surface, and an incorrect design can result in unsatisfactory performance.
 \\
\hline \textbf  { LQR \newline
 \cite{okasha2022design,masse2018modeling,16,91,294,304,amin2016review,roy2021review} 
 }
 &
(1) Achieves robust stability while minimizing energy consumption.
(2) Demonstrates computational efficiency.
(3) The effectiveness of the system is enhanced by incorporating the Kalman filter
&
(1) Complete access to the system states is necessary, but this is not always feasible.
(2) There is no assurance regarding the speed of response.
(3) It is not suitable for systems that demand a consistently minimal steady-state error.

\\
\hline \textbf { Gain Scheduling  \newline
 \cite{rugh2000research,bouzid2021pid,16,91,294,304,amin2016review,roy2021review} 
 } & 
(1) Facilitates rapid response of the controller to dynamic changes in operating conditions.
(2) The design approach seamlessly integrates with the overall problem, even when dealing with challenging nonlinear problems.
& 
(1) It is not time efficient.
(2) Gain scheduling heavily relies on conducting extensive simulations.
(3) There are no guaranteed performance outcomes.

\\
\hline \textbf { Backstepping  \newline
 \cite{glida2020optimal,abdou2018integral,16,91,294,304,amin2016review,roy2021review} }
&
(1) Demonstrates robustness in the face of constant external disturbances.
(2) Handles all states within the system and is capable of dealing with nonlinear systems.

&
(1) It is not efficient in terms of time.
(2) It is sensitive to variations in parameters.
(3) Implementation can be challenging.
\\
\hline \textbf {  H-Infinity \newline
 \cite{16,91,294,304,amin2016review,roy2021review,seif2020robust,abdou2018integral,masse2018modeling}} 
 &
  (1) Capable of operating in the presence of uncertainties within a system.
(2) Complex control problems are addressed in two subsections: stability and performance.
(3) Offers robust performance.
 &
(1) Involves intricate mathematical algorithms.
(2) Implementation can be challenging.
(3) It necessitates a reasonably accurate model of the system to be controlled.
 \\
\hline \textbf { Adaptive control \newline
\cite{300,301,414,415,419,420,16,91,294,304,amin2016review,roy2021review}} &
 
(1) Capable of handling systems with unpredictable parameter variations and disturbances.
(2) Capable of handling unmodeled dynamics.
(3) Exhibits rapid responsiveness to varying parameters.
&
(1) An accurate model of the system is necessary.
(2) Implementing the design can be time-consuming.
(3) It requires extensive design work before final implementations.
\\
\hline \textbf { AI: Fuzzy Logic and Neural Network \newline
\cite{16,91,294,304,amin2016review,roy2021review,nguyen2019fuzzy, ferdaus2020towards,wai2019adaptive,307,417}}
&
(1) The control action is heavily influenced by the provided rules.
(2) The controller can be manually prepared.
(3) Capable of withstanding unknown disturbances.
(4) Offers adaptive parameters for uncertain models.
(5) The selected control system can be trained.
&
(1) Stability cannot be guaranteed.
(2) Continuous tuning is necessary for critical systems.
(3) It consumes a significant amount of computational power.
(4) Offline learning may fail when uncertainties are present.

\\ \hline

\end{tabular}
%\end{adjustwidth}

\end{table*}

\subsection{Considerations for Selecting an Appropriate Control Algorithm}
Choosing the right control algorithm for a UAV is a crucial decision that depends on various factors, including the type of UAV, its mission objectives, environmental conditions, and available hardware and software resources. Several considerations should be taken into account when selecting or developing a control algorithm.
Firstly, it is essential to identify the UAV's requirements. Understanding the UAV's performance requirements, such as flight range, payload capacity, and environmental conditions, is crucial in determining the most suitable control algorithm.
Next, evaluate different control algorithms available for UAVs, including classical PID, PID2, LQR, sliding mode control, and model reference adaptive controller. Assess the strengths and weaknesses of each algorithm and consider how well they align with the UAV's requirements.
Consider the implementation complexity of the chosen control algorithm. Some algorithms may require significant computational resources or specialized hardware, such as ML-based algorithms or cooperative decision and control \cite{316}. Select an algorithm that can be easily implemented on the UAV's hardware platform, considering the available computing resources.
Once a control algorithm is chosen, optimize its performance by tuning its control parameters. Flight testing can be conducted to collect data on the UAV's flight behaviour and adjust the control algorithm's parameters for optimal performance.
Here are some key considerations when selecting or developing a control algorithm for a UAV \cite{91,294,325}:

\begin{enumerate}
\item \textbf{Stability and Control:} The algorithm should ensure the UAV's stability and controllability, even under turbulent or challenging conditions.
\item \textbf{Performance:} The algorithm should enable the UAV to achieve its performance objectives, such as speed, altitude, and manoeuvrability while minimizing power consumption and optimizing mission duration.
\item \textbf{Sensitivity to Environment:} The algorithm should consider environmental factors that can affect the UAV's performance, such as wind, temperature, and humidity.
\item \textbf{Responsiveness:} The algorithm should be capable of responding quickly to changes in the UAV's mission objectives or unexpected events, such as obstacles or other aircraft.
\item \textbf{Computational Requirements:} The algorithm should be computationally efficient and feasible for the available onboard processing hardware and software.
\item \textbf{Robustness:} The algorithm should be robust to uncertainties, such as sensor noise or errors in the UAV's kinematic model.
\item \textbf{Safety:} The algorithm should ensure the UAV operates safely and avoids collisions with other objects, people, or animals.
\item \textbf{Regulatory Compliance:} The algorithm should comply with local regulations and guidelines for UAV operations, such as flight altitude restrictions and flight path limitations.
\item \textbf{Human Interaction:} The algorithm should enable human operators to interact with the UAV and provide inputs or commands, if necessary.
\end{enumerate}

Overall, the technique for choosing the right control algorithm for where UAV involves careful consideration of where UAV's requirements, evaluation of different control algorithms, simulation and testing, implementation complexity, and optimization.

\section{UAVs Fundamental Hardware/Software Architectures: Applications and Issues }
\subsection{ The Hardware Architecture of UAVs} 
The hardware architecture of Unmanned Aerial Vehicles (UAVs) encompasses several critical components, such as the flight computer and controller, sensors, actuators, battery, communication interfaces, payload, and structural components. The flight computer and controller manage the flight path and stabilization of the UAV. Sensors provide crucial navigation, altitude, and orientation data, while actuators control the UAV's movement. The battery, chosen according to the application's requirements, powers the UAV. Communication interfaces enable remote control and data transmission, whereas the payload may include cameras, sensors, and other equipment specific to the application. Structural components, including the frame and arms, ensure support and stability for the UAV, allowing effective operation across applications. This hardware architecture significantly influences the capabilities and performance of UAVs across varying contexts \cite{339,9,164,322}.
Numerous studies have examined general hardware architecture. Multilevel architecture for UAVs has been proposed in the literature \cite{339,9,164,322}. The characteristics of modern UAVs have been thoroughly discussed in \cite{339, 340}. The aspects of miniature UAVs are treated in \cite{341}, and the required software components for real-time control implementation for UAVs are addressed in \cite{284}. Regardless of scale, UAVs typically include the same components, with additions and modifications according to the application. 

\subsubsection{ Flight Computer and Controller:} The flight computer serves as the primary processing unit that governs the UAV's navigation and flight. It amalgamates data from all the sensors and actuators to manage the UAV's flight trajectory and stability. For medium or large-scale UAVs, this unit is generally more advanced and potent, given the system's heightened complexity. A wealth of literature provides comprehensive reviews and surveys on flight controllers and computers \cite{164,344,10,346}. Other works delve into the design and implementation of a UAV flight controller \cite{347}.

\subsubsection{Sensors:} The sensor suite of a UAV typically encompasses a range of devices such as accelerometers, gyroscopes, magnetometers, barometers, GPS, and cameras. These sensors offer crucial information regarding the UAV's orientation, position, and surrounding environment. Medium or large-scale UAVs might additionally incorporate LIDAR \cite{sohail4348272deep}, radar, and other specialized sensors for applications like navigation and mapping \cite{himeur2022using}. A brief comparison of remote-sensing platforms is presented in \cite{348}. For agricultural applications, the sensors, and their deployment are discussed in \cite{349,350}, while their use in construction and civil applications is detailed in \cite{105}. An examination of electromagnetic interference on UAV sensors is covered in \cite{352}, and the application of different sensors in mining areas is discussed in \cite{107}. Multi-sensor data fusion using Deep Learning (DL) for security and surveillance purposes is reviewed in \cite{354}, and remote sensing for forest health monitoring is explored in \cite{355}.  

 \subsubsection{Actuators:} This includes motors, servos, and electronic speed controllers (ESCs) that control the UAV's movement and stability. Medium or large-scale UAVs are typically larger and more powerful to handle the increased weight and size of the UAV. UAV Electric Propulsion System reviewed in \cite{356,124}. The application of the development of an actuator is analysed in \cite{357}. Another development of propulsion system plasma actuators elaborated in \cite{358}.

\subsubsection{Battery:} This is the power source that provides energy to the UAV's components. A medium or large-scale UAV typically requires a larger battery to provide enough energy to power its components, many papers focused on UAV supply systems and management of its energy \cite{112}. While some \cite{362} focused on UAV architectures of power supply, their charging techniques are well discussed in \cite{363}, while the wirelessly charging of UAVs is covered in \cite{364}. In addition, \cite{365} presents the key technologies of the fuel cell of UAVs. Intelligent energy management and hybrid power supply for UAVs is reviewed in \cite{366,367}.

\subsubsection{Communication Interfaces:} This includes radios and other communication devices that enable the UAV to transmit data to and receive data from ground stations, other UAVs, or other devices. a medium or large-scale UAV are typically more sophisticated and capable of transmitting and receiving larger amounts of data over longer distances. Explore UAV communication networks issues characteristics, design issues, and applications is reviewed in \cite{18}, where the used communication process for UAV is discussed in \cite{369}. A survey in networking and communication technologies is well presented in \cite{49}. 5G systems and satellite communication for UAV navigation and surveillance are presented by \cite{371}. While \cite{140,373} analysed in-depth the swarm communication architectures. In addition, \cite{374} describes UAV-based IoT communication networks and the applications for IoT for sustainable smart farming are explored in \cite{375}.
 
\subsubsection{Payload:} This is the equipment that the UAV is carrying for its specific mission, such as cameras, sensors, or other specialized equipment. a medium or large-scale UAV is typically more sophisticated and specialized and may include specialized sensors, cameras, or other equipment to support its mission. Payload delivery UAV and dropping payload's capability are investigated respectively in \cite{250} and \cite{253}. 
  
\subsubsection{Structural Components:} This includes the frame, arms, and other components that provide the UAV with its physical structure and support the other components. The structural components of a medium or large-scale UAV are typically larger, stronger, and more complex to support its size and weight. Concept and initial design of a UAV addressed in \cite{380}. And UAV mechanical frame stress analysed by \cite{381}. While \cite{382}  reviewed finite element methods for UAV structural analysis. A UAV wing's structural analysis and optimization investigated by \cite{383}. and small-scale UAV structural design and optimization elaborated on \cite{384,385}.

This is a general hardware architecture for a UAV. Depending on the specific UAV and its mission, the architecture may vary, and additional components may be added to meet specific requirements. In some cases, UAVs may use internal combustion engines to generate electricity, which can be used to power the electric motors. However, this is relatively rare, as electric motors are typically more efficient, reliable, and environmentally friendly than internal combustion engines.

\subsection{The Software Architecture for UAVs} It is a critical component that enables the UAV to operate effectively and efficiently. The UAVs architecture as an autonomous system typically consists of several layers, including the firmware layer, operating system layer, middleware layer, and application layer \cite{388,389,391,392,393}.
The firmware layer includes the code that controls the hardware components of the UAV, such as the flight controller and sensors. The operating system layer provides the necessary interface between the firmware and higher-level software applications. The middleware layer includes software components that provide communication and data exchange between different parts of the system. This layer includes protocols and software libraries for data transmission, processing, and storage. Finally, the application layer includes the software programs that run on top of the other layers and provide the necessary functionality for the UAV's intended application. This layer includes software for flight planning, navigation, data analysis, and payload control. 

Overall, the software architecture for UAVs is a complex and highly integrated system that must be carefully designed to ensure reliable and efficient operation in different applications. Advanced software techniques such as ML and AI are also becoming increasingly important in UAV software architectures to improve autonomy and decision-making capabilities.

\subsection{UAVs Applications and Main Issues } There are numerous and diverse, ranging from commercial to military and scientific applications. The applications of UAVs have been discussed and categorized by numerous kinds of literature \cite{396,397,398,399,400}. Some of the most common applications of UAVs include, Aerial photography and videography UAVs can capture high-quality images and video footage from the air, making them ideal for applications such as filmmaking, real estate, and tourism. In agriculture \cite{401}, UAVs can be used to monitor crops, detect crop diseases, and optimize irrigation and fertilization, leading to more efficient and sustainable agriculture practices. UAVs in search and rescue applications can be equipped with specialized sensors and cameras to aid in search and rescue operations, particularly in remote or hard-to-reach areas. While in military and law enforcement, UAVs used for reconnaissance, surveillance, and target acquisition in military and law enforcement applications. \textcolor{black}{UAVs find applications in infrastructure inspection \cite{397,398}, aiding in the efficient and cost-effective inspection and monitoring of structures such as bridges \cite{402}, pipelines, and power lines. They also contribute significantly to environmental conservation efforts by monitoring parameters such as air and water quality, as well as wildlife populations \cite{29,ezequiel2014uav}. Furthermore, UAVs serve as essential tools in scientific research, facilitating environmental monitoring \cite{403}, atmospheric studies \cite{26,27,28,29,fudala2022use,wojcik2019investigation,zmarz2018application,fudala2022use}, and wildlife tracking. They have notably enhanced our understanding of cryospheric features, both on the surface and subsurface. UAVs have also revolutionized faunal studies by enabling non-invasive methods for accurate counting and morphometric analysis of various animal species. Atmospheric surveys conducted by UAVs offer swift and versatile data collection, including aerosol sample collection. The design and development of specialized platforms, tailored to the challenging Antarctic environment, have been instrumental in the successful deployment of these applications \cite{pina2022uavs}.} Consumer UAVs are designed for personal use, such as hobbyist drones, aerial photography drones, and racing drones. Overall, the applications of UAVs continue to expand and evolve as new technologies and capabilities emerge, providing significant opportunities for innovation and advancement in a wide range of fields. Due to the increasing of UAV applications, making use raises common concerns and issues across industries \cite{405,406, 409}. These include privacy concerns, regulatory compliance, technical challenges, specialized equipment requirements, and potential safety risks. Privacy concerns arise from capturing images and videos without consent, while regulatory compliance and technical challenges relate to operating UAVs in specialized environments. The need for specialized sensors and software, along with potential safety risks from accidents or crashes, are also common concerns for UAV use in different fields.

\section{Key Trends: Open-Source UAV Software and Hardware Projects}
There are many open-source projects related to the application of advanced control techniques in   UAVs. Here are a few examples:
\subsection{PX4 Autopilot} Is an open-source flight control platform \cite{465} used for Autonomous navigation and control applications, it supports a wide range of UAVs. The platform provides advanced control algorithms, such as model predictive control \cite{412,413}, adaptive control \cite{414,415}, and R \cite{416,417}, that can be used to improve the stability and performance of UAVs. It provides a flexible and modular platform for developing autonomous flight systems, including advanced navigation and control algorithms. The software is compatible with a range of hardware platforms and supports a variety of communication protocols.

\subsection{ArduPilot} Is an open-source autopilot platform \cite{418} used for autonomous navigation and control applications, it supports a variety of UAVs, including fixed-wing aircraft, multirotors, and rovers. The platform provides advanced control algorithms, such as adaptive control and AI \cite{419,420}, that can be used to improve the stability and performance of UAVs \cite{421}. It provides a range of features for autonomous navigation and control, including GPS waypoint navigation, automated takeoff and landing, and mission planning. The software is compatible with a wide range of hardware platforms and is actively maintained by a large community of developers. 
  
\subsection{TensorFlow for UAV} Is an open-source project software library \cite{422} for ML and AI applications. In the context of UAVs, TensorFlow can be used for a variety of applications, such as object detection and recognition, path planning and navigation \cite{423,424}, and sensor fusion. For example, TensorFlow can be used to train deep neural networks to recognize objects in images or videos captured by UAVs, which can be useful for applications such as search and rescue or surveillance. TensorFlow can also be used for path planning and navigation, by training ML models to predict the optimal trajectory for a UAV based on environmental and mission constraints. Additionally, TensorFlow can be used for sensor fusion, by integrating data from multiple sensors on a UAV to create a more accurate and complete picture of the environment.
\subsection{Paparazzi UAV} This is an open-source project \cite{425} that provides a complete autopilot system for UAVs, including advanced control algorithms and tools for flight planning and mission execution \cite{426,427}

\subsection{Ground Control} Including QGroundControl \cite{428}. Mission Planner \cite{429}. APM Planner 2 \cite{430} and UgCS \cite{431} are an open-source ground control stations that provides a graphical user interface for controlling UAVs \cite{432,433,434}. This platforms provides advanced control algorithms, such as adaptive control, that can be used to improve the stability and performance of UAVs 

\subsection{AirSim} Aerial Informatics and Robotics Simulation \cite{435}, is an open-source UAV simulator developed by Microsoft. It provides a realistic simulation environment for testing and developing UAV control algorithms, including advanced physics-based modeling of the UAV and its environment \cite{436,437,438}. The software is designed to be flexible and customizable, allowing users to experiment with different control strategies and sensor configurations. The platform provides advanced control algorithms, such as R, that can be used to improve the stability and performance of UAVs.
 \subsection{JdeRobot UAVs} Is an open-source toolkit \cite{439} for developing robotics including UAVs that provides a comprehensive set of tools and algorithms for controlling UAVs. The platform provides advanced control algorithms, such as model predictive control and R, that can be used to improve the stability and performance of UAVs \cite{440,441}.

\subsection{DroneKit and DroneKit-Python} Are open-source SDK framework \cite{442} for building apps that run on top of autopilot software such as ArduPilot and PX4. It provides a set of high-level APIs that allow developers to easily interact with the UAV autopilot system, access telemetry data, and send commands to the UAV. While DroneKit-Python \cite{443}. is a Python library that extends the capabilities of DroneKit and provides a simple and easy-to-use interface for developing Python-based UAV applications. It includes a number of high-level APIs for controlling and monitoring UAVs, as well as support for accessing sensor data, controlling actuators, and implementing custom behaviours. advantage of DroneKit ability to support a wide range of development platforms, including Linux, Windows, and Mac OS X and flexibility in supporting multiple autopilot software platforms, including ArduPilot and PX4. This allows developers to build applications that can work with a wide range of UAV hardware and software configurations. And provides a range of APIs for controlling and monitoring UAVs, including mission planning, telemetry, and vehicle control \cite{444,445,446}. 

\subsection{MAVLink} Is an open-source lightweight communication protocol \cite{447} widely used in the UAVs industry to facilitate communication between different components of a UAV system. Is designed to be platform-independent and can be implemented on a wide range of devices including ground control stations \cite{448}, onboard flight controllers, and other peripherals. MAVLink provides a flexible and extensible messaging system that allows different components of a UAV system to exchange information such as flight status, sensor data, and control commands \cite{449}. This enables the different components of a UAV system to work together in a coordinated and efficient manner. One of the key advantages of using MAVLink is its ability to support a wide range of hardware and software platforms. This allows UAV developers and manufacturers to leverage existing software and hardware components and build a system that meets their specific needs. Another advantage of MAVLink is its ability to support multiple communication protocols including serial, UDP, and TCP. This flexibility makes it possible to use MAVLink in a wide range of applications \cite{448,449,450,451} including ground control stations \cite{448,449}, autonomous vehicles, and remote sensing applications.

%-----------------------------------------------------------------
%-------------------------------

\subsection{ROS for UAVs}
The Robot Operating System (ROS) \cite{452} is an open-source software framework for robotics that offers developers libraries and tools to create and manage robotic systems. In the UAV field, ROS can be utilized to produce autonomous drones and other aerial robots. ROS finds application in various areas in the UAV field, including navigation with SLAM \cite{453} and path planning algorithms \cite{454} to aid UAVs in navigating complex environments and avoiding obstacles. It also facilitates object detection \cite{455,456,457,458}, surveillance, and mapping. By integrating sensors like cameras, LIDAR, and GPS, ROS enables UAVs to perceive their surroundings and make intelligent decisions based on sensor data. Additionally, ROS provides a control framework with interfaces for motor control, communication with onboard sensors \cite{459}, and sensor data processing for decision-making. Before deploying algorithms and control strategies in the real world, ROS offers a powerful simulation environment \cite{455} for testing the functionality of UAVs. One of the significant advantages of using ROS in the UAV field is the large and active community that provides abundant resources, including libraries, tutorials, and forums, to support developers in building and deploying UAVs using ROS. Overall, ROS in the UAV field provides a powerful set of tools for developing autonomous drones and other aerial robots, enabling advanced navigation \cite{460}, perception, control, and Swarm Controller capabilities \cite{456,461}.

These projects are just a few examples of the many open-source projects focused on the application of advanced control techniques in UAVs, such as Sparky2, CC3D and Atom, ArduPilot Mega APM, FlyMaple, Erle-Brain3. By utilizing these projects, researchers and developers gain access to cutting-edge tools and algorithms that can be used to develop advanced UAVs.

\section{High-Level UAV Development Software and Categorization}
High-level UAV development software refers to software that provides a user-friendly interface for developing, testing, and deploying control and navigation algorithms for UAVs. This type of software simplifies the process of developing and testing control algorithms by offering an abstract, high-level interface that does not require in-depth knowledge of the underlying hardware and software systems \cite{445}. High-level UAV development software can be categorized into five main classes, including:

%------------------
\subsection{Simulation Software}
Simulation software plays a crucial role in creating virtual environments for testing and evaluating UAVs, eliminating the need for physical prototypes. The simulation program offers accurate and plausible outcomes for financial system design prior to its actual implementation. Nevertheless, the simulation aspect may present certain obstacles that can impede the achievement of desired results if not adequately considered, such as the impact of disturbances like wind, abrupt weather changes, unexpected errors in the dynamic system, or any unforeseen circumstances.
Some widely used simulation software tools in the UAV industry include gazebosim \cite{462}, AirSim \cite{435}, Webots \cite{463}, Morse \cite{464}, jMAVSim \cite{465}, New Paparazzi Simulator \cite{466}, HackflightSim \cite{467}, and Matlab UAV Toolbox \cite{UAVToolbox}.
To differentiate the capabilities of each simulation software based on their advantages and disadvantages, presented in Table \ref{tab5} below:

\begin{table*}[t!]
\caption{Popular Simulation Software Comparison Pros and Cons.}
\label{tab5}
\begin{tabular}{
m{35mm}
m{60mm}
m{60mm}
}
\hline

Simulation Software & Pros & Cons   \\

\hline

\textbf{UAV Toolbox MATLAB
\newline
\cite{UAVToolbox,horri2022tutorial,aliane2022web,xing2015design}
\newline
------------------------
\newline
PL:$MATLAB$,$Fortran$,$C++$,$C$  
SOS:$Windows,Linux,MacOS$\newline
L:$Master License GPCL$
}    & 
\begin{itemize}
    \item The most renowned and widely utilized software
    \item Supports complex simulations
    \item User-friendly Equipped with readily available toolboxes and tools.
    \item The most prevalent software mentioned in research papers.
\end{itemize}
&
\begin{itemize}
    \item Occasionally resource-intensive and time-consuming.
\end{itemize}
\\ \hline
\textbf{GazeBoSim  
\newline
\cite{462,pinedaprm,ivaldi2014tools} 
\newline
------------------------
\newline
PL:$C++$ \newline
SOS: $Linux, MacOS$\newline
L:$Apache V2.0$
} & 
\begin{itemize}
    \item Highly realistic simulation environment with GUI
    \item Open-Source and Extensible
    \item Wide Range of Sensor Support
    \item Integration with ROS
\end{itemize}
&	
\begin{itemize}
    \item Steep Learning Curve
    \item Resource Intensive
    \item Limited Real-World Dynamics
    \item Occasional Stability Issues (Unreasonable robot jumps occur in collision)
\end{itemize}
\\ \hline
\textbf{AirSim
\newline
\cite{435,436,437,438,GPT-Drone2}
\newline
------------------------
\newline
PL:$C++$ \newline
SOS:$Windows, Linux$ \newline
L:$MIT$
}  & 
\begin{itemize}
    \item high-fidelity simulation environment offers accurate physics modeling
    \item Multi-vehicle and Sensor Support
    \item Ability of integration with Popular Frameworks
\end{itemize}
&	
\begin{itemize}
    \item Computationally demanding
    \item Occasional stability bugs (Sensor Synchronization, Communication Errors)
\end{itemize}
\\ 
\hline
\end{tabular}
\noindent{\footnotesize{\textsuperscript{} }}
\end{table*}

\footnotesize{\textbf{PL :} Programming Language, \textbf{SOS : } Supported Operating System, \textbf{L 
: } License}\\
\normalsize (default)
\quad

\subsection{Flight Control Software}
Flight control software is responsible for managing UAV flight operations and onboard systems, including cameras and sensors. Prominent flight control software platforms include ArduPilot \cite{418}, PX4 \cite{465}, Paparazzi \cite{425}, Multiwii series \cite{468}, Cleanflight \cite{469}, Betaflight \cite{470}, INAV \cite{471}), and OpenPilot series \cite{472} (LibrePilot \cite{473}, dRonin \cite{474}).

\subsection{Ground Control Software}
Ground control software enables remote monitoring and control of UAVs. It provides operators with telemetry data, waypoint setting capabilities, and flight control options. Popular ground control software platforms include Ardupilot Mission Planner \cite{429,430}, QGroundControl \cite{428}, and UgCS \cite{431}.

\subsection{Computer Vision Software}
Computer vision software plays a vital role in processing and analyzing images and videos captured by UAVs. It is extensively used in applications like aerial mapping, surveying, and inspection. Well-known computer vision software tools in the UAV industry include OpenCV \cite{478}, TensorFlow \cite{422}, and PyTorch \cite{479}.

\subsection{Sensor Integration Software}
Sensor integration software facilitates the integration of various sensors and systems into UAV platforms. It enables the seamless integration of sensors such as LiDAR, thermal cameras, and multispectral cameras. Prominent sensor integration software tools in the UAV industry include ROS \cite{452} and MAVLink \cite{447}.
These examples represent a fraction of the UAV development software tools commonly employed in the industry. Numerous other specialized software tools are available for specific applications and use cases. The open-source nature of many of these tools has fostered the rapid advancement of UAV technology and the flourishing of the industry.

\section{UAVs Open Issues and Future Research  Directions}
\subsection{Open Issues}

\textcolor{black}{The field of UAVs presents several open issues that researchers are currently grappling with. These issues encompass limitations in operability, including flight autonomy, path planning, battery endurance, flight time, and limited payload carrying capability \cite{mohsan2023unmanned}. Another significant challenge lies in the integration of UAVs within the relief chain to address the specific obstacles faced by international humanitarian organizations (IHOs) \cite{azmat2020potential}.}

\textcolor{black}{Deploying UAV swarms in diverse environments introduces various hurdles, such as decision-making, control, path planning, communication, monitoring, tracking, targeting, collision, and obstacle avoidance \cite{iqbal2022motion,136,140,150}. Additionally, concerns surrounding safety, privacy, security, and power in unmanned systems contribute to the existing challenges. For example, the absence of GPS alerts about the surrounding areas poses potential safety risks, while privacy concerns arise from the collection of personal data by UAVs. The vulnerability of UAV signals to hacking or jamming attempts also represents a security challenge \cite{mohsan2023unmanned}. Other challenges in UAVs can be summarized in ensuring the security of sensitive data, as the lack of encryption exposes UAVs to risks of data hijacking and potential threats of data leakage \cite{mohsan2023unmanned, vattapparamban2016drones}.}

\begin{figure*}[t!]
\includegraphics[width=1\textwidth]{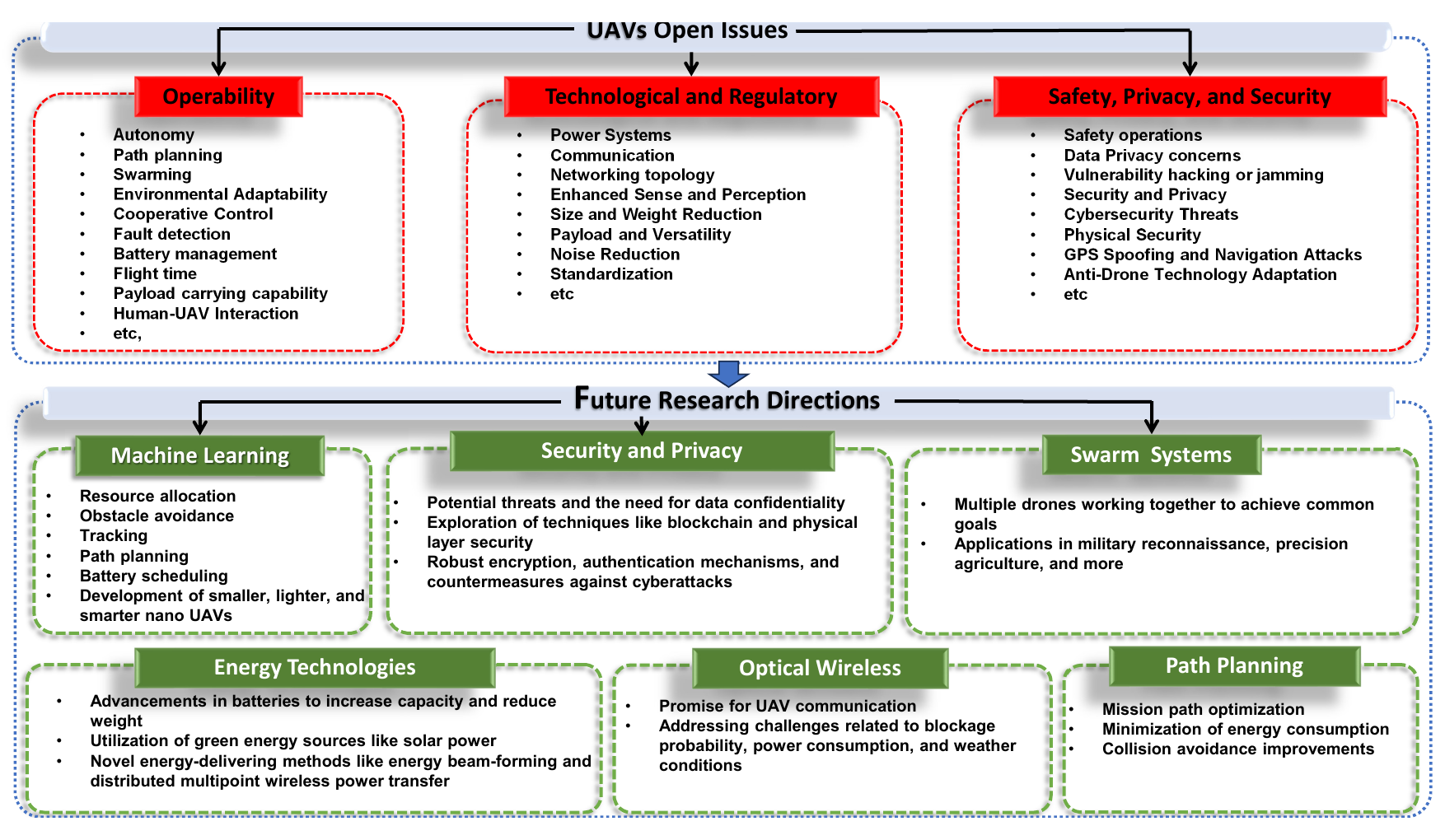}
\caption{ UAVs Open Issues and Future Research Directions.}
 \label{fig2222}
\end{figure*}

\textcolor{black}{Moreover, the need for longer flight times to achieve greater economic impact presents a power-related challenge \cite{mohsan2023unmanned}.  Additionally, the lack of standardization in UAV operations hinders their widespread use. Ambiguity and the absence of significant standards and regulations affect various aspects such as airspace regulations, weight and size limits, privacy considerations, and safety requirements \cite{mohsan2023unmanned, vattapparamban2016drones}.
Utilizing wireless sensors is crucial for enabling smart traffic control systems and enhancing UAV performance \cite{vattapparamban2016drones}. UAVs face limitations in transmission range, processing capability, and speed. Addressing these limitations requires research contributions to advance UAV technology \cite{vattapparamban2016drones}. Resource allocation poses a challenge in terms of optimizing UAV path planning and resource allocation to enhance operational efficiency and performance \cite{vattapparamban2016drones}. Speed limitations can be overcome by regulatory bodies allowing UAVs to operate at higher altitudes \cite{vattapparamban2016drones}. Power limitations and battery life are critical challenges that need to be addressed to improve UAV operations. This includes addressing energy consumption, extending battery life, and developing efficient recharging methods \cite{shakhatreh2019unmanned}.
Furthermore, \cite{kirschstein2021energy} identifies several challenges and problems, including social perception concerns, privacy and safety concerns, and environmental concerns. However, the major challenge in the UAV field is communication, which can be addressed by advancements in microprocessors to enable intelligent autonomous control of various systems. Drones possess distinguished features such as dynamic node mobility and network topology, variable network performance, flight range, autonomous and remote operations, fast data delivery, and cost-effectiveness \cite{gautam2022drone}.}
\textcolor{black}{We have classified the open issues of UAVs into different categories as shown in Figure \ref{fig2222}. These categories encompass challenges related to operability, technology, regulatory aspects, safety, privacy, and security. It's important to note that there may be overlaps between these classes, as certain research directions can have implications across multiple categories. Additionally, this classification is not exhaustive, and there could be alternative classifications based on different perspectives or viewpoints.}

\subsection{Future Research Directions}
 
\textcolor{black}{UAVs field is witnessing future research trends that encompass various areas. One prominent area is swarm UAV systems, where multiple drones collaborate to achieve common goals across various applications, such as military reconnaissance and precision agriculture. Resource allocation, obstacle avoidance, tracking, path planning, and battery scheduling in swarm UAV systems are greatly influenced by machine learning and deep learning algorithms. These algorithms contribute to the development of smaller, lighter, and smarter nano UAVs \cite{elmeseiry2021detailed,208}.}

\textcolor{black}{Security and privacy are critical considerations in UAV systems, given the potential threats and the need to safeguard data confidentiality. Further research is required to explore techniques such as blockchain and physical layer security to address these concerns effectively \cite{heidari2023secure,himeur2022blockchain}. Additionally, trajectory and path planning techniques should be improved to optimize mission paths, minimize energy consumption, and ensure collision avoidance.}

\textcolor{black}{Advancements in energy charging technologies are vital to enhance UAV performance. This includes developing enhanced batteries and utilizing green energy sources like solar power to extend flight times. Novel energy-delivering methods such as energy beam-forming and distributed multipoint wireless power transfer (WPT) can also enhance charging efficiency.}

\textcolor{black}{Optical Wireless Communications (OWCs) hold promise for UAV communication. However, challenges related to blockage probability, power consumption, and weather conditions need to be addressed to ensure reliable and efficient communication in UAV systems \cite{elmeseiry2021detailed}.}

\textcolor{black}{To enhance overall UAV performance, several recommendations can be implemented. Firstly, advancing battery technologies to increase capacity and reduce weight, along with the adoption of efficient charging techniques like wireless or solar power, would significantly benefit UAV operations. Secondly, collision avoidance systems can be improved through the development of advanced algorithms and sensor technologies. Artificial intelligence and machine learning can be utilized for autonomous decision-making in collision avoidance scenarios. Thirdly, the security of UAV systems should be strengthened through the implementation of robust encryption, authentication mechanisms, and effective countermeasures against cyberattacks. Lastly, it is crucial to establish clear and comprehensive regulations for UAV operation, including guidelines for autonomous UAV swarms. These regulations should address safety, privacy, and environmental considerations to ensure responsible and sustainable UAV deployment \cite{elmeseiry2021detailed,mohsan2023unmanned,6}.}

\section{ Conclusion}
In conclusion, this paper presents a comprehensive and in-depth analysis of UAV development, covering various research directions in the last three years, potential open development axes, aircraft control development axes, hardware and software architectures, applications, and key trends in the field. Through the use of the Scopus database and expert analysis, this paper sheds light on recent UAV research trends and the intra-interactions between different research directions.
Furthermore, this paper highlights the integration of cutting-edge technologies such as AI, communications, IoTs, aircraft detection, control and autonomous flight, perception and sensing, environmental monitoring and conservation, miniaturization, swarming and cooperative control, transformability capability as potential open development axes in UAVs. It also discusses classical, modern, intelligent, and adaptive control techniques, pushing the boundaries of UAV control with advanced trends, and the importance of selecting appropriate control algorithms. The paper also provides insights into the fundamental hardware and software architectures of UAVs, including flight computers and controllers, sensors, actuators, batteries, communication interfaces, payloads, and structural components. It discusses diverse applications of UAVs and their main issues, as well as key trends in open-source UAV software and hardware projects. In conclusion, this review paper serves as a valuable resource for researchers, developers, and practitioners interested in staying updated with the latest advancements in UAV development and research trends. It provides a comprehensive overview of key aspects of UAV development, identifies potential open development axes, discusses aircraft control development axes, and explores diverse applications and main issues in the field. With its comprehensive analysis and insights, this paper makes an indispensable contribution to the scientific literature on UAVs, providing valuable guidance for future research and development in this dynamic field.



\begin{thebibliography}{100}
\providecommand{\url}[1]{#1}
\csname url@samestyle\endcsname
\providecommand{\newblock}{\relax}
\providecommand{\bibinfo}[2]{#2}
\providecommand{\BIBentrySTDinterwordspacing}{\spaceskip=0pt\relax}
\providecommand{\BIBentryALTinterwordstretchfactor}{4}
\providecommand{\BIBentryALTinterwordspacing}{\spaceskip=\fontdimen2\font plus
\BIBentryALTinterwordstretchfactor\fontdimen3\font minus
  \fontdimen4\font\relax}
\providecommand{\BIBforeignlanguage}[2]{{%
\expandafter\ifx\csname l@#1\endcsname\relax
\typeout{** WARNING: IEEEtran.bst: No hyphenation pattern has been}%
\typeout{** loaded for the language `#1'. Using the pattern for}%
\typeout{** the default language instead.}%
\else
\language=\csname l@#1\endcsname
\fi
#2}}
\providecommand{\BIBdecl}{\relax}
\BIBdecl

\bibitem{himeur2023face}
Y.~Himeur, S.~Al-Maadeed, I.~Varlamis, N.~Al-Maadeed, K.~Abualsaud, and
  A.~Mohamed, ``Face mask detection in smart cities using deep and transfer
  learning: lessons learned from the covid-19 pandemic,'' \emph{Systems},
  vol.~11, no.~2, p. 107, 2023.

\bibitem{kheddar2023deep1}
H.~Kheddar, Y.~Himeur, and A.~I. Awad, ``Deep transfer learning applications in
  intrusion detection systems: A comprehensive review,'' \emph{arXiv preprint
  arXiv:2304.10550}, 2023.

\bibitem{atalla2023intelligent}
S.~Atalla, M.~Daradkeh, A.~Gawanmeh, H.~Khalil, W.~Mansoor, S.~Miniaoui, and
  Y.~Himeur, ``An intelligent recommendation system for automating academic
  advising based on curriculum analysis and performance modeling,''
  \emph{Mathematics}, vol.~11, no.~5, p. 1098, 2023.

\bibitem{copiaco2023innovative}
A.~Copiaco, Y.~Himeur, A.~Amira, W.~Mansoor, F.~Fadli, S.~Atalla, and S.~S.
  Sohail, ``An innovative deep anomaly detection of building energy consumption
  using energy time-series images,'' \emph{Engineering Applications of
  Artificial Intelligence}, vol. 119, p. 105775, 2023.

\bibitem{himeur2022next}
Y.~Himeur, M.~Elnour, F.~Fadli, N.~Meskin, I.~Petri, Y.~Rezgui, F.~Bensaali,
  and A.~Amira, ``Next-generation energy systems for sustainable smart cities:
  Roles of transfer learning,'' \emph{Sustainable Cities and Society}, p.
  104059, 2022.

\bibitem{elnour2022performance}
M.~Elnour, F.~Fadli, Y.~Himeur, I.~Petri, Y.~Rezgui, N.~Meskin, and A.~M.
  Ahmad, ``Performance and energy optimization of building automation and
  management systems: Towards smart sustainable carbon-neutral sports
  facilities,'' \emph{Renewable and Sustainable Energy Reviews}, vol. 162, p.
  112401, 2022.

\bibitem{kheddar2023deep2}
H.~Kheddar, Y.~Himeur, S.~Al-Maadeed, A.~Amira, and F.~Bensaali, ``Deep
  transfer learning for automatic speech recognition: Towards better
  generalization,'' \emph{arXiv preprint arXiv:2304.14535}, 2023.

\bibitem{atalla2023iot}
S.~Atalla, S.~Tarapiah, A.~Gawanmeh, M.~Daradkeh, H.~Mukhtar, Y.~Himeur,
  W.~Mansoor, K.~F.~B. Hashim, and M.~Daadoo, ``Iot-enabled precision
  agriculture: Developing an ecosystem for optimized crop management,''
  \emph{Information}, vol.~14, no.~4, p. 205, 2023.

\bibitem{al2022smart}
T.~M. Al-Hasan, A.~S. Shibeika, U.~Attique, F.~Bensaali, and Y.~Himeur, ``Smart
  speed camera based on automatic number plate recognition for residential
  compounds and institutions inside qatar,'' in \emph{2022 5th International
  Conference on Signal Processing and Information Security (ICSPIS)}.\hskip 1em
  plus 0.5em minus 0.4em\relax IEEE, 2022, pp. 42--45.

\bibitem{khalife2022achievability}
J.~Khalife and Z.~M. Kassas, ``On the achievability of submeter-accurate uav
  navigation with cellular signals exploiting loose network synchronization,''
  \emph{IEEE Transactions on Aerospace and Electronic Systems}, vol.~58, no.~5,
  pp. 4261--4278, 2022.

\bibitem{elharrouss2021panoptic}
O.~Elharrouss, S.~Al-Maadeed, N.~Subramanian, N.~Ottakath, N.~Almaadeed, and
  Y.~Himeur, ``Panoptic segmentation: a review,'' \emph{arXiv preprint
  arXiv:2111.10250}, 2021.

\bibitem{liu2022hybrid}
W.~Liu, T.~Zhang, S.~Huang, and K.~Li, ``A hybrid optimization framework for
  uav reconnaissance mission planning,'' \emph{Computers \& Industrial
  Engineering}, vol. 173, p. 108653, 2022.

\bibitem{1}
F.~Ahmed, J.~Mohanta, A.~Keshari, and P.~S. Yadav, ``Recent advances in
  unmanned aerial vehicles: A review,'' \emph{Arabian Journal for Science and
  Engineering}, vol.~47, no.~7, pp. 7963--7984, 2022.

\bibitem{2}
S.~N. Ghazbi, Y.~Aghli, M.~Alimohammadi, and A.~A. Akbari, ``Quadrotors
  unmanned aerial vehicles: A review,'' \emph{International journal on smart
  sensing and Intelligent Systems}, vol.~9, no.~1, pp. 309--333, 2016.

\bibitem{3}
M.~S. Ismail, A.~Ahmad, S.~Ismail, and N.~M.~M. Yusop, ``A review on unmanned
  aerial vehicle (uav) threats assessments,'' in \emph{AIP Conference
  Proceedings}, vol. 2617, no.~1.\hskip 1em plus 0.5em minus 0.4em\relax AIP
  Publishing LLC, 2022, p. 050007.

\bibitem{4}
G.~Macrina, L.~D.~P. Pugliese, F.~Guerriero, and G.~Laporte, ``Drone-aided
  routing: A literature review,'' \emph{Transportation Research Part C:
  Emerging Technologies}, vol. 120, p. 102762, 2020.

\bibitem{5}
R.~Zhang, J.~Zhang, and H.~Yu, ``Review of modeling and control in uav
  autonomous maneuvering flight,'' in \emph{2018 IEEE International Conference
  on Mechatronics and Automation (ICMA)}.\hskip 1em plus 0.5em minus
  0.4em\relax IEEE, 2018, pp. 1920--1925.

\bibitem{6}
H.~Liang, S.-C. Lee, W.~Bae, J.~Kim, and S.~Seo, ``Towards uavs in
  construction: Advancements, challenges, and future directions for monitoring
  and inspection,'' \emph{Drones}, vol.~7, no.~3, p. 202, 2023.

\bibitem{pina2022uavs}
P.~Pina and G.~Vieira, ``Uavs for science in antarctica,'' \emph{Remote
  Sensing}, vol.~14, no.~7, p. 1610, 2022.

\bibitem{7}
B.~Fan, Y.~Li, R.~Zhang, and Q.~Fu, ``Review on the technological development
  and application of uav systems,'' \emph{Chinese Journal of Electronics},
  vol.~29, no.~2, pp. 199--207, 2020.

\bibitem{8}
J.~Pasha, Z.~Elmi, S.~Purkayastha, A.~M. Fathollahi-Fard, Y.-E. Ge, Y.-Y. Lau,
  and M.~A. Dulebenets, ``The drone scheduling problem: A systematic
  state-of-the-art review,'' \emph{IEEE Transactions on Intelligent
  Transportation Systems}, 2022.

\bibitem{9}
J.~M{\'e}szar{\'o}s, ``Aerial surveying uav based on open-source hardware and
  software,'' \emph{The International Archives of the Photogrammetry, Remote
  Sensing and Spatial Information Sciences}, vol.~38, pp. 155--159, 2012.

\bibitem{10}
E.~Ebeid, M.~Skriver, K.~H. Terkildsen, K.~Jensen, and U.~P. Schultz, ``A
  survey of open-source uav flight controllers and flight simulators,''
  \emph{Microprocessors and Microsystems}, vol.~61, pp. 11--20, 2018.

\bibitem{12}
L.~O. Rojas-Perez and J.~Mart{\'\i}nez-Carranza, ``On-board processing for
  autonomous drone racing: an overview,'' \emph{Integration}, vol.~80, pp.
  46--59, 2021.

\bibitem{13}
A.~Aabid, B.~Parveez, N.~Parveen, S.~A. Khan, J.~Zayan, and O.~Shabbir,
  ``Reviews on design and development of unmanned aerial vehicle (drone) for
  different applications,'' \emph{J. Mech. Eng. Res. Dev}, vol.~45, no.~2, pp.
  53--69, 2022.

\bibitem{14}
M.~Galimov, R.~Fedorenko, and A.~Klimchik, ``Uav positioning mechanisms in
  landing stations: Classification and engineering design review,''
  \emph{Sensors}, vol.~20, no.~13, p. 3648, 2020.

\bibitem{15}
R.~Amin, L.~Aijun, and S.~Shamshirband, ``A review of quadrotor uav: control
  methodologies and performance evaluation,'' \emph{International Journal of
  Automation and Control}, vol.~10, no.~2, pp. 87--103, 2016.

\bibitem{17}
M.~Campion, P.~Ranganathan, and S.~Faruque, ``Uav swarm communication and
  control architectures: a review,'' \emph{Journal of Unmanned Vehicle
  Systems}, vol.~7, no.~2, pp. 93--106, 2018.

\bibitem{18}
H.~Nawaz, H.~M. Ali, and A.~A. Laghari, ``Uav communication networks issues: a
  review,'' \emph{Archives of Computational Methods in Engineering}, vol.~28,
  pp. 1349--1369, 2021.

\bibitem{19}
Q.~T. Do, D.~S. Lakew, A.~T. Tran, D.~T. Hua, and S.~Cho, ``A review on recent
  approaches in mmwave uav-aided communication networks and open issues,'' in
  \emph{2023 International Conference on Information Networking (ICOIN)}.\hskip
  1em plus 0.5em minus 0.4em\relax IEEE, 2023, pp. 728--731.

\bibitem{20}
R.~Singh, K.~D. Ballal, M.~S. Berger, and L.~Dittmann, ``Overview of drone
  communication requirements in 5g,'' \emph{Internet of Things: 5th The Global
  IoT Summit, GIoTS 2022, Dublin, Ireland, June 20--23, 2022, Revised Selected
  Papers}, pp. 3--16, 2023.

\bibitem{21}
L.~P. Osco, J.~M. Junior, A.~P.~M. Ramos, L.~A. de~Castro~Jorge, S.~N.
  Fatholahi, J.~de~Andrade~Silva, E.~T. Matsubara, H.~Pistori, W.~N.
  Gon{\c{c}}alves, and J.~Li, ``A review on deep learning in uav remote
  sensing,'' \emph{International Journal of Applied Earth Observation and
  Geoinformation}, vol. 102, p. 102456, 2021.

\bibitem{22}
A.~Puente-Castro, D.~Rivero, A.~Pazos, and E.~Fernandez-Blanco, ``A review of
  artificial intelligence applied to path planning in uav swarms,''
  \emph{Neural Computing and Applications}, pp. 1--18, 2022.

\bibitem{chen2023yolo}
C.~Chen, Z.~Zheng, T.~Xu, S.~Guo, S.~Feng, W.~Yao, and Y.~Lan, ``Yolo-based uav
  technology: A review of the research and its applications,'' \emph{Drones},
  vol.~7, no.~3, p. 190, 2023.

\bibitem{othman2023development}
N.~A. Othman and I.~Aydin, ``Development of a novel lightweight cnn model for
  classification of human actions in uav-captured videos,'' \emph{Drones},
  vol.~7, no.~3, p. 148, 2023.

\bibitem{24}
M.~Hassanalian and A.~Abdelkefi, ``Classifications, applications, and design
  challenges of drones: A review,'' \emph{Progress in Aerospace Sciences},
  vol.~91, pp. 99--131, 2017.

\bibitem{fudala2022use}
K.~Fudala and R.~J. Bialik, ``The use of drone-based aerial photogrammetry in
  population monitoring of southern giant petrels in asma 1, king george
  island, maritime antarctica,'' \emph{Global Ecology and Conservation},
  vol.~33, p. e01990, 2022.

\bibitem{26}
D.~Ventura, A.~Bonifazi, M.~F. Gravina, G.~D. Ardizzone \emph{et~al.},
  ``Unmanned aerial systems (uass) for environmental monitoring: A review with
  applications in coastal habitats,'' \emph{Aerial Robots-Aerodynamics, Control
  and Applications}, pp. 165--184, 2017.

\bibitem{27}
Z.~Yang, X.~Yu, S.~Dedman, M.~Rosso, J.~Zhu, J.~Yang, Y.~Xia, Y.~Tian,
  G.~Zhang, and J.~Wang, ``Uav remote sensing applications in marine
  monitoring: Knowledge visualization and review,'' \emph{Science of The Total
  Environment}, p. 155939, 2022.

\bibitem{28}
R.~H. Kabir and K.~Lee, ``Wildlife monitoring using a multi-uav system with
  optimal transport theory,'' \emph{Applied Sciences}, vol.~11, no.~9, p. 4070,
  2021.

\bibitem{29}
L.~J. Mangewa, P.~A. Ndakidemi, and L.~K. Munishi, ``Integrating uav technology
  in an ecological monitoring system for community wildlife management areas in
  tanzania,'' \emph{Sustainability}, vol.~11, no.~21, p. 6116, 2019.

\bibitem{zmarz2018application}
A.~Zmarz, M.~Rodzewicz, M.~D{\k{a}}bski, I.~Karsznia, M.~Korczak-Abshire, and
  K.~J. Chwedorzewska, ``Application of uav bvlos remote sensing data for
  multi-faceted analysis of antarctic ecosystem,'' \emph{Remote Sensing of
  Environment}, vol. 217, pp. 375--388, 2018.

\bibitem{ezequiel2014uav}
C.~A.~F. Ezequiel, M.~Cua, N.~C. Libatique, G.~L. Tangonan, R.~Alampay, R.~T.
  Labuguen, C.~M. Favila, J.~L.~E. Honrado, V.~Canos, C.~Devaney \emph{et~al.},
  ``Uav aerial imaging applications for post-disaster assessment, environmental
  management and infrastructure development,'' in \emph{2014 International
  Conference on Unmanned Aircraft Systems (ICUAS)}.\hskip 1em plus 0.5em minus
  0.4em\relax IEEE, 2014, pp. 274--283.

\bibitem{32}
F.~Fumian, D.~Di~Giovanni, L.~Martellucci, R.~Rossi, and P.~Gaudio,
  ``Application of miniaturized sensors to unmanned aerial systems, a new
  pathway for the survey of polluted areas: Preliminary results,''
  \emph{Atmosphere}, vol.~11, no.~5, p. 471, 2020.

\bibitem{34}
K.~Patnaik and W.~Zhang, ``Towards reconfigurable and flexible multirotors: A
  literature survey and discussion on potential challenges,''
  \emph{International Journal of Intelligent Robotics and Applications},
  vol.~5, no.~3, pp. 365--380, 2021.

\bibitem{35}
D.~A. Ta, I.~Fantoni, and R.~Lozano, ``Modeling and control of a convertible
  mini-uav,'' \emph{IFAC Proceedings Volumes}, vol.~44, no.~1, pp. 1492--1497,
  2011.

\bibitem{36}
M.~A. da~Silva~Ferreira, M.~F.~T. Begazo, G.~C. Lopes, A.~F. de~Oliveira, E.~L.
  Colombini, and A.~da~Silva~Sim{\~o}es, ``Drone reconfigurable architecture
  (dra): A multipurpose modular architecture for unmanned aerial vehicles
  (uavs),'' \emph{Journal of Intelligent \& Robotic Systems}, vol.~99, no. 3-4,
  pp. 517--534, 2020.

\bibitem{37}
F.~Schiano, P.~M. Kornatowski, L.~Cencetti, and D.~Floreano, ``Reconfigurable
  drone system for transportation of parcels with variable mass and size,''
  \emph{IEEE Robotics and Automation Letters}, vol.~7, no.~4, pp.
  12\,150--12\,157, 2022.

\bibitem{39}
G.~Singhal, B.~Bansod, and L.~Mathew, ``Unmanned aerial vehicle classification,
  applications and challenges: A review,'' 2018.

\bibitem{mohsan2023unmanned}
S.~A.~H. Mohsan, N.~Q.~H. Othman, Y.~Li, M.~H. Alsharif, and M.~A. Khan,
  ``Unmanned aerial vehicles (uavs): Practical aspects, applications, open
  challenges, security issues, and future trends,'' \emph{Intelligent Service
  Robotics}, vol.~16, no.~1, pp. 109--137, 2023.

\bibitem{arafat2023vision}
M.~Y. Arafat, M.~M. Alam, and S.~Moh, ``Vision-based navigation techniques for
  unmanned aerial vehicles: Review and challenges,'' \emph{Drones}, vol.~7,
  no.~2, p.~89, 2023.

\bibitem{42}
R.~Kapoor, A.~Shukla, and V.~Goyal, ``Unmanned aerial vehicle (uav)
  communications using multiple antennas,'' in \emph{Advanced Computational
  Paradigms and Hybrid Intelligent Computing: Proceedings of ICACCP
  2021}.\hskip 1em plus 0.5em minus 0.4em\relax Springer, 2022, pp. 261--272.

\bibitem{43}
C.~Yan, L.~Fu, J.~Zhang, and J.~Wang, ``A comprehensive survey on uav
  communication channel modeling,'' \emph{IEEE Access}, vol.~7, pp.
  107\,769--107\,792, 2019.

\bibitem{44}
Q.~Song, Y.~Zeng, J.~Xu, and S.~Jin, ``A survey of prototype and experiment for
  uav communications,'' \emph{Science China Information Sciences}, vol.~64, pp.
  1--21, 2021.

\bibitem{45}
J.~Zhao, F.~Gao, G.~Ding, T.~Zhang, W.~Jia, and A.~Nallanathan, ``Integrating
  communications and control for uav systems: Opportunities and challenges,''
  \emph{IEEE Access}, vol.~6, pp. 67\,519--67\,527, 2018.

\bibitem{53}
M.~Mozaffari, W.~Saad, M.~Bennis, and M.~Debbah, ``Drone-based antenna array
  for service time minimization in wireless networks,'' in \emph{2018 IEEE
  international conference on communications (ICC)}.\hskip 1em plus 0.5em minus
  0.4em\relax IEEE, 2018, pp. 1--6.

\bibitem{55}
M.~M. Khan, S.~Hossain, P.~Majumder, S.~Akter, and R.~H. Ashique, ``A review on
  machine learning and deep learning for various antenna design applications,''
  \emph{Heliyon}, p. e09317, 2022.

\bibitem{56}
C.~Wu and C.-F. Lai, ``A survey on improving the wireless communication with
  adaptive antenna selection by intelligent method,'' \emph{Computer
  Communications}, vol. 181, pp. 374--403, 2022.

\bibitem{58}
R.~Kapoor, A.~Shukla, and V.~Goyal, ``Analysis of multiple antenna techniques
  for unmanned aerial vehicle (uav) communication,'' in \emph{IOT with Smart
  Systems: Proceedings of ICTIS 2021, Volume 2}.\hskip 1em plus 0.5em minus
  0.4em\relax Springer, 2022, pp. 347--357.

\bibitem{59}
N.~Parvaresh, M.~Kulhandjian, H.~Kulhandjian, C.~D'Amours, and B.~Kantarci, ``A
  tutorial on ai-powered 3d deployment of drone base stations: State of the
  art, applications and challenges,'' \emph{Vehicular Communications}, p.
  100474, 2022.

\bibitem{SHI2021104340}
\BIBentryALTinterwordspacing
L.~Shi, N.~J.~H. Marcano, and R.~H. Jacobsen, ``A review on communication
  protocols for autonomous unmanned aerial vehicles for inspection
  application,'' \emph{Microprocessors and Microsystems}, vol.~86, p. 104340,
  2021. [Online]. Available:
  \url{https://www.sciencedirect.com/science/article/pii/S014193312100497X}
\BIBentrySTDinterwordspacing

\bibitem{hy}
M.~Khan, I.~Qureshi, and F.~Khanzada, ``A hybrid communication scheme for
  efficient and low-cost deployment of future flying ad-hoc network (fanet),''
  \emph{Drones}, vol.~3, no.~1, p.~16, 2019.

\bibitem{550}
T.~Brown, B.~Argrow, C.~Dixon, S.~Doshi, R.-G. Thekkekunnel, and D.~Henkel,
  ``Ad hoc uav ground network (augnet),'' in \emph{AIAA 3rd" Unmanned
  Unlimited" Technical Conference, Workshop and Exhibit}, 2004, p. 6321.

\bibitem{560}
E.~Yanmaz, S.~Hayat, J.~Scherer, and C.~Bettstetter, ``Experimental performance
  analysis of two-hop aerial 802.11 networks,'' in \emph{2014 IEEE Wireless
  Communications and Networking Conference (WCNC)}.\hskip 1em plus 0.5em minus
  0.4em\relax IEEE, 2014, pp. 3118--3123.

\bibitem{580}
M.~Asadpour, D.~Giustiniano, K.~A. Hummel, S.~Heimlicher, and S.~Egli, ``Now or
  later? delaying data transfer in time-critical aerial communication,'' in
  \emph{Proceedings of the ninth ACM conference on Emerging networking
  experiments and technologies}, 2013, pp. 127--132.

\bibitem{590}
S.~Morgenthaler, T.~Braun, Z.~Zhao, T.~Staub, and M.~Anwander, ``Uavnet: A
  mobile wireless mesh network using unmanned aerial vehicles,'' in \emph{2012
  IEEE globecom workshops}.\hskip 1em plus 0.5em minus 0.4em\relax IEEE, 2012,
  pp. 1603--1608.

\bibitem{600}
F.~Lv, H.~Zhu, H.~Xue, Y.~Zhu, S.~Chang, M.~Dong, and M.~Li, ``An empirical
  study on urban ieee 802.11 p vehicle-to-vehicle communication,'' in
  \emph{2016 13th Annual IEEE International Conference on Sensing,
  Communication, and Networking (SECON)}.\hskip 1em plus 0.5em minus
  0.4em\relax IEEE, 2016, pp. 1--9.

\bibitem{zigbee}
D.~S. Pereira, M.~R. De~Morais, L.~B. Nascimento, P.~J. Alsina, V.~G. Santos,
  D.~H. Fernandes, and M.~R. Silva, ``Zigbee protocol-based communication
  network for multi-unmanned aerial vehicle networks,'' \emph{IEEE Access},
  vol.~8, pp. 57\,762--57\,771, 2020.

\bibitem{santoszigbee}
V.~G. SANTOS, D.~H. FERNANDES, and M.~R. SILVA, ``Zigbee protocol-based
  communication network for multi-unmanned aerial vehicle networks.''

\bibitem{2022comparative}
G.~A. QasMarrogy and A.~J. Fadhil, ``A comparative study of different fanet
  802.11 wireless protocols with different data loads,'' \emph{Polytechnic
  Journal}, vol.~12, no.~1, pp. 61--66, 2022.

\bibitem{paredes2023lora}
W.~D. Paredes, H.~Kaushal, I.~Vakilinia, and Z.~Prodanoff, ``Lora technology in
  flying ad hoc networks: A survey of challenges and open issues,''
  \emph{Sensors}, vol.~23, no.~5, p. 2403, 2023.

\bibitem{noor2020review}
F.~Noor, M.~A. Khan, A.~Al-Zahrani, I.~Ullah, and K.~A. Al-Dhlan, ``A review on
  communications perspective of flying ad-hoc networks: key enabling wireless
  technologies, applications, challenges and open research topics,''
  \emph{Drones}, vol.~4, no.~4, p.~65, 2020.

\bibitem{martinez2019iot}
J.-M. Martinez-Caro and M.-D. Cano, ``Iot system integrating unmanned aerial
  vehicles and lora technology: A performance evaluation study,''
  \emph{Wireless Communications and Mobile Computing}, vol. 2019, pp. 1--12,
  2019.

\bibitem{500}
L.~Zhu, D.~Yin, J.~Yang, and L.~Shen, ``Research of remote measurement and
  control technology of uav based on mobile communication networks,'' in
  \emph{2015 ieee international conference on information and
  automation}.\hskip 1em plus 0.5em minus 0.4em\relax IEEE, 2015, pp.
  2517--2522.

\bibitem{510}
Y.~Xu and G.~Gui, ``Optimal resource allocation for wireless powered
  multi-carrier backscatter communication networks,'' \emph{IEEE Wireless
  Communications Letters}, vol.~9, no.~8, pp. 1191--1195, 2020.

\bibitem{520}
H.~C. Nguyen, R.~Amorim, J.~Wigard, I.~Z. Kov{\'a}cs, T.~B. S{\o}rensen, and
  P.~E. Mogensen, ``How to ensure reliable connectivity for aerial vehicles
  over cellular networks,'' \emph{Ieee Access}, vol.~6, pp. 12\,304--12\,317,
  2018.

\bibitem{371}
N.~Hosseini, H.~Jamal, J.~Haque, T.~Magesacher, and D.~W. Matolak, ``Uav
  command and control, navigation and surveillance: A review of potential 5g
  and satellite systems,'' in \emph{2019 IEEE Aerospace Conference}.\hskip 1em
  plus 0.5em minus 0.4em\relax IEEE, 2019, pp. 1--10.

\bibitem{colajanni2022service}
G.~Colajanni, P.~Daniele, L.~Galluccio, C.~Grasso, and G.~Schembra, ``Service
  chain placement optimization in 5g fanet-based network edge,'' \emph{IEEE
  Communications Magazine}, vol.~60, no.~11, pp. 60--65, 2022.

\bibitem{amponis2022drones}
G.~Amponis, T.~Lagkas, M.~Zevgara, G.~Katsikas, T.~Xirofotos, I.~Moscholios,
  and P.~Sarigiannidis, ``Drones in b5g/6g networks as flying base stations,''
  \emph{drones}, vol.~6, no.~2, p.~39, 2022.

\bibitem{grasso2022tailoring}
C.~Grasso, R.~Raftopoulos, and G.~Schembra, ``Tailoring fanet-based 6g network
  slices in remote areas for low-latency applications,'' \emph{Procedia
  Computer Science}, vol. 203, pp. 69--78, 2022.

\bibitem{60}
M.~T.~d. Oliveira, R.~K. Miranda, J.~P.~C. da~Costa, A.~L. de~Almeida, and
  R.~T.~d. Sousa, ``Low cost antenna array based drone tracking device for
  outdoor environments,'' \emph{Wireless Communications and Mobile Computing},
  vol. 2019, 2019.

\bibitem{61}
S.~R. Ganti and Y.~Kim, ``Design of low-cost on-board auto-tracking antenna for
  small uas,'' in \emph{2015 12th International Conference on Information
  Technology-New Generations}.\hskip 1em plus 0.5em minus 0.4em\relax IEEE,
  2015, pp. 273--279.

\bibitem{62}
A.~F.~A. Carneiro, J.~P.~N. Torres, A.~Baptista, and M.~J.~M. Martins, ``Smart
  antenna for application in uavs,'' \emph{Information}, vol.~9, no.~12, p.
  328, 2018.

\bibitem{64}
A.~Abdelmaboud, ``The internet of drones: Requirements, taxonomy, recent
  advances, and challenges of research trends,'' \emph{Sensors}, vol.~21,
  no.~17, p. 5718, 2021.

\bibitem{65}
N.~Goyal, S.~Sharma, A.~K. Rana, and S.~L. Tripathi, \emph{Internet of Things:
  Robotic and Drone Technology}.\hskip 1em plus 0.5em minus 0.4em\relax CRC
  Press, 2022.

\bibitem{67}
A.~Israr, G.~E.~M. Abro, M.~Sadiq Ali~Khan, M.~Farhan, and S.~u.~A. Bin
  Mohd~Zulkifli, ``Internet of things (iot)-enabled unmanned aerial vehicles
  for the inspection of construction sites: a vision and future directions,''
  \emph{Mathematical Problems in Engineering}, vol. 2021, pp. 1--15, 2021.

\bibitem{68}
Y.~Aldeen and H.~M. Abdulhadi, ``Data communication for drone-enabled internet
  of things,'' \emph{Indonesian Journal of Electrical Engineering and Computer
  Science}, vol.~22, no.~2, pp. 1216--1222, 2021.

\bibitem{72}
Y.~Sun, H.~Fesenko, V.~Kharchenko, L.~Zhong, I.~Kliushnikov, O.~Illiashenko,
  O.~Morozova, and A.~Sachenko, ``Uav and iot-based systems for the monitoring
  of industrial facilities using digital twins: Methodology, reliability
  models, and application,'' \emph{Sensors}, vol.~22, no.~17, p. 6444, 2022.

\bibitem{74}
S.~A. Lakshman and D.~Ebenezer, ``Integration of internet of things and drones
  and its future applications,'' \emph{Materials Today: Proceedings}, vol.~47,
  pp. 944--949, 2021.

\bibitem{88}
S.~Shen, Y.~Mulgaonkar, N.~Michael, and V.~Kumar, ``Multi-sensor fusion for
  robust autonomous flight in indoor and outdoor environments with a rotorcraft
  mav,'' in \emph{2014 IEEE International Conference on Robotics and Automation
  (ICRA)}.\hskip 1em plus 0.5em minus 0.4em\relax IEEE, 2014, pp. 4974--4981.

\bibitem{94}
L.~Xin, Z.~Tang, W.~Gai, and H.~Liu, ``Vision-based autonomous landing for the
  uav: A review,'' \emph{Aerospace}, vol.~9, no.~11, p. 634, 2022.

\bibitem{89}
S.~Y. Choi and D.~Cha, ``Unmanned aerial vehicles using machine learning for
  autonomous flight; state-of-the-art,'' \emph{Advanced Robotics}, vol.~33,
  no.~6, pp. 265--277, 2019.

\bibitem{90}
M.~Y. B.~M. Noor, M.~Ismail, M.~F. b~Khyasudeen, A.~Shariffuddin, N.~Kamel, and
  S.~R. Azzuhri, ``Autonomous precision landing for commercial uav: A review.''
  \emph{FSDM}, pp. 459--468, 2017.

\bibitem{91}
A.~Zulu and S.~John, ``A review of control algorithms for autonomous
  quadrotors,'' \emph{arXiv preprint arXiv:1602.02622}, 2016.

\bibitem{93}
Z.~Li, W.-H. Chen, and C.~Liu, ``Review of uav-based autonomous search
  algorithms for hazardous sources.''

\bibitem{99}
S.~Crommelinck, R.~Bennett, M.~Gerke, F.~Nex, M.~Y. Yang, and G.~Vosselman,
  ``Review of automatic feature extraction from high-resolution optical sensor
  data for uav-based cadastral mapping,'' \emph{Remote Sensing}, vol.~8, no.~8,
  p. 689, 2016.

\bibitem{100}
F.~Corradi and F.~Fioranelli, ``Radar perception for autonomous unmanned aerial
  vehicles: a survey,'' \emph{System Engineering for constrained embedded
  systems}, pp. 14--20, 2022.

\bibitem{104}
Z.~Sun, X.~Wang, Z.~Wang, L.~Yang, Y.~Xie, and Y.~Huang, ``Uavs as remote
  sensing platforms in plant ecology: review of applications and challenges,''
  \emph{Journal of Plant Ecology}, vol.~14, no.~6, pp. 1003--1023, 2021.

\bibitem{105}
S.~Guan, Z.~Zhu, and G.~Wang, ``A review on uav-based remote sensing
  technologies for construction and civil applications,'' \emph{Drones},
  vol.~6, no.~5, p. 117, 2022.

\bibitem{107}
H.~Ren, Y.~Zhao, W.~Xiao, and Z.~Hu, ``A review of uav monitoring in mining
  areas: Current status and future perspectives,'' \emph{International Journal
  of Coal Science \& Technology}, vol.~6, pp. 320--333, 2019.

\bibitem{108}
R.~Bailon-Ruiz and S.~Lacroix, ``Wildfire remote sensing with uavs: A review
  from the autonomy point of view,'' in \emph{2020 international conference on
  unmanned aircraft systems (ICUAS)}.\hskip 1em plus 0.5em minus 0.4em\relax
  IEEE, 2020, pp. 412--420.

\bibitem{109}
L.~Merino, F.~Caballero, J.~Ferruz, J.~Wiklund, P.-E. Forss{\'e}n, and
  A.~Ollero, ``Multi-uav cooperative perception techniques,'' \emph{Multiple
  heterogeneous unmanned aerial vehicles}, pp. 67--110, 2007.

\bibitem{110}
M.~Chodnicki, B.~Siemiatkowska, W.~Stecz, and S.~St{\k{e}}pie{\'n}, ``Energy
  efficient uav flight control method in an environment with obstacles and
  gusts of wind,'' \emph{Energies}, vol.~15, no.~10, p. 3730, 2022.

\bibitem{112}
M.~N. Boukoberine, Z.~Zhou, and M.~Benbouzid, ``A critical review on unmanned
  aerial vehicles power supply and energy management: Solutions, strategies,
  and prospects,'' \emph{Applied Energy}, vol. 255, p. 113823, 2019.

\bibitem{124}
C.~Amici, F.~Ceresoli, M.~Pasetti, M.~Saponi, M.~Tiboni, and S.~Zanoni,
  ``Review of propulsion system design strategies for unmanned aerial
  vehicles,'' \emph{Applied Sciences}, vol.~11, no.~11, p. 5209, 2021.

\bibitem{115}
O.~Gur and A.~Rosen, ``Optimizing electric propulsion systems for unmanned
  aerial vehicles,'' \emph{Journal of aircraft}, vol.~46, no.~4, pp.
  1340--1353, 2009.

\bibitem{116}
N.~J.~P. Betancourth, J.~E.~P. Villamarin, J.~J.~V. Rios, P.~D. Bravo-Mosquera,
  and H.~D. Cer{\'o}n-Mu{\~n}oz, ``Design and manufacture of a solar-powered
  unmanned aerial vehicle for civilian surveillance missions,'' \emph{Journal
  of Aerospace Technology and Management}, vol.~8, pp. 385--396, 2016.

\bibitem{117}
P.~Rajendran and H.~Smith, ``Review of solar and battery power system
  development for solar-powered electric unmanned aerial vehicles,'' in
  \emph{Advanced Materials Research}, vol. 1125.\hskip 1em plus 0.5em minus
  0.4em\relax Trans Tech Publ, 2015, pp. 641--647.

\bibitem{119}
I.~A. Nemer, T.~R. Sheltami, S.~Belhaiza, and A.~S. Mahmoud, ``Energy-efficient
  uav movement control for fair communication coverage: A deep reinforcement
  learning approach,'' \emph{Sensors}, vol.~22, no.~5, p. 1919, 2022.

\bibitem{120}
T.~Xiao, W.~Wei, H.~Hongliang, and R.~Zhang, ``Energy-efficient data collection
  for uav-assisted iot: Joint trajectory and resource optimization,''
  \emph{Chinese Journal of Aeronautics}, vol.~35, no.~9, pp. 95--105, 2022.

\bibitem{121}
H.~T. Do, L.~H. Truong, M.~T. Nguyen, C.-F. Chien, H.~T. Tran, H.~T. Hua, C.~V.
  Nguyen, H.~T. Nguyen, and N.~T. Nguyen, ``Energy-efficient unmanned aerial
  vehicle (uav) surveillance utilizing artificial intelligence (ai),''
  \emph{Wireless Communications and Mobile Computing}, vol. 2021, pp. 1--11,
  2021.

\bibitem{125}
R.~Jiao, Z.~Wang, R.~Chu, M.~Dong, Y.~Rong, and W.~Chou, ``An intuitive
  end-to-end human-uav interaction system for field exploration,''
  \emph{Frontiers in Neurorobotics}, vol.~13, p. 117, 2020.

\bibitem{128}
M.~A. Kassab, M.~Ahmed, A.~Maher, and B.~Zhang, ``Real-time human-uav
  interaction: New dataset and two novel gesture-based interacting systems,''
  \emph{IEEE Access}, vol.~8, pp. 195\,030--195\,045, 2020.

\bibitem{126}
T.~M{\"u}ezzino{\u{g}}lu and M.~Karak{\"o}se, ``An intelligent human--unmanned
  aerial vehicle interaction approach in real time based on machine learning
  using wearable gloves,'' \emph{Sensors}, vol.~21, no.~5, p. 1766, 2021.

\bibitem{127}
B.~Chen, C.~Hua, D.~Li, Y.~He, and J.~Han, ``Intelligent human--uav interaction
  system with joint cross-validation over action--gesture recognition and scene
  understanding,'' \emph{Applied Sciences}, vol.~9, no.~16, p. 3277, 2019.

\bibitem{129}
A.~Maher, C.~Li, H.~Hu, and B.~Zhang, ``Realtime human-uav interaction using
  deep learning,'' in \emph{Biometric Recognition: 12th Chinese Conference,
  CCBR 2017, Shenzhen, China, October 28-29, 2017, Proceedings 12}.\hskip 1em
  plus 0.5em minus 0.4em\relax Springer, 2017, pp. 511--519.

\bibitem{130}
S.~Rajappa, H.~B{\"u}lthoff, and P.~Stegagno, ``Design and implementation of a
  novel architecture for physical human-uav interaction,'' \emph{The
  International Journal of Robotics Research}, vol.~36, no. 5-7, pp. 800--819,
  2017.

\bibitem{131}
S.~Rajappa \emph{et~al.}, \emph{Towards Human-UAV Physical Interaction and
  Fully Actuated Aerial Vehicles}.\hskip 1em plus 0.5em minus 0.4em\relax Logos
  Verlag, 2018.

\bibitem{134}
T.~Nisser and C.~Westin, ``Human factors challenges in unmanned aerial vehicles
  (uavs): A literature review,'' \emph{School of Aviation of the Lund
  University, Ljungbyhed}, 2006.

\bibitem{135}
S.~Hart, V.~Steane, S.~Bullock, and J.~M. Noyes, ``Understanding human
  decision-making when controlling uavs in a search and rescue application,''
  2022.

\bibitem{133}
D.~Tezza and M.~Andujar, ``The state-of-the-art of human--drone interaction: A
  survey,'' \emph{IEEE Access}, vol.~7, pp. 167\,438--167\,454, 2019.

\bibitem{136}
Y.~Zhou, B.~Rao, and W.~Wang, ``Uav swarm intelligence: Recent advances and
  future trends,'' \emph{Ieee Access}, vol.~8, pp. 183\,856--183\,878, 2020.

\bibitem{150}
M.~Khelifi and I.~Butun, ``Swarm unmanned aerial vehicles (suavs): A
  comprehensive analysis of localization, recent aspects, and future trends,''
  \emph{Journal of Sensors}, vol. 2022, 2022.

\bibitem{140}
M.~Campion, P.~Ranganathan, and S.~Faruque, ``A review and future directions of
  uav swarm communication architectures,'' in \emph{2018 IEEE international
  conference on electro/information technology (EIT)}.\hskip 1em plus 0.5em
  minus 0.4em\relax IEEE, 2018, pp. 0903--0908.

\bibitem{141}
M.~Paulsson, ``High-level control of uav swarms with rssi based position
  estimation,'' 2017.

\bibitem{151}
V.~K. KA, R.~Priyadarshini, P.~Kathik, E.~Madhan, and A.~Sonya,
  ``Self-co-ordination algorithm (sca) for multi-uav systems using fair
  scheduling queue,'' \emph{Sensor Review}, no. ahead-of-print, 2022.

\bibitem{143}
M.~M. Iqbal, Z.~A. Ali, R.~Khan, and M.~Shafiq, ``Motion planning of uav swarm:
  Recent challenges and approaches,'' \emph{Aeronautics-New Advances}, 2022.

\bibitem{144}
X.~Zhu, Z.~Liu, and J.~Yang, ``Model of collaborative uav swarm toward
  coordination and control mechanisms study,'' \emph{Procedia Computer
  Science}, vol.~51, pp. 493--502, 2015.

\bibitem{148}
Q.~Peng, H.~Wu, and R.~Xue, ``Review of dynamic task allocation methods for uav
  swarms oriented to ground targets,'' \emph{Complex System Modeling and
  Simulation}, vol.~1, no.~3, pp. 163--175, 2021.

\bibitem{149}
S.~Qamar, S.~H. Khan, M.~A. Arshad, M.~Qamar, and A.~Khan, ``Autonomous drone
  swarm navigation and multi-target tracking in 3d environments with dynamic
  obstacles,'' \emph{arXiv preprint arXiv:2202.06253}, 2022.

\bibitem{77}
B.~Taha and A.~Shoufan, ``Machine learning-based drone detection and
  classification: State-of-the-art in research,'' \emph{IEEE access}, vol.~7,
  pp. 138\,669--138\,682, 2019.

\bibitem{79}
R.~Bhalara, R.~J. Shilu, and D.~Nandi, ``A review on aircraft detection
  techniques and feature extraction using deep learning.''

\bibitem{82}
L.~Zhou, H.~Yan, Y.~Shan, C.~Zheng, Y.~Liu, X.~Zuo, and B.~Qiao, ``Aircraft
  detection for remote sensing images based on deep convolutional neural
  networks,'' \emph{Journal of Electrical and Computer Engineering}, vol. 2021,
  pp. 1--16, 2021.

\bibitem{83}
E.~Kiyak and G.~Unal, ``Small aircraft detection using deep learning,''
  \emph{Aircraft Engineering and Aerospace Technology}, vol.~93, no.~4, pp.
  671--681, 2021.

\bibitem{84}
A.~K. Thoudoju, ``Detection of aircraft, vehicles and ships in aerial and
  satellite imagery using evolutionary deep learning,'' 2021.

\bibitem{211}
M.~F. Ferraz, L.~B. J{\'u}nior, A.~S. Komori, L.~C. Rech, G.~H. Schneider,
  G.~S. Berger, {\'A}.~R. Cantieri, J.~Lima, and M.~A. Wehrmeister,
  ``Artificial intelligence architecture based on planar lidar scan data to
  detect energy pylon structures in a uav autonomous detailed inspection
  process,'' in \emph{Optimization, Learning Algorithms and Applications: First
  International Conference, OL2A 2021, Bragan{\c{c}}a, Portugal, July 19--21,
  2021, Revised Selected Papers 1}.\hskip 1em plus 0.5em minus 0.4em\relax
  Springer, 2021, pp. 430--443.

\bibitem{himeur2022latest}
Y.~Himeur, S.~S. Sohail, F.~Bensaali, A.~Amira, and M.~Alazab, ``Latest trends
  of security and privacy in recommender systems: a comprehensive review and
  future perspectives,'' \emph{Computers \& Security}, vol. 118, p. 102746,
  2022.

\bibitem{AI2}
W.~F. Hendria, Q.~T. Phan, F.~Adzaka, and C.~Jeong, ``Combining transformer and
  cnn for object detection in uav imagery,'' \emph{ICT Express}, 2021.

\bibitem{AI3}
M.~Radovic, O.~Adarkwa, and Q.~Wang, ``Object recognition in aerial images
  using convolutional neural networks,'' \emph{Journal of Imaging}, vol.~3,
  no.~2, p.~21, 2017.

\bibitem{AI4}
M.~Rahnemoonfar, R.~Murphy, M.~V. Miquel, D.~Dobbs, and A.~Adams, ``Flooded
  area detection from uav images based on densely connected recurrent neural
  networks,'' in \emph{IGARSS 2018-2018 IEEE international geoscience and
  remote sensing symposium}.\hskip 1em plus 0.5em minus 0.4em\relax IEEE, 2018,
  pp. 1788--1791.

\bibitem{AI1}
M.~B. Bejiga, A.~Zeggada, and F.~Melgani, ``Convolutional neural networks for
  near real-time object detection from uav imagery in avalanche search and
  rescue operations,'' in \emph{2016 IEEE International Geoscience and Remote
  Sensing Symposium (IGARSS)}.\hskip 1em plus 0.5em minus 0.4em\relax IEEE,
  2016, pp. 693--696.

\bibitem{himeur2023video}
Y.~Himeur, S.~Al-Maadeed, H.~Kheddar, N.~Al-Maadeed, K.~Abualsaud, A.~Mohamed,
  and T.~Khattab, ``Video surveillance using deep transfer learning and deep
  domain adaptation: Towards better generalization,'' \emph{Engineering
  Applications of Artificial Intelligence}, vol. 119, p. 105698, 2023.

\bibitem{212}
S.~Li, Y.~Jia, F.~Yang, Q.~Qin, H.~Gao, and Y.~Zhou, ``Collaborative
  decision-making method for multi-uav based on multiagent reinforcement
  learning,'' \emph{IEEE Access}, vol.~10, pp. 91\,385--91\,396, 2022.

\bibitem{208}
S.~Rezwan and W.~Choi, ``Artificial intelligence approaches for uav navigation:
  Recent advances and future challenges,'' \emph{IEEE Access}, 2022.

\bibitem{209}
R.~Yin, W.~Li, Z.-q. Wang, and X.-x. Xu, ``The application of artificial
  intelligence technology in uav,'' in \emph{2020 5th international conference
  on information science, computer technology and transportation
  (ISCTT)}.\hskip 1em plus 0.5em minus 0.4em\relax IEEE, 2020, pp. 238--241.

\bibitem{OpenAI}
\BIBentryALTinterwordspacing
Openai chatgpt. [Online]. Available: \url{https://chat.openai.com/}
\BIBentrySTDinterwordspacing

\bibitem{sohail2023using}
S.~S. Sohail, D.~{\O}. Madsen, Y.~Himeur, and M.~Ashraf, ``Using chatgpt to
  navigate ambivalent and contradictory research findings on artificial
  intelligence,'' \emph{Available at SSRN 4413913}, 2023.

\bibitem{chatgpt3}
B.~Zhang and H.~Soh, ``Large language models as zero-shot human models for
  human-robot interaction,'' \emph{arXiv preprint arXiv:2303.03548}, 2023.

\bibitem{chatgpt4}
J.~X. Liu, Z.~Yang, I.~Idrees, S.~Liang, B.~Schornstein, S.~Tellex, and
  A.~Shah, ``Lang2ltl: Translating natural language commands to temporal robot
  task specification,'' \emph{arXiv preprint arXiv:2302.11649}, 2023.

\bibitem{sohail2023future}
S.~S. Sohail, F.~Farhat, Y.~Himeur, M.~Nadeem, D.~{\O}. Madsen, Y.~Singh,
  S.~Atalla, and W.~Mansoor, ``The future of gpt: A taxonomy of existing
  chatgpt research, current challenges, and possible future directions,''
  \emph{Current Challenges, and Possible Future Directions (April 8, 2023)},
  2023.

\bibitem{chatgpt6}
Y.~Hong, Q.~Wu, Y.~Qi, C.~Rodriguez-Opazo, and S.~Gould, ``A recurrent
  vision-and-language bert for navigation,'' \emph{arXiv preprint
  arXiv:2011.13922}, 2020.

\bibitem{chatgpt7}
A.~Bucker, L.~Figueredo, S.~Haddadinl, A.~Kapoor, S.~Ma, and R.~Bonatti,
  ``Reshaping robot trajectories using natural language commands: A study of
  multi-modal data alignment using transformers,'' in \emph{2022 IEEE/RSJ
  International Conference on Intelligent Robots and Systems (IROS)}.\hskip 1em
  plus 0.5em minus 0.4em\relax IEEE, 2022, pp. 978--984.

\bibitem{vemprala2023chatgpt}
S.~Vemprala, R.~Bonatti, A.~Bucker, and A.~Kapoor, ``Chatgpt for robotics:
  Design principles and model abilities,'' \emph{2023}, 2023.

\bibitem{chatgpt2}
\BIBentryALTinterwordspacing
Microsoft airsim chatgpt. [Online]. Available:
  \url{https://youtu.be/NYd0QcZcS6Q}
\BIBentrySTDinterwordspacing

\bibitem{chatgpt1}
\BIBentryALTinterwordspacing
Airsim-chatgpt, promptcraft code. [Online]. Available:
  \url{https://github.com/microsoft/PromptCraft-Robotics}
\BIBentrySTDinterwordspacing

\bibitem{GPT-Drone2}
\BIBentryALTinterwordspacing
Microsoft airsim chatgpt for industrial inspection. [Online]. Available:
  \url{https://www.youtube.com/watch?v=38lA3U2J43w&feature=youtu.be}
\BIBentrySTDinterwordspacing

\bibitem{tiwary2021monitoring}
A.~Tiwary, B.~Rimal, Y.~Himeur, and A.~Amira, ``Monitoring nature-based
  engineering projects in mountainous region incorporating spatial imaging:
  Case study of a hydroelectric project in nepal,'' in \emph{CITIES
  20.50--Creating Habitats for the 3rd Millennium: Smart--Sustainable--Climate
  Neutral. Proceedings of REAL CORP 2021, 26th International Conference on
  Urban Development, Regional Planning and Information Society}.\hskip 1em plus
  0.5em minus 0.4em\relax CORP--Competence Center of Urban and Regional
  Planning, 2021, pp. 535--538.

\bibitem{31}
J.~G. Serna, F.~Vanegas, F.~Gonzalez, and D.~Flannery, ``A review of current
  approaches for uav autonomous mission planning for mars biosignatures
  detection,'' in \emph{2020 IEEE Aerospace Conference}.\hskip 1em plus 0.5em
  minus 0.4em\relax IEEE, 2020, pp. 1--15.

\bibitem{himeur2022ai}
Y.~Himeur, M.~Elnour, F.~Fadli, N.~Meskin, I.~Petri, Y.~Rezgui, F.~Bensaali,
  and A.~Amira, ``Ai-big data analytics for building automation and management
  systems: a survey, actual challenges and future perspectives,''
  \emph{Artificial Intelligence Review}, pp. 1--93, 2022.

\bibitem{wojcik2019investigation}
K.~A. W{\'o}jcik, R.~J. Bialik, M.~Osi{\'n}ska, and M.~Figielski,
  ``Investigation of sediment-rich glacial meltwater plumes using a
  high-resolution multispectral sensor mounted on an unmanned aerial vehicle,''
  \emph{Water}, vol.~11, no.~11, p. 2405, 2019.

\bibitem{220}
A.~Bauranov and J.~Rakas, ``Designing airspace for urban air mobility: A review
  of concepts and approaches,'' \emph{Progress in Aerospace Sciences}, vol.
  125, p. 100726, 2021.

\bibitem{221}
L.~A. Garrow, B.~J. German, and C.~E. Leonard, ``Urban air mobility: A
  comprehensive review and comparative analysis with autonomous and electric
  ground transportation for informing future research,'' \emph{Transportation
  Research Part C: Emerging Technologies}, vol. 132, p. 103377, 2021.

\bibitem{222}
R.~Rothfeld, A.~Straubinger, M.~Fu, C.~Al~Haddad, and C.~Antoniou, ``Urban air
  mobility,'' in \emph{Demand for Emerging Transportation Systems}.\hskip 1em
  plus 0.5em minus 0.4em\relax Elsevier, 2020, pp. 267--284.

\bibitem{223}
R.~Shrestha, I.~Oh, and S.~Kim, ``A survey on operation concept, advancements,
  and challenging issues of urban air traffic management,'' \emph{Frontiers in
  Future Transportation}, vol.~2, p.~1, 2021.

\bibitem{33}
D.~Di~Giovanni, F.~Fumian, A.~Chierici, M.~Bianchelli, L.~Martellucci,
  G.~Carminati, A.~Malizia, F.~d’Errico, P.~Gaudio \emph{et~al.}, ``Design of
  miniaturized sensors for a mission-oriented uav application: A new pathway
  for early warning,'' \emph{INTERNATIONAL JOURNAL OF SAFETY AND SECURITY
  ENGINEERING}, vol.~11, no.~4, pp. 435--444, 2021.

\bibitem{232}
S.~Matalonga, S.~White, J.~Hartmann, and J.~Riordan, ``A review of the legal,
  regulatory and practical aspects needed to unlock autonomous beyond visual
  line of sight unmanned aircraft systems operations,'' \emph{Journal of
  Intelligent \& Robotic Systems}, vol. 106, no.~1, p.~10, 2022.

\bibitem{233}
M.~Skowron, W.~Chmielowiec, K.~Glowacka, M.~Krupa, and A.~Srebro, ``Sense and
  avoid for small unmanned aircraft systems: Research on methods and best
  practices,'' \emph{Proceedings of the Institution of Mechanical Engineers,
  Part G: Journal of Aerospace Engineering}, vol. 233, no.~16, pp. 6044--6062,
  2019.

\bibitem{236}
R.~J. a.~L. Hartley, I.~L. Henderson, and C.~L. Jackson, ``Bvlos unmanned
  aircraft operations in forest environments,'' \emph{Drones}, vol.~6, no.~7,
  p. 167, 2022.

\bibitem{240}
J.~Gray, ``Design optimization of a boundary layer ingestion propulsor using a
  coupled aeropropulsive model,'' Ph.D. dissertation, 2018.

\bibitem{241}
Y.~Chen, ``Overview of solar uav power system,'' \emph{Academic Journal of
  Science and Technology}, vol.~4, no.~1, pp. 80--82, 2022.

\bibitem{242}
S.~Pal, A.~Mishra, and P.~Singh, ``Recent trends in long endurance solar
  powered uavs: A review,'' 2020.

\bibitem{243}
M.~Harun, S.~Abdullah, M.~Aras, and M.~Bahar, ``Collision avoidance control for
  unmanned autonomous vehicles (uav): Recent advancements and future
  prospects,'' 2021.

\bibitem{245}
X.~Gan, Y.~Wu, P.~Liu, and Q.~Wang, ``Dynamic collision avoidance zone modeling
  method based on uav emergency collision avoidance trajectory,'' in \emph{2020
  IEEE International Conference on Artificial Intelligence and Information
  Systems (ICAIIS)}.\hskip 1em plus 0.5em minus 0.4em\relax IEEE, 2020, pp.
  693--696.

\bibitem{246}
E.~Kakaletsis, C.~Symeonidis, M.~Tzelepi, I.~Mademlis, A.~Tefas, N.~Nikolaidis,
  and I.~Pitas, ``Computer vision for autonomous uav flight safety: an overview
  and a vision-based safe landing pipeline example,'' \emph{Acm Computing
  Surveys (Csur)}, vol.~54, no.~9, pp. 1--37, 2021.

\bibitem{253}
G.~S. Hadi, R.~Varianto, B.~Trilaksono, and A.~Budiyono, ``Autonomous uav
  system development for payload dropping mission,'' \emph{The Journal of
  Instrumentation, Automation and Systems}, vol.~1, no.~2, pp. 72--22, 2014.

\bibitem{312}
P.~J. Cruz and R.~Fierro, ``Cable-suspended load lifting by a quadrotor uav:
  hybrid model, trajectory generation, and control,'' \emph{Autonomous Robots},
  vol.~41, pp. 1629--1643, 2017.

\bibitem{313}
M.~A. Santos, B.~Rego, G.~V. Raffo, and A.~Ferramosca, ``Suspended load path
  tracking control strategy using a tilt-rotor uav,'' \emph{Journal of Advanced
  Transportation}, vol. 2017, 2017.

\bibitem{314}
K.~Mohammadi, ``Passivity-based control of multiple quad-copters with a
  cable-suspended payload,'' Ph.D. dissertation, 2021.

\bibitem{256}
S.~H. Derrouaoui, Y.~Bouzid, M.~Guiatni, and I.~Dib, ``A comprehensive review
  on reconfigurable drones: Classification, characteristics, design and control
  technologies,'' \emph{Unmanned Systems}, vol.~10, no.~01, pp. 3--29, 2022.

\bibitem{259}
A.~Moosavian, F.~Xi, and S.~M. Hashemi, ``Design and motion control of fully
  variable morphing wings,'' \emph{Journal of aircraft}, vol.~50, no.~4, pp.
  1189--1201, 2013.

\bibitem{261}
E.~Ajanic, M.~Feroskhan, S.~Mintchev, F.~Noca, and D.~Floreano, ``Bioinspired
  wing and tail morphing extends drone flight capabilities,'' \emph{Science
  Robotics}, vol.~5, no.~47, p. eabc2897, 2020.

\bibitem{263}
H.~Jiakun, H.~Zhe, T.~Fangbao, and C.~Gang, ``Review on bio-inspired flight
  systems and bionic aerodynamics,'' \emph{Chinese Journal of Aeronautics},
  vol.~34, no.~7, pp. 170--186, 2021.

\bibitem{264}
L.~Kilian, F.~Shahid, J.-S. Zhao, and C.~N. Nayeri, ``Bioinspired morphing
  wings: mechanical design and wind tunnel experiments,'' \emph{Bioinspiration
  \& Biomimetics}, vol.~17, no.~4, p. 046019, 2022.

\bibitem{267}
E.~J. Miller, W.~A. Lokos, J.~Cruz, G.~Crampton, C.~A. Stephens, S.~Kota,
  G.~Ervin, and P.~Flick, ``Approach for structurally clearing an adaptive
  compliant trailing edge flap for flight,'' in \emph{Society of Flight Test
  Engineers International Annual Symposium}, no. DFRC-E-DAA-TN24640, 2015.

\bibitem{frigioescu2023preliminary}
T.-F. Frigioescu, M.~R. Condruz, T.~A. Badea, and A.~Paraschiv, ``A preliminary
  study on the development of a new uav concept and the associated flight
  method,'' \emph{Drones}, vol.~7, no.~3, p. 166, 2023.

\bibitem{gokbel2023improvement}
E.~G{\"o}kbel, A.~G{\"u}ll{\"u}, and S.~Ersoy, ``Improvement of uav: design and
  implementation on launchability,'' \emph{Aircraft Engineering and Aerospace
  Technology}, vol.~95, no.~5, pp. 734--740, 2023.

\bibitem{defrangesco2022big}
R.~DeFrangesco and S.~DeFrangesco, \emph{The Big Book of Drones}.\hskip 1em
  plus 0.5em minus 0.4em\relax CRC Press, 2022.

\bibitem{268}
D.~Falanga, K.~Kleber, S.~Mintchev, D.~Floreano, and D.~Scaramuzza, ``The
  foldable drone: A morphing quadrotor that can squeeze and fly,'' \emph{IEEE
  Robotics and Automation Letters}, vol.~4, no.~2, pp. 209--216, 2018.

\bibitem{272}
\BIBentryALTinterwordspacing
The foldable drone, laboratory of intelligent systems at epfl,. [Online].
  Available:
  \url{https://actu.epfl.ch/news/new-foldable-drone-flies-through-narrow-holes-in-r}
\BIBentrySTDinterwordspacing

\bibitem{273}
M.~Pods{\k{e}}dkowski, R.~Konopi{\'n}ski, D.~Obidowski, and K.~Koter,
  ``Variable pitch propeller for uav-experimental tests,'' \emph{Energies},
  vol.~13, no.~20, p. 5264, 2020.

\bibitem{274}
A.~Abhishek, A.~Duhoon, M.~Kothari, S.~Kadukar, L.~Rane, and G.~Suryavanshi,
  ``Design, development, and closed-loop flight-testing of a single power plant
  variable pitch quadrotor unmanned air vehicle,'' in \emph{Proceedings of the
  73rd American Helicopter Society Annual Forum}, 2017, pp. 205--218.

\bibitem{275}
X.~Wu, ``Design and development of variable pitch quadcopter for long endurance
  flight,'' Ph.D. dissertation, Oklahoma State University, 2018.

\bibitem{276}
Z.~Liu, Y.~He, L.~Yang, and J.~Han, ``Control techniques of tilt rotor unmanned
  aerial vehicle systems: A review,'' \emph{Chinese Journal of Aeronautics},
  vol.~30, no.~1, pp. 135--148, 2017.

\bibitem{277}
A.~Misra, S.~Jayachandran, S.~Kenche, A.~Katoch, A.~Suresh, E.~Gundabattini,
  S.~K. Selvaraj, A.~A. Legesse \emph{et~al.}, ``A review on vertical take-off
  and landing (vtol) tilt-rotor and tilt wing unmanned aerial vehicles
  (uavs),'' \emph{Journal of Engineering}, vol. 2022, 2022.

\bibitem{280}
M.~Kamel, S.~Verling, O.~Elkhatib, C.~Sprecher, P.~Wulkop, Z.~Taylor,
  R.~Siegwart, and I.~Gilitschenski, ``The voliro omniorientational hexacopter:
  An agile and maneuverable tiltable-rotor aerial vehicle,'' \emph{IEEE
  Robotics \& Automation Magazine}, vol.~25, no.~4, pp. 34--44, 2018.

\bibitem{281}
\BIBentryALTinterwordspacing
Voliro hexcopter, eth zurich team. [Online]. Available:
  \url{https://voliro.com}
\BIBentrySTDinterwordspacing

\bibitem{282}
\BIBentryALTinterwordspacing
Gl-10, nasa greased lightning. [Online]. Available:
  \url{https://ntrs.nasa.gov/citations/20180000765}
\BIBentrySTDinterwordspacing

\bibitem{284}
V.~Kangunde, R.~S. Jamisola, and E.~K. Theophilus, ``A review on drones
  controlled in real-time,'' \emph{International journal of dynamics and
  control}, pp. 1--15, 2021.

\bibitem{285}
B.~Rub{\'\i}, R.~P{\'e}rez, and B.~Morcego, ``A survey of path following
  control strategies for uavs focused on quadrotors,'' \emph{Journal of
  Intelligent \& Robotic Systems}, vol.~98, no.~2, pp. 241--265, 2020.

\bibitem{290}
G.~Sonugur, ``A review of quadrotor uav: Control and slam methodologies ranging
  from conventional to innovative approaches,'' \emph{Robotics and Autonomous
  Systems}, p. 104342, 2022.

\bibitem{291}
A.~S. Hanif, X.~Han, and S.-H. Yu, ``Independent control spraying system for
  uav-based precise variable sprayer: A review,'' \emph{Drones}, vol.~6,
  no.~12, p. 383, 2022.

\bibitem{294}
M.~Idrissi, M.~Salami, and F.~Annaz, ``A review of quadrotor unmanned aerial
  vehicles: applications, architectural design and control algorithms,''
  \emph{Journal of Intelligent \& Robotic Systems}, vol. 104, no.~2, p.~22,
  2022.

\bibitem{311}
Y.~{\'A}. L{\'o}pez, M.~Garcia-Fernandez, G.~Alvarez-Narciandi, and F.~L.-H.
  Andr{\'e}s, ``Unmanned aerial vehicle-based ground-penetrating radar systems:
  A review,'' \emph{IEEE Geoscience and Remote Sensing Magazine}, vol.~10,
  no.~2, pp. 66--86, 2022.

\bibitem{297}
G.~Farid, M.~Hongwei, S.~M. Ali, and Q.~Liwei, ``A review on linear and
  nonlinear control techniques for position and attitude control of a
  quadrotor,'' \emph{Control and Intelligent Systems}, vol.~45, no.~1, pp.
  43--57, 2017.

\bibitem{298}
S.~H. Derrouaoui, Y.~Bouzid, and M.~Guiatni, ``Nonlinear robust control of a
  new reconfigurable unmanned aerial vehicle,'' \emph{Robotics}, vol.~10,
  no.~2, p.~76, 2021.

\bibitem{299}
P.~Priya and S.~S. Kamlu, ``Robust control algorithm for drones,'' in
  \emph{Aeronautics-New Advances}.\hskip 1em plus 0.5em minus 0.4em\relax
  IntechOpen, 2022.

\bibitem{300}
S.~Cao and H.~Yu, ``An adaptive control framework for the autonomous aerobatic
  maneuvers of fixed-wing unmanned aerial vehicle,'' \emph{Drones}, vol.~6,
  no.~11, p. 316, 2022.

\bibitem{301}
V.~Nguen, A.~Putov, and T.~Nguen, ``Adaptive control of an unmanned aerial
  vehicle,'' in \emph{AIP Conference Proceedings}, vol. 1798, no.~1.\hskip 1em
  plus 0.5em minus 0.4em\relax AIP Publishing LLC, 2017, p. 020124.

\bibitem{303}
T.~N. Dief and S.~Yoshida, ``Modeling and classical controller of quad-rotor,''
  \emph{arXiv preprint arXiv:1707.04173}, 2017.

\bibitem{304}
B.~Dolega, G.~Kopecki, D.~Kordos, and T.~Rogalski, ``Review of chosen control
  algorithms used for small uav control,'' in \emph{Solid State Phenomena},
  vol. 260.\hskip 1em plus 0.5em minus 0.4em\relax Trans Tech Publ, 2017, pp.
  175--183.

\bibitem{306}
A.~Joukhadar, M.~Alchehabi, and A.~Jejeh, ``Advanced uavs nonlinear control
  systems and applications,'' in \emph{Unmanned Robotic Systems and
  Applications}.\hskip 1em plus 0.5em minus 0.4em\relax IntechOpen, 2019,
  p.~79.

\bibitem{307}
M.~Jafari and H.~Xu, ``Intelligent control for unmanned aerial systems with
  system uncertainties and disturbances using artificial neural network,''
  \emph{Drones}, vol.~2, no.~3, p.~30, 2018.

\bibitem{308}
G.~Vachtsevanos, L.~Tang, G.~Drozeski, and L.~Gutierrez, ``Intelligent control
  of unmanned aerial vehicles for improved autonomy and reliability,''
  \emph{IFAC Proceedings Volumes}, vol.~37, no.~8, pp. 852--861, 2004.

\bibitem{309}
A.~T. Azar, A.~Koubaa, N.~Ali~Mohamed, H.~A. Ibrahim, Z.~F. Ibrahim, M.~Kazim,
  A.~Ammar, B.~Benjdira, A.~M. Khamis, I.~A. Hameed \emph{et~al.}, ``Drone deep
  reinforcement learning: A review,'' \emph{Electronics}, vol.~10, no.~9, p.
  999, 2021.

\bibitem{414}
A.~Goel, J.~A. Paredes, H.~Dadhaniya, S.~A.~U. Islam, A.~M. Salim, S.~Ravela,
  and D.~Bernstein, ``Experimental implementation of an adaptive digital
  autopilot,'' in \emph{2021 American Control Conference (ACC)}.\hskip 1em plus
  0.5em minus 0.4em\relax IEEE, 2021, pp. 3737--3742.

\bibitem{417}
D.~Park, H.~Yu, N.~Xuan-Mung, J.~Lee, and S.~K. Hong, ``Multicopter pid
  attitude controller gain auto-tuning through reinforcement learning neural
  networks,'' in \emph{Proceedings of the 2019 2nd International Conference on
  Control and Robot Technology}, 2019, pp. 80--84.

\bibitem{421}
B.~R. Trilaksono, S.~H. Nasution, E.~B. Purwanto \emph{et~al.}, ``Design and
  implementation of hardware-in-the-loop-simulation for uav using pid control
  method,'' in \emph{2013 3rd International Conference on Instrumentation,
  Communications, Information Technology and Biomedical Engineering
  (ICICI-BME)}.\hskip 1em plus 0.5em minus 0.4em\relax IEEE, 2013, pp.
  124--130.

\bibitem{412}
M.~Bangura and R.~Mahony, ``Real-time model predictive control for
  quadrotors,'' \emph{IFAC Proceedings Volumes}, vol.~47, no.~3, pp.
  11\,773--11\,780, 2014.

\bibitem{413}
C.~A. Amadi \emph{et~al.}, ``Design and implementation of a model predictive
  control on a pixhawk flight controller.'' Ph.D. dissertation, Stellenbosch:
  Stellenbosch University, 2018.

\bibitem{415}
E.~A. Niit and W.~J. Smit, ``Integration of model reference adaptive control
  (mrac) with px4 firmware for quadcopters,'' in \emph{2017 24th International
  Conference on Mechatronics and Machine Vision in Practice (M2VIP)}.\hskip 1em
  plus 0.5em minus 0.4em\relax IEEE, 2017, pp. 1--6.

\bibitem{420}
A.~K. Yadav and P.~Gaur, ``Ai-based adaptive control and design of autopilot
  system for nonlinear uav,'' \emph{Sadhana}, vol.~39, pp. 765--783, 2014.

\bibitem{480}
E.-H. Zheng, J.-J. Xiong, and J.-L. Luo, ``Second order sliding mode control
  for a quadrotor uav,'' \emph{ISA transactions}, vol.~53, no.~4, pp.
  1350--1356, 2014.

\bibitem{481}
A.~Benallegue, A.~Mokhtari, and L.~Fridman, ``High-order sliding-mode observer
  for a quadrotor uav,'' \emph{International Journal of Robust and Nonlinear
  Control: IFAC-Affiliated Journal}, vol.~18, no. 4-5, pp. 427--440, 2008.

\bibitem{316}
P.~R. Chandler, M.~Pachter, D.~Swaroop, J.~M. Fowler, J.~K. Howlett,
  S.~Rasmussen, C.~Schumacher, and K.~Nygard, ``Complexity in uav cooperative
  control,'' in \emph{Proceedings of the 2002 American Control Conference (IEEE
  Cat. No. CH37301)}, vol.~3.\hskip 1em plus 0.5em minus 0.4em\relax IEEE,
  2002, pp. 1831--1836.

\bibitem{edwards1998sliding}
C.~Edwards and S.~Spurgeon, \emph{Sliding mode control: theory and
  applications}.\hskip 1em plus 0.5em minus 0.4em\relax Crc Press, 1998.

\bibitem{lee2007chattering}
H.~Lee and V.~I. Utkin, ``Chattering suppression methods in sliding mode
  control systems,'' \emph{Annual reviews in control}, vol.~31, no.~2, pp.
  179--188, 2007.

\bibitem{young1999control}
K.~D. Young, V.~I. Utkin, and U.~Ozguner, ``A control engineer's guide to
  sliding mode control,'' \emph{IEEE transactions on control systems
  technology}, vol.~7, no.~3, pp. 328--342, 1999.

\bibitem{tokat2015classification}
S.~Tokat, M.~S. Fadali, and O.~Eray, ``A classification and overview of sliding
  mode controller sliding surface design methods,'' \emph{Recent Advances in
  Sliding Modes: From Control to Intelligent Mechatronics}, pp. 417--439, 2015.

\bibitem{zheng2014second}
E.-H. Zheng, J.-J. Xiong, and J.-L. Luo, ``Second order sliding mode control
  for a quadrotor uav,'' \emph{ISA transactions}, vol.~53, no.~4, pp.
  1350--1356, 2014.

\bibitem{bartoszewicz2015new}
A.~Bartoszewicz and P.~Le{\'s}niewski, ``New switching and nonswitching type
  reaching laws for smc of discrete time systems,'' \emph{IEEE Transactions on
  Control Systems Technology}, vol.~24, no.~2, pp. 670--677, 2015.

\bibitem{latosinski2021non}
P.~Latosi{\'n}ski and M.~Herkt, ``Non-switching reaching law based dsmc
  strategies in the context of robustness comparison,'' in \emph{Proceedings of
  the 27th International Conference on Systems Engineering, ICSEng 2020}.\hskip
  1em plus 0.5em minus 0.4em\relax Springer, 2021, pp. 81--92.

\bibitem{choi2019unmanned}
S.~Y. Choi and D.~Cha, ``Unmanned aerial vehicles using machine learning for
  autonomous flight; state-of-the-art,'' \emph{Advanced Robotics}, vol.~33,
  no.~6, pp. 265--277, 2019.

\bibitem{khan2019unmanned}
A.~I. Khan and Y.~Al-Mulla, ``Unmanned aerial vehicle in the machine learning
  environment,'' \emph{Procedia computer science}, vol. 160, pp. 46--53, 2019.

\bibitem{ben2022uav}
S.~Ben~Aissa and A.~Ben~Letaifa, ``Uav communications with machine learning:
  challenges, applications and open issues,'' \emph{Arabian Journal for Science
  and Engineering}, vol.~47, no.~2, pp. 1559--1579, 2022.

\bibitem{gu2020uav}
W.~Gu, K.~P. Valavanis, M.~J. Rutherford, and A.~Rizzo, ``Uav model-based
  flight control with artificial neural networks: a survey,'' \emph{Journal of
  Intelligent \& Robotic Systems}, vol. 100, pp. 1469--1491, 2020.

\bibitem{cordoba2007autonomous}
G.~M.~A. Cordoba, ``Autonomous intelligent fuzzy logic guidance, and flight
  control system for the efigenia ej-1b mozart unmanned aerial vehicle uav,''
  \emph{IFAC Proceedings Volumes}, vol.~40, no.~7, pp. 31--36, 2007.

\bibitem{hajiyev2015fuzzy}
C.~Hajiyev, H.~Ersin~Soken, S.~Yenal~Vural, C.~Hajiyev, H.~E. Soken, and S.~Y.
  Vural, ``Fuzzy logic-based controller design,'' \emph{State Estimation and
  Control for Low-cost Unmanned Aerial Vehicles}, pp. 201--221, 2015.

\bibitem{kurnaz2010adaptive}
S.~Kurnaz, O.~Cetin, and O.~Kaynak, ``Adaptive neuro-fuzzy inference system
  based autonomous flight control of unmanned air vehicles,'' \emph{Expert
  systems with Applications}, vol.~37, no.~2, pp. 1229--1234, 2010.

\bibitem{sargolzaei2020control}
A.~Sargolzaei, A.~Abbaspour, and C.~D. Crane, ``Control of cooperative unmanned
  aerial vehicles: review of applications, challenges, and algorithms,''
  \emph{Optimization, Learning, and Control for Interdependent Complex
  Networks}, pp. 229--255, 2020.

\bibitem{ziquan2022review}
Y.~Ziquan, Y.~Zhang, B.~Jiang, F.~Jun, and J.~Ying, ``A review on
  fault-tolerant cooperative control of multiple unmanned aerial vehicles,''
  \emph{Chinese Journal of Aeronautics}, vol.~35, no.~1, pp. 1--18, 2022.

\bibitem{ryan2004overview}
A.~Ryan, M.~Zennaro, A.~Howell, R.~Sengupta, and J.~K. Hedrick, ``An overview
  of emerging results in cooperative uav control,'' in \emph{2004 43rd IEEE
  Conference on Decision and Control (CDC)(IEEE Cat. No. 04CH37601)},
  vol.~1.\hskip 1em plus 0.5em minus 0.4em\relax IEEE, 2004, pp. 602--607.

\bibitem{kada2020distributed}
B.~Kada, M.~Khalid, and M.~S. Shaikh, ``Distributed cooperative control of
  autonomous multi-agent uav systems using smooth control,'' \emph{Journal of
  Systems Engineering and Electronics}, vol.~31, no.~6, pp. 1297--1307, 2020.

\bibitem{334}
I.~Sadeghzadeh and Y.~Zhang, ``A review on fault-tolerant control for unmanned
  aerial vehicles (uavs),'' \emph{Infotech@ Aerospace 2011}, p. 1472, 2011.

\bibitem{330}
A.~Bondyra, M.~Ko{\l}odziejczak, R.~Kulikowski, and W.~Giernacki, ``An acoustic
  fault detection and isolation system for multirotor uav,'' \emph{Energies},
  vol.~15, no.~11, p. 3955, 2022.

\bibitem{331}
R.~Puchalski and W.~Giernacki, ``Uav fault detection methods,
  state-of-the-art,'' \emph{Drones}, vol.~6, no.~11, p. 330, 2022.

\bibitem{332}
G.~Ducard, ``Actuator fault detection in uavs, in handbook of unmanned
  aircraft,'' \emph{Handbook of Unmanned Aerial Vehicles}, 2014.

\bibitem{333}
G.~K. Fourlas and G.~C. Karras, ``A survey on fault diagnosis and
  fault-tolerant control methods for unmanned aerial vehicles,''
  \emph{Machines}, vol.~9, no.~9, p. 197, 2021.

\bibitem{fourlas2021survey}
------, ``A survey on fault diagnosis and fault-tolerant control methods for
  unmanned aerial vehicles,'' \emph{Machines}, vol.~9, no.~9, p. 197, 2021.

\bibitem{gao2022adaptive}
B.~Gao, Y.-J. Liu, and L.~Liu, ``Adaptive neural fault-tolerant control of a
  quadrotor uav via fast terminal sliding mode,'' \emph{Aerospace Science and
  Technology}, p. 107818, 2022.

\bibitem{bu2023prescribed}
X.~Bu, ``Prescribed performance control approaches, applications and
  challenges: A comprehensive survey,'' \emph{Asian Journal of Control},
  vol.~25, no.~1, pp. 241--261, 2023.

\bibitem{song2015integrated}
H.~Song, T.~Zhang, G.~Zhang, and C.~Lu, ``Integrated interceptor guidance and
  control with prescribed performance,'' \emph{International Journal of Robust
  and Nonlinear Control}, vol.~25, no.~16, pp. 3179--3194, 2015.

\bibitem{huang2022prescribed}
H.~Huang, C.~Luo, and B.~Han, ``Prescribed performance fuzzy back-stepping
  control of a flexible air-breathing hypersonic vehicle subject to input
  constraints,'' \emph{Journal of Intelligent Manufacturing}, vol.~33, no.~3,
  pp. 853--866, 2022.

\bibitem{chen2020adaptive}
B.-W. Chen and L.-G. Tan, ``Adaptive anti-saturation tracking control with
  prescribed performance for hypersonic vehicle,'' \emph{International Journal
  of Control, Automation and Systems}, vol.~18, no.~2, pp. 394--404, 2020.

\bibitem{bu2018prescribed}
X.~Bu and Y.~Xiao, ``Prescribed performance-based low-computational cost fuzzy
  control of a hypersonic vehicle using non-affine models,'' \emph{Advances in
  Mechanical Engineering}, vol.~10, no.~2, p. 1687814018757261, 2018.

\bibitem{bu2017prescribed}
X.~Bu, Y.~Xiao, and K.~Wang, ``A prescribed performance control approach
  guaranteeing small overshoot for air-breathing hypersonic vehicles via neural
  approximation,'' \emph{Aerospace Science and Technology}, vol.~71, pp.
  485--498, 2017.

\bibitem{li2020adaptive}
S.~Li, T.~Ma, X.~Luo, and Z.~Yang, ``Adaptive fuzzy output regulation for
  unmanned surface vehicles with prescribed performance,'' \emph{International
  Journal of Control, Automation and Systems}, vol.~18, pp. 405--414, 2020.

\bibitem{yu2020decentralized}
Z.~Yu, Y.~Zhang, Z.~Liu, Y.~Qu, C.-Y. Su, and B.~Jiang, ``Decentralized
  finite-time adaptive fault-tolerant synchronization tracking control for
  multiple uavs with prescribed performance,'' \emph{Journal of the Franklin
  Institute}, vol. 357, no.~16, pp. 11\,830--11\,862, 2020.

\bibitem{koksal2020backstepping}
N.~Koksal, H.~An, and B.~Fidan, ``Backstepping-based adaptive control of a
  quadrotor uav with guaranteed tracking performance,'' \emph{ISA
  transactions}, vol. 105, pp. 98--110, 2020.

\bibitem{16}
B.~Han, Y.~Zhou, K.~K. Deveerasetty, and C.~Hu, ``A review of control
  algorithms for quadrotor,'' in \emph{2018 IEEE international conference on
  information and automation (ICIA)}.\hskip 1em plus 0.5em minus 0.4em\relax
  IEEE, 2018, pp. 951--956.

\bibitem{amin2016review}
R.~Amin, L.~Aijun, and S.~Shamshirband, ``A review of quadrotor uav: control
  methodologies and performance evaluation,'' \emph{International Journal of
  Automation and Control}, vol.~10, no.~2, pp. 87--103, 2016.

\bibitem{roy2021review}
R.~Roy, M.~Islam, N.~Sadman, M.~P. Mahmud, K.~D. Gupta, and M.~M. Ahsan, ``A
  review on comparative remarks, performance evaluation and improvement
  strategies of quadrotor controllers,'' \emph{Technologies}, vol.~9, no.~2,
  p.~37, 2021.

\bibitem{hasseni2021parameters}
S.-E.-I. Hasseni, L.~Abdou, and H.-E. Glida, ``Parameters tuning of a quadrotor
  pid controllers by using nature-inspired algorithms,'' \emph{Evolutionary
  Intelligence}, vol.~14, pp. 61--73, 2021.

\bibitem{hasseni2018decentralized}
S.-E.-I. Hasseni and L.~Abdou, ``Decentralized pid control by using ga
  optimization applied to a quadrotor,'' \emph{Journal of Automation Mobile
  Robotics and Intelligent Systems}, vol.~12, no.~2, pp. 33--44, 2018.

\bibitem{castillo2018comparison}
J.~J. Castillo-Zamora, K.~A. Camarillo-Gomez, G.~I. Perez-Soto, and
  J.~Rodriguez-Resendiz, ``Comparison of pd, pid and sliding-mode position
  controllers for v--tail quadcopter stability,'' \emph{Ieee Access}, vol.~6,
  pp. 38\,086--38\,096, 2018.

\bibitem{10110876}
X.~Guo, S.~Hou, P.~Niu, and D.~Zhao, ``A review of control methods for
  quadrotor uavs,'' in \emph{2022 5th International Conference on Electronics
  and Electrical Engineering Technology (EEET)}, 2022, pp. 132--138.

\bibitem{okasha2022design}
M.~Okasha, J.~Kralev, and M.~Islam, ``Design and experimental comparison of
  pid, lqr and mpc stabilizing controllers for parrot mambo mini-drone.
  aerospace 2022, 9, 298,'' 2022.

\bibitem{masse2018modeling}
C.~MASSÉ, O.~GOUGEON, D.-T. NGUYEN, and D.~SAUSSIÉ, ``Modeling and control of
  a quadcopter flying in a wind field: A comparison between lqr and structured
  $\mathscr{H}$$\infty$ control techniques,'' in \emph{2018 International
  Conference on Unmanned Aircraft Systems (ICUAS)}, 2018, pp. 1408--1417.

\bibitem{rugh2000research}
W.~J. Rugh and J.~S. Shamma, ``Research on gain scheduling,''
  \emph{Automatica}, vol.~36, no.~10, pp. 1401--1425, 2000.

\bibitem{bouzid2021pid}
Y.~Bouzid, S.~H. Derrouaoui, and M.~Guiatni, ``Pid gain scheduling for 3d
  trajectory tracking of a quadrotor with rotating and extendable arms,'' in
  \emph{2021 International Conference on Recent Advances in Mathematics and
  Informatics (ICRAMI)}.\hskip 1em plus 0.5em minus 0.4em\relax IEEE, 2021, pp.
  1--4.

\bibitem{glida2020optimal}
H.~E. Glida, L.~Abdou, A.~Chelihi, C.~Sentouh, and S.-E.-I. Hasseni, ``Optimal
  model-free backstepping control for a quadrotor helicopter,'' \emph{Nonlinear
  Dynamics}, vol. 100, pp. 3449--3468, 2020.

\bibitem{abdou2018integral}
L.~Abdou \emph{et~al.}, ``Integral backstepping/lft-lpv h∞ control for the
  trajectory tracking of a quadcopter,'' in \emph{2018 7th International
  Conference on Systems and Control (ICSC)}.\hskip 1em plus 0.5em minus
  0.4em\relax IEEE, 2018, pp. 348--353.

\bibitem{seif2020robust}
H.~Seif-El-Islam and L.~Abdou, ``Robust lpv control for attitude stabilization
  of a quadrotor helicopter under input saturations,'' \emph{Advances in
  Technology Innovation}, vol.~5, no.~2, p.~98, 2020.

\bibitem{419}
S.~Baldi, D.~Sun, X.~Xia, G.~Zhou, and D.~Liu, ``Ardupilot-based adaptive
  autopilot: architecture and software-in-the-loop experiments,'' \emph{IEEE
  Transactions on Aerospace and Electronic Systems}, vol.~58, no.~5, pp.
  4473--4485, 2022.

\bibitem{nguyen2019fuzzy}
A.-T. Nguyen, T.~Taniguchi, L.~Eciolaza, V.~Campos, R.~Palhares, and M.~Sugeno,
  ``Fuzzy control systems: Past, present and future,'' \emph{IEEE Computational
  Intelligence Magazine}, vol.~14, no.~1, pp. 56--68, 2019.

\bibitem{ferdaus2020towards}
M.~M. Ferdaus, S.~G. Anavatti, M.~Pratama, and M.~A. Garratt, ``Towards the use
  of fuzzy logic systems in rotary wing unmanned aerial vehicle: a review,''
  \emph{Artificial Intelligence Review}, vol.~53, no.~1, pp. 257--290, 2020.

\bibitem{wai2019adaptive}
R.-J. Wai and A.~S. Prasetia, ``Adaptive neural network control and optimal
  path planning of uav surveillance system with energy consumption
  prediction,'' \emph{Ieee Access}, vol.~7, pp. 126\,137--126\,153, 2019.

\bibitem{325}
P.~G. Fahlstrom, T.~J. Gleason, and M.~H. Sadraey, \emph{Introduction to UAV
  systems}.\hskip 1em plus 0.5em minus 0.4em\relax John Wiley \& Sons, 2022.

\bibitem{339}
L.~Bigazzi, M.~Basso, E.~Boni, G.~Innocenti, and M.~Pieraccini, ``A multilevel
  architecture for autonomous uavs,'' \emph{drones}, vol.~5, no.~3, p.~55,
  2021.

\bibitem{164}
F.~Ahmed and M.~Jenihhin, ``A survey on uav computing platforms: A hardware
  reliability perspective,'' \emph{Sensors}, vol.~22, no.~16, p. 6286, 2022.

\bibitem{322}
W.~Changpradith, ``Application of object detection using hardware acceleration
  for autonomous uav,'' 2022.

\bibitem{340}
G.~V. HrISToV, P.~Z. ZAHArIEV, and I.~H. BELoEV, ``A review of the
  characteristics of modern unmanned aerial vehicles,'' \emph{Acta technologica
  agriculturae}, vol.~19, no.~2, pp. 33--38, 2016.

\bibitem{341}
\BIBentryALTinterwordspacing
J.~Colorado~Monta{\~n}o, ``Towards mav autonomous flight: A modeling and
  control approach,'' Ph.D. dissertation, Industriales, May 2010, unpublished.
  [Online]. Available:
  \url{http://robcib.etsii.upm.es/index.php?option=com\%5fcontent&task=view&id=129&Itemid=154}
\BIBentrySTDinterwordspacing

\bibitem{344}
B.~L. Sharma, N.~Khatri, and A.~Sharma, ``An analytical review on fpga based
  autonomous flight control system for small uavs,'' in \emph{2016
  International Conference on Electrical, Electronics, and Optimization
  Techniques (ICEEOT)}.\hskip 1em plus 0.5em minus 0.4em\relax IEEE, 2016, pp.
  1369--1372.

\bibitem{346}
H.~Chao, Y.~Cao, and Y.~Chen, ``Autopilots for small unmanned aerial vehicles:
  a survey,'' \emph{International Journal of Control, Automation and Systems},
  vol.~8, pp. 36--44, 2010.

\bibitem{347}
N.~Monterrosa, J.~Montoya, F.~Jarqu{\'\i}n, and C.~Bran, ``Design, development
  and implementation of a uav flight controller based on a state machine
  approach using a fpga embedded system,'' in \emph{2016 IEEE/AIAA 35th Digital
  Avionics Systems Conference (DASC)}.\hskip 1em plus 0.5em minus 0.4em\relax
  IEEE, 2016, pp. 1--8.

\bibitem{sohail4348272deep}
S.~S. Sohail, Y.~Himeur, A.~Amira, F.~Fadli, W.~Mansoor, S.~Atalla, and
  A.~Copiaco, ``Deep transfer learning for 3d point cloud understanding: A
  comprehensive survey,'' \emph{Available at SSRN 4348272}.

\bibitem{himeur2022using}
Y.~Himeur, B.~Rimal, A.~Tiwary, and A.~Amira, ``Using artificial intelligence
  and data fusion for environmental monitoring: A review and future
  perspectives,'' \emph{Information Fusion}, vol. 86-87, pp. 44--75, 2022.

\bibitem{348}
L.~P{\'a}dua, J.~Vanko, J.~Hru{\v{s}}ka, T.~Ad{\~a}o, J.~J. Sousa, E.~Peres,
  and R.~Morais, ``Uas, sensors, and data processing in agroforestry: A review
  towards practical applications,'' \emph{International journal of remote
  sensing}, vol.~38, no. 8-10, pp. 2349--2391, 2017.

\bibitem{349}
N.~Amarasingam, S.~Salgadoe, K.~Powell, L.~F. Gonzalez, and S.~Natarajan, ``A
  review of uav platforms, sensors, and applications for monitoring of
  sugarcane crops,'' \emph{Remote Sensing Applications: Society and
  Environment}, p. 100712, 2022.

\bibitem{350}
D.~Olson and J.~Anderson, ``Review on unmanned aerial vehicles, remote sensors,
  imagery processing, and their applications in agriculture,'' \emph{Agronomy
  Journal}, vol. 113, no.~2, pp. 971--992, 2021.

\bibitem{352}
S.-G. Kim, E.~Lee, I.-P. Hong, and J.-G. Yook, ``Review of intentional
  electromagnetic interference on uav sensor modules and experimental study,''
  \emph{Sensors}, vol.~22, no.~6, p. 2384, 2022.

\bibitem{354}
S.~Samaras, E.~Diamantidou, D.~Ataloglou, N.~Sakellariou, A.~Vafeiadis,
  V.~Magoulianitis, A.~Lalas, A.~Dimou, D.~Zarpalas, K.~Votis \emph{et~al.},
  ``Deep learning on multi sensor data for counter uav applications—a
  systematic review,'' \emph{Sensors}, vol.~19, no.~22, p. 4837, 2019.

\bibitem{355}
S.~Ecke, J.~Dempewolf, J.~Frey, A.~Schwaller, E.~Endres, H.-J. Klemmt,
  D.~Tiede, and T.~Seifert, ``Uav-based forest health monitoring: A systematic
  review,'' \emph{Remote Sensing}, vol.~14, no.~13, p. 3205, 2022.

\bibitem{356}
D.~Joshi, D.~Deb, and S.~Muyeen, ``Comprehensive review on electric propulsion
  system of unmanned aerial vehicles,'' \emph{Frontiers in Energy Research}, p.
  739, 2022.

\bibitem{357}
D.~Teubl, T.~Bitenc, and M.~Hornung, ``Design and development of an actuator
  control and monitoring unit for small and medium size research uavs,''
  \emph{Deutsche Gesellschaft f{\"u}r Luft-und Raumfahrt-Lilienthal-Oberth eV,
  Bonn}, 2021.

\bibitem{358}
S.~Grundmann, M.~Frey, and C.~Tropea, ``Unmanned aerial vehicle (uav) with
  plasma actuators for separation control,'' in \emph{47th AIAA Aerospace
  Sciences Meeting including The New Horizons Forum and Aerospace Exposition},
  2009, p. 698.

\bibitem{362}
M.~N. Boukoberine, Z.~Zhou, and M.~Benbouzid, ``Power supply architectures for
  drones-a review,'' in \emph{IECON 2019-45th Annual Conference of the IEEE
  Industrial Electronics Society}, vol.~1.\hskip 1em plus 0.5em minus
  0.4em\relax IEEE, 2019, pp. 5826--5831.

\bibitem{363}
S.~A.~H. Mohsan, N.~Q.~H. Othman, M.~A. Khan, H.~Amjad, and J.~{\.Z}ywio{\l}ek,
  ``A comprehensive review of micro uav charging techniques,''
  \emph{Micromachines}, vol.~13, no.~6, p. 977, 2022.

\bibitem{364}
P.~K. Chittoor, B.~Chokkalingam, and L.~Mihet-Popa, ``A review on uav wireless
  charging: Fundamentals, applications, charging techniques and standards,''
  \emph{IEEE access}, vol.~9, pp. 69\,235--69\,266, 2021.

\bibitem{365}
L.~Xu, Y.~Huangfu, R.~Ma, R.~Xie, Z.~Song, D.~Zhao, Y.~Yang, Y.~Wang, and
  L.~Xu, ``A comprehensive review on fuel cell uav key technologies: Propulsion
  system, management strategy and design procedure,'' \emph{IEEE Transactions
  on Transportation Electrification}, 2022.

\bibitem{366}
C.~Zhang, Y.~Qiu, J.~Chen, Y.~Li, Z.~Liu, Y.~Liu, J.~Zhang, and C.~S. Hwa, ``A
  comprehensive review of electrochemical hybrid power supply systems and
  intelligent energy managements for unmanned aerial vehicles in public
  services,'' \emph{Energy and AI}, p. 100175, 2022.

\bibitem{367}
A.~Townsend, I.~N. Jiya, C.~Martinson, D.~Bessarabov, and R.~Gouws, ``A
  comprehensive review of energy sources for unmanned aerial vehicles, their
  shortfalls and opportunities for improvements,'' \emph{Heliyon}, vol.~6,
  no.~11, p. e05285, 2020.

\bibitem{369}
G.~Alsuhli, A.~Fahim, and Y.~Gadallah, ``A survey on the role of uavs in the
  communication process: A technological perspective,'' \emph{Computer
  Communications}, 2022.

\bibitem{49}
A.~Sharma, P.~Vanjani, N.~Paliwal, C.~M.~W. Basnayaka, D.~N.~K. Jayakody, H.-C.
  Wang, and P.~Muthuchidambaranathan, ``Communication and networking
  technologies for uavs: A survey,'' \emph{Journal of Network and Computer
  Applications}, vol. 168, p. 102739, 2020.

\bibitem{373}
X.~Chen, J.~Tang, and S.~Lao, ``Review of unmanned aerial vehicle swarm
  communication architectures and routing protocols,'' \emph{Applied Sciences},
  vol.~10, no.~10, p. 3661, 2020.

\bibitem{374}
M.~H.~M. Ghazali, K.~Teoh, and W.~Rahiman, ``A systematic review of real-time
  deployments of uav-based lora communication network,'' \emph{IEEE Access},
  vol.~9, pp. 124\,817--124\,830, 2021.

\bibitem{375}
N.~Islam, M.~M. Rashid, F.~Pasandideh, B.~Ray, S.~Moore, and R.~Kadel, ``A
  review of applications and communication technologies for internet of things
  (iot) and unmanned aerial vehicle (uav) based sustainable smart farming,''
  \emph{Sustainability}, vol.~13, no.~4, p. 1821, 2021.

\bibitem{250}
J.~Scott and C.~Scott, ``Drone delivery models for healthcare,'' 2017.

\bibitem{380}
S.~Brischetto and R.~Torre, ``Preliminary finite element analysis and flight
  simulations of a modular drone built through fused filament fabrication,''
  \emph{Journal of Composites Science}, vol.~5, no.~11, p. 293, 2021.

\bibitem{381}
A.~Martinetti, M.~Margaryan, and L.~van Dongen, ``Simulating mechanical stress
  on a micro unmanned aerial vehicle (uav) body frame for selecting maintenance
  actions,'' \emph{Procedia manufacturing}, vol.~16, pp. 61--66, 2018.

\bibitem{382}
A.~Mishra, S.~Pal, G.~Malhi, and P.~Singh, ``Structural analysis of uav
  airframe by using fem techniques: A review,'' \emph{International Journal of
  Mechanical and Production, ISSN}, pp. 2249--6890, 2020.

\bibitem{383}
J.~F. M.~A. Ferreira, ``Structural analysis and optimization of a uav wing,''
  2018.

\bibitem{384}
G.~Landolfo and A.~Altman, ``Aerodynamic and structural design of a small
  nonplanar wing uav,'' in \emph{47th AIAA Aerospace Sciences Meeting Including
  the New Horizons Forum and Aerospace Exposition}, 2009, p. 1068.

\bibitem{385}
M.~Kim, H.~Joo, and B.~Jang, ``Conceptual multicopter sizing and performance
  analysis via component database,'' in \emph{2017 Ninth International
  Conference on Ubiquitous and Future Networks (ICUFN)}.\hskip 1em plus 0.5em
  minus 0.4em\relax IEEE, 2017, pp. 105--109.

\bibitem{388}
L.~S. Souza, F.~G. Rocha, and M.~S. Soares, ``A review on software/systems
  architecture description for autonomous systems,'' \emph{Recent Advances in
  Computer Science and Communications (Formerly: Recent Patents on Computer
  Science)}, vol.~16, no.~3, pp. 52--60, 2023.

\bibitem{389}
R.~Spica, P.~R. Giordano, M.~Ryll, H.~H. B{\"u}lthoff, and A.~Franchi, ``An
  open-source hardware/software architecture for quadrotor uavs,'' \emph{IFAC
  Proceedings Volumes}, vol.~46, no.~30, pp. 198--205, 2013.

\bibitem{391}
{\'A}.~Madridano, A.~Al-Kaff, P.~Flores, D.~Mart{\'\i}n, and A.~de~la Escalera,
  ``Software architecture for autonomous and coordinated navigation of uav
  swarms in forest and urban firefighting,'' \emph{Applied Sciences}, vol.~11,
  no.~3, p. 1258, 2021.

\bibitem{392}
T.~Kekec, B.~C. Ustundag, M.~A. Guney, A.~Yildirim, and M.~Unel, ``A modular
  software architecture for uavs,'' in \emph{IECON 2013-39th Annual Conference
  of the IEEE Industrial Electronics Society}.\hskip 1em plus 0.5em minus
  0.4em\relax IEEE, 2013, pp. 4037--4042.

\bibitem{393}
E.~Pastor, J.~Lopez, and P.~Royo, ``A hardware/software architecture for uav
  payload and mission control,'' in \emph{2006 ieee/aiaa 25TH Digital Avionics
  Systems Conference}.\hskip 1em plus 0.5em minus 0.4em\relax IEEE, 2006, pp.
  1--8.

\bibitem{396}
------, ``A hardware/software architecture for uav payload and mission
  control,'' in \emph{2006 ieee/aiaa 25TH Digital Avionics Systems
  Conference}.\hskip 1em plus 0.5em minus 0.4em\relax IEEE, 2006, pp. 1--8.

\bibitem{397}
D.~D. York, A.~J. Al-Bayati, and Z.~Y. Al-Shabbani, ``Potential applications of
  uav within the construction industry and the challenges limiting
  implementation,'' in \emph{Construction Research Congress 2020: Project
  Management and Controls, Materials, and Contracts}.\hskip 1em plus 0.5em
  minus 0.4em\relax American Society of Civil Engineers Reston, VA, 2020, pp.
  31--39.

\bibitem{398}
N.~D. Opfer and D.~R. Shields, ``Unmanned aerial vehicle applications and
  issues for construction,'' in \emph{2014 ASEE Annual Conference \&
  Exposition}, 2014, pp. 24--1302.

\bibitem{399}
M.~Javaid, A.~Haleem, I.~H. Khan, R.~P. Singh, R.~Suman, and S.~Mohan,
  ``Significant features and applications of drones for healthcare: An
  overview,'' \emph{Journal of Industrial Integration and Management}, 2022.

\bibitem{400}
G.~Singhal, B.~Bansod, and L.~Mathew, ``Unmanned aerial vehicle classification,
  applications and challenges: A review,'' 2018.

\bibitem{401}
S.~Ahirwar, R.~Swarnkar, S.~Bhukya, and G.~Namwade, ``Application of drone in
  agriculture,'' \emph{International Journal of Current Microbiology and
  Applied Sciences}, vol.~8, no.~01, pp. 2500--2505, 2019.

\bibitem{402}
Y.~Li and C.~Liu, ``Applications of multirotor drone technologies in
  construction management,'' \emph{International Journal of Construction
  Management}, vol.~19, no.~5, pp. 401--412, 2019.

\bibitem{403}
S.~Manfreda, M.~F. McCabe, P.~E. Miller, R.~Lucas, V.~Pajuelo~Madrigal,
  G.~Mallinis, E.~Ben~Dor, D.~Helman, L.~Estes, G.~Ciraolo \emph{et~al.}, ``On
  the use of unmanned aerial systems for environmental monitoring,''
  \emph{Remote sensing}, vol.~10, no.~4, p. 641, 2018.

\bibitem{405}
R.~Majeed, N.~A. Abdullah, M.~F. Mushtaq, and R.~Kazmi, ``Drone security:
  Issues and challenges,'' \emph{Parameters}, vol.~2, p. 5GHz, 2021.

\bibitem{406}
S.~Mirri, C.~Prandi, and P.~Salomoni, ``Human-drone interaction: state of the
  art, open issues and challenges,'' in \emph{Proceedings of the ACM SIGCOMM
  2019 Workshop on Mobile AirGround Edge Computing, Systems, Networks, and
  Applications}, 2019, pp. 43--48.

\bibitem{409}
H.~Nawaz, H.~M. Ali, and A.~A. Laghari, ``Uav communication networks issues: a
  review,'' \emph{Archives of Computational Methods in Engineering}, vol.~28,
  pp. 1349--1369, 2021.

\bibitem{465}
\BIBentryALTinterwordspacing
Px4, pixhawk engineering team. [Online]. Available:
  \url{https://github.com/PX4/jMAVSim}
\BIBentrySTDinterwordspacing

\bibitem{416}
L.~He, N.~Aouf, and B.~Song, ``Explainable deep reinforcement learning for uav
  autonomous path planning,'' \emph{Aerospace science and technology}, vol.
  118, p. 107052, 2021.

\bibitem{418}
\BIBentryALTinterwordspacing
 [Online]. Available: \url{https://ardupilot.org/ardupilot}
\BIBentrySTDinterwordspacing

\bibitem{422}
\BIBentryALTinterwordspacing
Tensorflow. [Online]. Available: \url{https://www.tensorflow.org}
\BIBentrySTDinterwordspacing

\bibitem{423}
M.~T. Topalli, M.~Yilmaz, and M.~F. {\c{C}}orapsiz, ``Real time implementation
  of drone detection using tensorflow and mobilenetv2-ssd,'' in \emph{2021 7th
  International Conference on Electrical, Electronics and Information
  Engineering (ICEEIE)}.\hskip 1em plus 0.5em minus 0.4em\relax IEEE, 2021, pp.
  436--439.

\bibitem{424}
Z.~Jiang, Y.~Liu, B.~Wu, and Q.~Zhu, ``Monocular vision based uav target
  detection and ranging system implemented on opencv and tensor flow,'' in
  \emph{2019 18th International Symposium on Distributed Computing and
  Applications for Business Engineering and Science (DCABES)}.\hskip 1em plus
  0.5em minus 0.4em\relax IEEE, 2019, pp. 88--91.

\bibitem{425}
\BIBentryALTinterwordspacing
 [Online]. Available: \url{https://wiki.paparazziuav.org}
\BIBentrySTDinterwordspacing

\bibitem{426}
B.~Remes, D.~Hensen, F.~Van~Tienen, C.~De~Wagter, E.~Van~der Horst, and
  G.~De~Croon, ``Paparazzi: how to make a swarm of parrot ar drones fly
  autonomously based on gps,'' in \emph{International Micro Air Vehicle
  Conference and Flight Competition (IMAV2013)}, 2013, pp. 17--20.

\bibitem{427}
J.~Garcia, A.~Brock, N.~Saporito, G.~Hattenberger, X.~Paris, M.~Gorraz, and
  Y.~Jestin, ``Designing human-drone interactions with the paparazzi uav
  system,'' in \emph{1st International Workshop on Human-Drone
  Interaction-CHI'19}, 2019.

\bibitem{428}
\BIBentryALTinterwordspacing
 [Online]. Available: \url{http://www.qgroundcontrol.org}
\BIBentrySTDinterwordspacing

\bibitem{429}
\BIBentryALTinterwordspacing
Mission planner. [Online]. Available: \url{http://ardupilot.org/planner}
\BIBentrySTDinterwordspacing

\bibitem{430}
\BIBentryALTinterwordspacing
Apm planner 2. [Online]. Available: \url{https://ardupilot.org/planner2/}
\BIBentrySTDinterwordspacing

\bibitem{431}
\BIBentryALTinterwordspacing
Ugcs. [Online]. Available: \url{https://www.ugcs.com}
\BIBentrySTDinterwordspacing

\bibitem{432}
C.~Ramirez-Atencia and D.~Camacho, ``Extending qgroundcontrol for automated
  mission planning of uavs,'' \emph{Sensors}, vol.~18, no.~7, p. 2339, 2018.

\bibitem{433}
T.~Dardoize, N.~Ciochetto, J.-H. Hong, and H.-S. Shin, ``Implementation of
  ground control system for autonomous multi-agents using qgroundcontrol,'' in
  \emph{2019 Workshop on Research, Education and Development of Unmanned Aerial
  Systems (RED UAS)}.\hskip 1em plus 0.5em minus 0.4em\relax IEEE, 2019, pp.
  24--30.

\bibitem{434}
G.~Vachtsevanos, L.~Tang, G.~Drozeski, and L.~Gutierrez, ``From mission
  planning to flight control of unmanned aerial vehicles: Strategies and
  implementation tools,'' \emph{Annual Reviews in Control}, vol.~29, no.~1, pp.
  101--115, 2005.

\bibitem{435}
\BIBentryALTinterwordspacing
Airsim. [Online]. Available: \url{https://microsoft.github.io/AirSim}
\BIBentrySTDinterwordspacing

\bibitem{436}
R.~Madaan, N.~Gyde, S.~Vemprala, M.~Brown, K.~Nagami, T.~Taubner,
  E.~Cristofalo, D.~Scaramuzza, M.~Schwager, and A.~Kapoor, ``Airsim drone
  racing lab,'' in \emph{Neurips 2019 competition and demonstration
  track}.\hskip 1em plus 0.5em minus 0.4em\relax PMLR, 2020, pp. 177--191.

\bibitem{437}
D.~Villota~Miranda, M.~Gil~Mart{\'\i}nez, and J.~Rico-Azagra, ``A3c for drone
  autonomous driving using airsim,'' in \emph{XLII Jornadas de
  Autom{\'a}tica}.\hskip 1em plus 0.5em minus 0.4em\relax Universidade da
  Coru{\~n}a, Servizo de Publicaci{\'o}ns, 2021, pp. 203--209.

\bibitem{438}
C.~Ma, Y.~Zhou, and Z.~Li, ``A new simulation environment based on airsim, ros,
  and px4 for quadcopter aircrafts,'' in \emph{2020 6th International
  Conference on Control, Automation and Robotics (ICCAR)}.\hskip 1em plus 0.5em
  minus 0.4em\relax IEEE, 2020, pp. 486--490.

\bibitem{439}
\BIBentryALTinterwordspacing
Jderobot. [Online]. Available:
  \url{https://jderobot.github.io/projects/drones/drones}
\BIBentrySTDinterwordspacing

\bibitem{440}
J.~M. Ca{\~n}as, E.~Perdices, L.~Garc{\'\i}a-P{\'e}rez, and
  J.~Fern{\'a}ndez-Conde, ``A ros-based open tool for intelligent robotics
  education,'' \emph{Applied Sciences}, vol.~10, no.~21, p. 7419, 2020.

\bibitem{441}
P.~Arias-Perez, J.~Fern{\'a}ndez-Conde, D.~Martin~Gomez, J.~M. Ca{\~n}as, and
  P.~Campoy, ``A middleware infrastructure for programming vision-based
  applications in uavs,'' \emph{Drones}, vol.~6, no.~11, p. 369, 2022.

\bibitem{442}
\BIBentryALTinterwordspacing
Dronekit. [Online]. Available: \url{https://dronekit.io}
\BIBentrySTDinterwordspacing

\bibitem{443}
\BIBentryALTinterwordspacing
Dronekit-python. [Online]. Available: \url{dronekit-python.readthedocs.io}
\BIBentrySTDinterwordspacing

\bibitem{444}
A.~M. Gaber, R.~A. Rashid, N.~Nasir, R.~A. Rahim, M.~A. Sarijari, A.~S.
  Abdullah, O.~A. Aziz, S.~Z.~A. Hamid, and S.~Ali, ``Development of an
  autonomous iot-based drone for campus security,'' \emph{ELEKTRIKA-Journal of
  Electrical Engineering}, vol.~20, no. 2-2, pp. 70--76, 2021.

\bibitem{445}
M.-T. Lee, M.-L. Chuang, S.-T. Kuo, and Y.-R. Chen, ``Uav swarm real-time
  rerouting by edge computing d* lite algorithm,'' \emph{Applied Sciences},
  vol.~12, no.~3, p. 1056, 2022.

\bibitem{446}
P.~Y. Ingle, Y.~Kim, and Y.-G. Kim, ``Dvs: A drone video synopsis towards
  storing and analyzing drone surveillance data in smart cities,''
  \emph{Systems}, vol.~10, no.~5, p. 170, 2022.

\bibitem{447}
\BIBentryALTinterwordspacing
Mavlink. [Online]. Available: \url{https://mavlink.io/}
\BIBentrySTDinterwordspacing

\bibitem{448}
P.~Pratik, P.~Agarwal \emph{et~al.}, ``Review on mav link for unmanned air
  vehicle to ground control station communication,'' \emph{Think India
  Journal}, vol.~22, no.~4, pp. 7093--7103, 2019.

\bibitem{449}
S.~Badole, S.~Choudhary, A.~Titarmare, and P.~Khergade, ``Review on ground
  control station design for remotely piloted aircraft system,'' in \emph{2022
  10th International Conference on Emerging Trends in Engineering and
  Technology-Signal and Information Processing (ICETET-SIP-22)}.\hskip 1em plus
  0.5em minus 0.4em\relax IEEE, 2022, pp. 1--6.

\bibitem{450}
U.~R. Mogili and B.~Deepak, ``An intelligent drone for agriculture applications
  with the aid of the mavlink protocol,'' in \emph{Innovative Product Design
  and Intelligent Manufacturing Systems: Select Proceedings of ICIPDIMS
  2019}.\hskip 1em plus 0.5em minus 0.4em\relax Springer, 2020, pp. 195--205.

\bibitem{451}
A.~Koub{\^a}a, A.~Allouch, M.~Alajlan, Y.~Javed, A.~Belghith, and M.~Khalgui,
  ``Micro air vehicle link (mavlink) in a nutshell: A survey,'' \emph{IEEE
  Access}, vol.~7, pp. 87\,658--87\,680, 2019.

\bibitem{452}
\BIBentryALTinterwordspacing
Ros robot operating system. [Online]. Available: \url{https://www.ros.org}
\BIBentrySTDinterwordspacing

\bibitem{453}
B.~Abbyasov, R.~Lavrenov, A.~Zakiev, T.~Tsoy, E.~Magid, M.~Svinin, and E.~A.
  Martinez-Garcia, ``Comparative analysis of ros-based centralized methods for
  conducting collaborative monocular visual slam using a pair of uavs,'' in
  \emph{Proceedings of the 23rd International Conference on Climbing and
  Walking Robots and Support Technologies for Mobile Machines}, 2020, pp.
  113--120.

\bibitem{454}
D.~Canpolat~Tosun and Y.~I{\c{s}}{\i}k, ``A ros-based hybrid algorithm for the
  uav path planning problem,'' \emph{Aircraft Engineering and Aerospace
  Technology}, vol.~95, no.~5, pp. 784--798, 2023.

\bibitem{455}
R.~K. Megalingam, D.~V. Prithvi, N.~C.~S. Kumar, and V.~Egumadiri, ``Drone
  stability simulation using ros and gazebo,'' in \emph{Advanced Computing and
  Intelligent Technologies: Proceedings of ICACIT 2021}.\hskip 1em plus 0.5em
  minus 0.4em\relax Springer, 2022, pp. 131--143.

\bibitem{456}
R.~Mardiyanto, M.~N. Hisak, T.~Mujiono, and H.~Suryoatmojo, ``Robot operating
  system (ros) framework for swarm drone flight controller,'' in \emph{2022
  International Seminar on Intelligent Technology and Its Applications
  (ISITIA)}.\hskip 1em plus 0.5em minus 0.4em\relax IEEE, 2022, pp. 297--303.

\bibitem{457}
N.~Jain, A.~K. Gupta, and P.~Mathur, ``Autonomous drone using ros for
  surveillance and 3d mapping using satellite map,'' in \emph{Proceedings of
  the Second International Conference on Information Management and Machine
  Intelligence: ICIMMI 2020}.\hskip 1em plus 0.5em minus 0.4em\relax Springer,
  2021, pp. 255--266.

\bibitem{458}
Y.~Yu, X.~Wang, Z.~Zhong, and Y.~Zhang, ``Ros-based uav control using hand
  gesture recognition,'' in \emph{2017 29th Chinese Control And Decision
  Conference (CCDC)}.\hskip 1em plus 0.5em minus 0.4em\relax IEEE, 2017, pp.
  6795--6799.

\bibitem{459}
T.~Zhao and H.~Jiang, ``Landing system for ar. drone 2.0 using onboard camera
  and ros,'' in \emph{2016 IEEE Chinese Guidance, Navigation and Control
  Conference (CGNCC)}.\hskip 1em plus 0.5em minus 0.4em\relax IEEE, 2016, pp.
  1098--1102.

\bibitem{460}
W.~Meng, Y.~Hu, J.~Lin, F.~Lin, and R.~Teo, ``Ros+ unity: An efficient
  high-fidelity 3d multi-uav navigation and control simulator in gps-denied
  environments,'' in \emph{IECON 2015-41st Annual Conference of the IEEE
  Industrial Electronics Society}.\hskip 1em plus 0.5em minus 0.4em\relax IEEE,
  2015, pp. 002\,562--002\,567.

\bibitem{461}
A.~P. Lamping, J.~N. Ouwerkerk, and K.~Cohen, ``Multi-uav control and
  supervision with ros,'' in \emph{2018 aviation technology, integration, and
  operations conference}, 2018, p. 4245.

\bibitem{462}
\BIBentryALTinterwordspacing
Gazebosim, gazebo. [Online]. Available: \url{https://gazebosim.org/home}
\BIBentrySTDinterwordspacing

\bibitem{463}
\BIBentryALTinterwordspacing
Webots, cyberbotics. [Online]. Available: \url{https://cyberbotics.com}
\BIBentrySTDinterwordspacing

\bibitem{464}
\BIBentryALTinterwordspacing
Morse, laas and onera lab. [Online]. Available:
  \url{https://github.com/morse-simulator/morse}
\BIBentrySTDinterwordspacing

\bibitem{466}
\BIBentryALTinterwordspacing
Paparazzi. [Online]. Available: \url{https://github.com/paparazzi}
\BIBentrySTDinterwordspacing

\bibitem{467}
\BIBentryALTinterwordspacing
Hackflightsim. [Online]. Available:
  \url{https://github.com/simondlevy/HackflightSim}
\BIBentrySTDinterwordspacing

\bibitem{UAVToolbox}
\BIBentryALTinterwordspacing
Uav toolbox. [Online]. Available:
  \url{https://www.mathworks.com/products/uav.html}
\BIBentrySTDinterwordspacing

\bibitem{horri2022tutorial}
N.~Horri and M.~Pietraszko, ``A tutorial and review on flight control
  co-simulation using matlab/simulink and flight simulators,''
  \emph{Automation}, vol.~3, no.~3, pp. 486--510, 2022.

\bibitem{aliane2022web}
N.~Aliane, C.~Q.~G. Mu{\~n}oz, and J.~S{\'a}nchez-Soriano, ``Web and
  matlab-based platform for uav flight management and multispectral image
  processing,'' \emph{Sensors}, vol.~22, no.~11, p. 4243, 2022.

\bibitem{xing2015design}
Z.~Xing, Y.~He, and C.~Jian, ``Design and implementation of uav flight
  simulation based on matlab/simulink,'' in \emph{2015 International Conference
  on Advances in Mechanical Engineering and Industrial Informatics}, 2015, pp.
  190--193.

\bibitem{pinedaprm}
F.~Pineda-Torres and L.~A. Arias-Barrag{\'a}n, ``Prm navigation in trading
  drone and gazebo simulation.''

\bibitem{ivaldi2014tools}
S.~Ivaldi, V.~Padois, and F.~Nori, ``Tools for dynamics simulation of robots: a
  survey based on user feedback,'' \emph{arXiv preprint arXiv:1402.7050}, 2014.

\bibitem{468}
\BIBentryALTinterwordspacing
Multiwii series. [Online]. Available: \url{https://github.com/multiwii}
\BIBentrySTDinterwordspacing

\bibitem{469}
\BIBentryALTinterwordspacing
Cleanflight. [Online]. Available: \url{http://cleanflight.com/}
\BIBentrySTDinterwordspacing

\bibitem{470}
\BIBentryALTinterwordspacing
Betaflight. [Online]. Available:
  \url{https://github.com/betaflight/betaflight/wiki/670}
\BIBentrySTDinterwordspacing

\bibitem{471}
\BIBentryALTinterwordspacing
Inav, inav flight. [Online]. Available:
  \url{https://github.com/iNavFlight/inav/wiki}
\BIBentrySTDinterwordspacing

\bibitem{472}
\BIBentryALTinterwordspacing
Openpilot, openpilot wiki. [Online]. Available:
  \url{https://opwiki.readthedocs.io/en/latest}
\BIBentrySTDinterwordspacing

\bibitem{473}
\BIBentryALTinterwordspacing
Librepilot project. [Online]. Available: \url{http://librepilot.org}
\BIBentrySTDinterwordspacing

\bibitem{474}
\BIBentryALTinterwordspacing
dronin. [Online]. Available: \url{http://dronin.org/}
\BIBentrySTDinterwordspacing

\bibitem{478}
\BIBentryALTinterwordspacing
Opencv. [Online]. Available: \url{https://opencv.org}
\BIBentrySTDinterwordspacing

\bibitem{479}
\BIBentryALTinterwordspacing
Pytorch. [Online]. Available: \url{https://pytorch.org}
\BIBentrySTDinterwordspacing

\bibitem{azmat2020potential}
M.~Azmat and S.~Kummer, ``Potential applications of unmanned ground and aerial
  vehicles to mitigate challenges of transport and logistics-related critical
  success factors in the humanitarian supply chain,'' \emph{Asian journal of
  sustainability and social responsibility}, vol.~5, no.~1, pp. 1--22, 2020.

\bibitem{iqbal2022motion}
M.~M. Iqbal, Z.~A. Ali, R.~Khan, and M.~Shafiq, ``Motion planning of uav swarm:
  Recent challenges and approaches,'' \emph{Aeronautics-New Advances}, 2022.

\bibitem{vattapparamban2016drones}
E.~Vattapparamban, I.~G{\"u}ven{\c{c}}, A.~I. Yurekli, K.~Akkaya, and
  S.~Ulua{\u{g}}a{\c{c}}, ``Drones for smart cities: Issues in cybersecurity,
  privacy, and public safety,'' in \emph{2016 international wireless
  communications and mobile computing conference (IWCMC)}.\hskip 1em plus 0.5em
  minus 0.4em\relax IEEE, 2016, pp. 216--221.

\bibitem{shakhatreh2019unmanned}
H.~Shakhatreh, A.~H. Sawalmeh, A.~Al-Fuqaha, Z.~Dou, E.~Almaita, I.~Khalil,
  N.~S. Othman, A.~Khreishah, and M.~Guizani, ``Unmanned aerial vehicles
  (uavs): A survey on civil applications and key research challenges,''
  \emph{Ieee Access}, vol.~7, pp. 48\,572--48\,634, 2019.

\bibitem{kirschstein2021energy}
T.~Kirschstein, ``Energy demand of parcel delivery services with a mixed fleet
  of electric vehicles,'' \emph{Cleaner Engineering and Technology}, vol.~5, p.
  100322, 2021.

\bibitem{gautam2022drone}
T.~Gautam and R.~Johari, ``Drone: A systematic review of uav technologies,'' in
  \emph{International Conference on Computing, Communications, and
  Cyber-Security}.\hskip 1em plus 0.5em minus 0.4em\relax Springer, 2022, pp.
  147--158.

\bibitem{elmeseiry2021detailed}
N.~Elmeseiry, N.~Alshaer, and T.~Ismail, ``A detailed survey and future
  directions of unmanned aerial vehicles (uavs) with potential applications,''
  \emph{Aerospace}, vol.~8, no.~12, p. 363, 2021.

\bibitem{heidari2023secure}
A.~Heidari, N.~J. Navimipour, and M.~Unal, ``A secure intrusion detection
  platform using blockchain and radial basis function neural networks for
  internet of drones,'' \emph{IEEE Internet of Things Journal}, 2023.

\bibitem{himeur2022blockchain}
Y.~Himeur, A.~Sayed, A.~Alsalemi, F.~Bensaali, A.~Amira, I.~Varlamis,
  M.~Eirinaki, C.~Sardianos, and G.~Dimitrakopoulos, ``Blockchain-based
  recommender systems: Applications, challenges and future opportunities,''
  \emph{Computer Science Review}, vol.~43, p. 100439, 2022.

\end{thebibliography}
\end{document}